\documentclass[10pt,twocolumn,journal]{IEEEtran}
\usepackage{xr}
\usepackage{cite}
\usepackage{amsmath}
\usepackage[colorlinks=true,linkcolor=black,citecolor=black,urlcolor=black]{hyperref}
\usepackage{cases}
\usepackage[utf8]{inputenc}
\usepackage[english]{babel}
\usepackage{footnote}
\usepackage{booktabs}
\usepackage{tablefootnote}
\usepackage[mathscr]{eucal}
\usepackage{epsfig,epsf,psfrag}
\usepackage{amssymb,amsmath,amsfonts,latexsym}
\usepackage{amsmath,graphicx,bm,xcolor,url}
\usepackage[caption=false]{subfig} 
\usepackage{fixltx2e}%ordering of single and double column floats
\usepackage{array}%array and tabular environments
\usepackage{verbatim}
\usepackage{bm}
\usepackage{algpseudocode, cite}
\usepackage{algorithm}
\usepackage{verbatim}
\usepackage{textcomp}
\usepackage{mathrsfs}
\usepackage{epstopdf}
\usepackage{relsize}
\usepackage{cleveref} 
\usepackage{subfig}

\catcode`~=11 \def\UrlSpecials{\do\~{\kern -.15em\lower .7ex\hbox{~}\kern .04em}} \catcode`~=13 

\allowdisplaybreaks[3]

\newcommand{\nn}{\nonumber}

\newcommand{\calP}{\mathcal{P}}

\newcommand{\bA}{\mathbf{A}}

\newcommand{\bD}{\mathbf{D}}
\newcommand{\be}{\mathbf{e}}
\newcommand{\bE}{\mathbf{E}}

\newcommand{\bh}{\mathbf{h}}
\newcommand{\bH}{\mathbf{H}}

\newcommand{\bI}{\mathbf{I}}

\newcommand{\bp}{\mathbf{p}}
\newcommand{\bP}{\mathbf{P}}

\newcommand{\bS}{\mathbf{S}}

\newcommand{\bu}{\mathbf{u}}
\newcommand{\bU}{\mathbf{U}}
\newcommand{\bv}{\mathbf{v}}
\newcommand{\bV}{\mathbf{V}}
\newcommand{\bw}{\mathbf{w}}
\newcommand{\bW}{\mathbf{W}}
\newcommand{\bx}{\mathbf{x}}
\newcommand{\bX}{\mathbf{X}}
\newcommand{\by}{\mathbf{y}}

\newcommand{\bz}{\mathbf{z}}

\newcommand{\rmF}{\mathrm{F}}

\newcommand{\rmp}{\mathrm{p}}

\newcommand{\bbP}{\mathbb{P}}

\newcommand{\bbR}{\mathbb{R}}

\DeclareMathAlphabet{\mathbsf}{OT1}{cmss}{bx}{n}
\DeclareMathAlphabet{\mathssf}{OT1}{cmss}{m}{sl}% slanted sans serif

\DeclareSymbolFont{bsfletters}{OT1}{cmss}{bx}{n}  
\DeclareSymbolFont{ssfletters}{OT1}{cmss}{m}{n}
\DeclareMathSymbol{\bsfGamma}{0}{bsfletters}{'000}
\DeclareMathSymbol{\ssfGamma}{0}{ssfletters}{'000}
\DeclareMathSymbol{\bsfDelta}{0}{bsfletters}{'001}
\DeclareMathSymbol{\ssfDelta}{0}{ssfletters}{'001}
\DeclareMathSymbol{\bsfTheta}{0}{bsfletters}{'002}
\DeclareMathSymbol{\ssfTheta}{0}{ssfletters}{'002}
\DeclareMathSymbol{\bsfLambda}{0}{bsfletters}{'003}
\DeclareMathSymbol{\ssfLambda}{0}{ssfletters}{'003}
\DeclareMathSymbol{\bsfXi}{0}{bsfletters}{'004}
\DeclareMathSymbol{\ssfXi}{0}{ssfletters}{'004}
\DeclareMathSymbol{\bsfPi}{0}{bsfletters}{'005}
\DeclareMathSymbol{\ssfPi}{0}{ssfletters}{'005}
\DeclareMathSymbol{\bsfSigma}{0}{bsfletters}{'006}
\DeclareMathSymbol{\ssfSigma}{0}{ssfletters}{'006}
\DeclareMathSymbol{\bsfUpsilon}{0}{bsfletters}{'007}
\DeclareMathSymbol{\ssfUpsilon}{0}{ssfletters}{'007}
\DeclareMathSymbol{\bsfPhi}{0}{bsfletters}{'010}
\DeclareMathSymbol{\ssfPhi}{0}{ssfletters}{'010}
\DeclareMathSymbol{\bsfPsi}{0}{bsfletters}{'011}
\DeclareMathSymbol{\ssfPsi}{0}{ssfletters}{'011}
\DeclareMathSymbol{\bsfOmega}{0}{bsfletters}{'012}
\DeclareMathSymbol{\ssfOmega}{0}{ssfletters}{'012}

\newcommand{\hatK}{\hat{K}}

\newcommand{\blambda}{\bm{\lambda}}

\newcommand{\bSigma	}{\bm{\Sigma}}

\DeclareMathOperator*{\argmax}{arg\,max}

\newtheorem{theorem}{Theorem} 
\newtheorem{lemma}[theorem]{Lemma}
\newtheorem{definition}{Definition}

\newcommand{\qednew}{\nobreak \ifvmode \relax \else
      \ifdim\lastskip<1.5em \hskip-\lastskip
      \hskip1.5em plus0em minus0.5em \fi \nobreak
      \vrule height0.75em width0.5em depth0.25em\fi}

\graphicspath{{../figures/}}

\newtheorem{remark}{Remark}

\begin{document}
%\nocite{*}
%\nocite{*}

%\title{Relative Error Bounds  for  NMF under a Geometric Assumption  and an Initialization Strategy}
%\title{An Initialization Method for NMF based on Rank-One NMF and Relative Error Bounds   under a Geometric Assumption  }
\title{Rank-One NMF-Based Initialization for NMF and Relative Error Bounds under a Geometric Assumption  }

\author{Zhaoqiang~Liu 
        and~Vincent~Y.~F.~Tan, \IEEEmembership{Senior~Member,~IEEE}
        
%\thanks{A preliminary work has been published in ICASSP 2016 \cite{Zhao_16}.} 

%\thanks{Copyright (c) 2015 IEEE. Personal use of this material is permitted. However, permission to use this material for any other purposes must be obtained from the IEEE by sending a request to pubs-permissions@ieee.org.}

\thanks{This paper was presented in part at ICASSP 2017.} 

\thanks{The authors are with the Department of Mathematics, National University of Singapore (NUS).  The second author is also with the Department of Electrical and Computer Engineering, NUS.}
\thanks{The work of Z.~Liu (zqliu12@gmail.com) is supported by an NUS  Research Scholarship. The work of V.~Y.~F~Tan (vtan@nus.edu.sg)   is supported  in part by an NUS grant (R-263-000-B37-133). }
}
%\thanks{M. Shell was with the Department
%of Electrical and Computer Engineering, Georgia Institute of Technology, Atlanta,
%GA, 30332 USA e-mail: (see http://www.michaelshell.org/contact.html).}% <-this % stops a space
%\thanks{J. Doe and J. Doe are with Anonymous University.}% <-this % stops a space
%\thanks{Manuscript received April 19, 2005; revised August 26, 2015.}}

\markboth{IEEE Transactions on Signal Processing,~Vol.~, No.~, 2016}%
{Shell \MakeLowercase{\textit{Liu et al.}}: Rank-One NMF-Based Initialization for NMF and Relative Error Bounds under a Geometric Assumption}

\maketitle

\begin{abstract}
We propose a geometric assumption on nonnegative data matrices such that under this assumption,  we are able to provide upper bounds (both deterministic and probabilistic) on the relative error of nonnegative matrix factorization (NMF). The algorithm we propose first uses the geometric assumption to obtain an exact clustering of the columns of the data matrix; subsequently, it employs several rank-one NMFs to obtain the final decomposition. When applied to data matrices generated from our statistical model, we observe that our proposed algorithm produces factor matrices with comparable relative errors vis-\`a-vis classical NMF algorithms but with much faster speeds. 
On face image and hyperspectral imaging datasets, we demonstrate that our algorithm provides an excellent initialization for applying other NMF algorithms at a low computational cost. Finally, we show on face and text datasets that the combinations of our algorithm and several classical NMF algorithms outperform other algorithms in terms of clustering performance. 
\end{abstract}


%\vspace{-.1in}
\section{Introduction}
\label{sec:intro}
%\vspace{-.05in}

The nonnegative matrix factorization (NMF) problem can be formulated as follows: Given a nonnegative data matrix $\bV \in \mathbb{R}^{F\times N}_{+}$  and a positive integer $K$, we seek  nonnegative factor matrices $\bW \in \mathbb{R}^{F\times K}_{+}$ and $\bH \in \mathbb{R}^{K\times N}_{+}$, such that the distance (measured in some norm) between $\bV$ and $\bW \bH$ is minimized. Due to its non-subtractive, parts-based property which enhances interpretability, NMF has been widely used in  machine learning \cite{Ci_09} and signal processing \cite{Bu_08} among others. In addition, there are many fundamental algorithms to approximately solve the NMF problem, including the multiplicative update algorithms \cite{Lee_00}, the alternating (nonnegative) least-squares-type algorithms \cite{Chu_04, Kim_08b, Kim_08a, Kim_11}, and the hierarchical alternating least square algorithms \cite{Cichocki07} (also called the rank-one residual iteration \cite{Ho_11}). However, it is proved in \cite{Va_09} that NMF problem is NP-hard and all the basic algorithms simply either  ensure that the sequence of  objective functions is non-increasing or that the algorithm converges to the set of stationary points \cite{Lin_07a, Lin_07b, Ho_11}. To the best of our knowledge, none of these algorithms is suitable for analyzing a bound on the approximation error of NMF.

In an effort to find computationally tractable algorithms for NMF and to provide theoretical guarantees on the errors of these algorithms, researchers have revisited the so-called {\em separability assumption} proposed by Donoho and Stodden~\cite{Donoho_04}.  
An exact nonnegative factorization $\bV=\bW \bH$ is {\em separable} if for any $k \in \{1,2,\ldots,K\}$, there is an $n(k) \in \{1,2,\ldots,F\}$ such that $\bW(n(k),j)=0$ for all $j \neq k$ and $\bW (n(k),k)>0$. That is, an exact nonnegative factorization is separable if all the  features can be represented as nonnegative linear combinations of $K$ features. It is proved in \cite{Arora_12} that under the separability condition, there is an algorithm that runs in time polynomial in $F$, $N$ and $K$ and outputs a separable nonnegative factorization $\bV=\bW^{*}\bH^{*}$ with the number of columns of $\bW^{*}$ being at most $K$. Furthermore,  to handle noisy data, a perturbation analysis of their algorithm is presented. The authors assumed that  $\bV$ is normalized such that every row of it has unit $\ell_{1}$ norm and $\bV$ has a separable nonnegative factorization $\bV=\bW \bH$. In addition, each row of $\bV$ is perturbed by adding a vector of small $\ell_{1}$ norm to obtain a new data matrix $\bV'$. With  additional assumptions on the noise and $\bH$, their algorithm leads to an approximate nonnegative matrix factorization $\bW'\bH'$ of $\bV'$ with a provable error bound for the $\ell_{1}$ norm of each row of $\bV'-\bW'\bH'$. To develop more efficient algorithms and to extend the basic formulation to more general noise models, a collection of elegant papers dealing with NMF under various separability conditions has emerged~\cite{Gillis_14a ,Bittorf_12, Kumar_13, Ben_14, Gillis_14b}.  

%In the applications of the separability of NMF, people usually use certain salient features to represent all the features, instead of using certain special samples to represent all the samples. Thus it is unrealistic to give probabilistic estimates for the error bound. 
%\vspace{-.05in}
\subsection{Main Contributions}%\vspace{-.05in}
\subsubsection{Theoretical Contributions} \label{sec:thm_contribution}
We introduce a geometric assumption on the data matrix $\bV$ that allows us to correctly group columns of $\bV$ into disjoint subsets. This naturally suggests an algorithm that first clusters the columns and subsequently uses a rank-one approximate NMF algorithm~\cite{Gon_09} to obtain  the final decomposition. We analyze the error performance and provide a deterministic  upper bound on the relative error.  We also consider a random data generation model and provide a probabilistic relative error bound. Our geometric assumption  can be considered as a special case of  the separability (or, more precisely, the near-separability)  assumption~\cite{Donoho_04}. However, there are certain key differences: First, because our assumption is based on a notion of clusterability \cite{Ack_09}, our proof technique is different from those in the literature  that leverage the separability condition. Second, unlike most works that assume separability~\cite{Gillis_14a ,Bittorf_12, Kumar_13, Ben_14, Gillis_14b}, we exploit the $\ell_2$ norm of vectors instead of the $\ell_1$ norm of vectors/matrices. Third, $\bV$ does not need to be assumed to be normalized. As pointed out in \cite{Kumar_13}, normalization, especially in the $\ell_{1}$-norm for the rows of data matrices may deteriorate the clustering performance for text datasets significantly. Fourth, we provide an upper bound for relative error instead of the absolute error. Our work is the first to provide theoretical analyses for the relative error for near-separable-type NMF problems. Finally, we assume all the samples can be approximately represented by certain special samples (e.g., centroids) instead of using a small set of salient features to represent all the features. Mathematically, these two approximations may appear to be equivalent. However, our assumption and analysis techniques enable us to provide an efficient algorithm and  tight probabilistic relative error bounds for the NMF approximation (cf.~Theorem~\ref{thm:prob}).

\subsubsection{Experimental Evaluations}\label{sec: main_contr_exp}
Empirically, we show that this algorithm performs well in practice. When applied to data matrices generated from our statistical model, our algorithm yields comparable relative errors vis-\`a-vis several classical NMF algorithms including the multiplicative algorithm, the alternating nonnegative least square algorithm with block pivoting, and the hierarchical alternating least square algorithm. However, our algorithm is {\em significantly faster} as it simply involves calculating rank-one SVDs. It is also  well-known that NMF is sensitive to initializations. The authors in~\cite{Wild04,Bou08} use spherical k-means and an SVD-based technique to initialize NMF. We verify on several image  and hyperspectral  datasets that our algorithm, when combined with several classical NMF algorithms, achieves the best convergence rates and/or the smallest final relative errors. We also provide intuition for why our algorithm serves as an effective initializer for other NMF algorithms. Finally, combinations of our algorithm and several NMF algorithms achieve the best clustering performance for  several  face and text datasets. These experimental results substantiate that our algorithm can be used as a good initializer for standard NMF techniques.

\subsection{Related Work} \label{sec:priorart}
We now describe some works that are related to ours.
%We introduce several highly related problems in Section \ref{sec:priorart}. We present our geometric assumption and introduce useful lemmas in Section \ref{sec:prob_form}. In Section \ref{sec:non-prob}, we present non-probabilistic upper bound for relative error of NMF under our geometric assumption and related theorems. In Section \ref{sec:prob}, we propose a reasonable statistical model for generating data points and achieve tighter relative error upper bound under this model. We demonstrate that a simple technique can be used to automatically determine the optimal latent dimensionality of NMF under our statistical model in Section \ref{sec:num of clusters}. In Section \ref{sec:experiment}, we perform numerical experiments on both synthetic datasets and face datasets to demonstrate that geometric assumption is reasonable for certain real datasets and our algorithm can be used as a good initialization technique for NMF.
%We verify the correctness of our theoretical results through numerical simulations on synthetic datasets. 
\subsubsection{Near-Separable NMF} \label{sec:near-sep}
Arora et al.~\cite{Arora_12} provide an algorithm that runs in time polynomial in $F$, $N$ and $K$ to find the correct factor matrices under the separability condition. Furthermore, the authors consider the near-separable case and prove an approximation error bound when the original data matrix $\bV$ is slightly perturbed from being separable. %The authors in~\cite{Arora_12} provide an algorithm that runs in time polynomial in $F$, $N$ and $K$ under the separability condition. Furthermore, the authors consider the near-separable case and proved that if $\bV$ is normalized such that every row of it has unit $l_{1}$ norm and $\bV$ has a separable nonnegative factorization $\bV=\bW \bH$, in addition, each row of $\bV$ is perturbed by adding a vector of small $l_{1}$ norm to obtain a new data matrix $\bV'$, then with additional assumptions on the noise and $\bH$, their algorithm leads to an approximate nonnegative matrix factorization $\bW'\bH'$ of $\bV'$ with a provable error bound for the $l_{1}$ norm of each row of $\bV'-\bW'\bH'$. 
The algorithm and the theorem for near-separable case is also presented in \cite{Bittorf_12}. The main ideas behind the theorem are as follows: first, $\bV$ must be normalized such that every row of it has unit $\ell_{1}$ norm; this assumption simplifies the conical hull for exact NMF to a convex hull. Second, the rows of $\bH$ need to be robustly simplicial, i.e., every row of $\bH$ should not be contained in the convex hull of all other rows and the largest perturbation of the rows of $\bV$ should be bounded by a function of the smallest distance from a row of $\bH$ to the convex hull of all other rows. Later we will show in Section \ref{sec:prob_form} that our geometric assumption stated in inequality  \eqref{eq:geo_ass} is similar to this key idea in~\cite{Arora_12}. Although we impose a clustering-type generating assumption for data matrix, we do not need the normalization assumption in \cite{Arora_12}, which is stated in \cite{Kumar_13} that may lead to bad clustering performance for text datasets. In addition, because we do not impose this normalization assumption, instead of providing an upper bound on the approximation error, we   provide the upper bound for relative error, which is arguably more natural. 

\subsubsection{Initialization Techniques for NMF}\label{sec:init_NMF}
Similar to k-means, NMF can easily be trapped at bad local optima and is sensitive to initialization. We find that our algorithm is particularly amenable to provide good initial factor matrices for subsequently applying standard NMF algorithms. Thus, here we mention some works on initialization for NMF. Spherical k-means (\texttt{spkm}) is a simple clustering method and it is shown to be one of the most efficient algorithms for document clustering \cite{Dhi01}. The authors in~\cite{Wild04} consider  using \texttt{spkm} for initializing the left factor matrix $\bW$ and observe  a better convergence rate compared to random initialization. Other clustering-based initialization approaches for NMF including divergence-based k-means \cite{Xue08} and fuzzy clustering \cite{Zheng07}. It is also natural to consider using singular value decomposition (SVD) to initialize NMF. In fact, if there is no nonnegativity constraint, we can obtain the best rank-$K$ approximation of a given matrix directly using SVD, and there are strong relations between NMF and SVD. For example, we can obtain the best rank-one  NMF from the best rank-one SVD (see Lemma \ref{lem:rank_one}), and if the best rank-two approximation matrix of a nonnegative data matrix is also nonnegative, then we can also obtain best rank-two NMF \cite{Gon_09}. Moreover, for  a general positive integer $K$, it is shown in \cite{Bou08} that nonnegative double singular value decomposition (\texttt{nndsvd}), a deterministic SVD-based  approach, can be used to enhance the initialization  of NMF, leading to a faster reduction of the approximation error of many NMF algorithms. The CUR decomposition-based initialization method \cite{Langville06} is another factorization-based initialization approach for NMF. We compare our algorithm to widely-used algorithms for initializing NMF  in Section \ref{sec: init_cluster}. %We present that our algorithm can also be used as a good initialization method for NMF with real-dataset numerical experiments.

\subsection{Notations}%\vspace{-.05in}
We use upper case boldface letters to denote matrices and we use lower case boldface letters to denote vectors. We use Matlab-style notation for indexing, e.g., $\bV(i,j)$ denotes the entry of $\bV$ in the $i$-th row and $j$-th column, $\bV(i,:)$ denotes the $i$-th row of $\bV$, $\bV(:,j)$ denotes the $j$-th column of $\bV$ and $\bV(:,\mathscr{K})$ denotes the columns of $\bV$ indexed by $\mathscr{K}$. $\|\bV\|_{\rmF}$ represents the Frobenius norm of $\bV$ and $[N]:=\{1,2,\ldots,N\}$ for any positive integer $N$. Inequalities $\bv\geq 0$ or $\bV \geq 0$ denote element-wise nonnegativity. Let $\bV_{1} \in \mathbb{R}^{F\times N_{1}}$ and $\bV_{2} \in \mathbb{R}^{F\times N_{2}}$, we denote by $\left[\bV_{1}, \bV_{2}\right]$ the horizontal concatenation of the two matrices. Similarly, let $\bV_{1} \in \mathbb{R}^{F_{1}\times N}$ and $\bV_{2} \in \mathbb{R}^{F_{2}\times N}$. We denote by $\left[\bV_{1}; \bV_{2}\right]$ the vertical concatenation of the two matrices. We use  $\bbR_+$ and $\mathbb{R}_{++}$ to represent the set of nonnegative and positive numbers respectively. We denote the nonnegative orthant $\mathbb{R}_{+}^{F}$ as~$\mathcal{P}$. We use $\xrightarrow{\rmp}$ to denote convergence in probability.

\section{Problem Formulation}\label{sec:prob_form}
In this section, we first present our geometric assumption and prove that the exact clustering can be obtained for the normalized data points under the geometric assumption. Next, we introduce several useful lemmas in preparation for the proofs of the main theorems in subsequent sections.
\subsection{Our Geometric Assumption on $\bV$} \label{sec:geom} %\vspace{-.05in}
We assume the columns of $\bV$ lie in $K$ circular cones which satisfy a geometric assumption presented in~\eqref{eq:geo_ass} to follow.  We define {\em circular cones} as follows:
\begin{definition} %\vspace{-.05in}
Given  $\bu \in \mathbb{R}^{F}_{+}$ with unit $\ell_{2}$ norm and an angle $\alpha \in (0, {\pi}/{2})$, the {\em circular cone  with respect to (w.r.t.) $\bu$ and $\alpha$} is defined as
\begin{align}
\mathcal{C}(\bu,\alpha):=\Big\{\bx \in \mathbb{R}^{F} \setminus\{0\}: \frac{\bx^{T}\bu}{\|\bx\|_{2}} \geq \cos\alpha\Big\}.
\end{align}
\end{definition} %
%\vspace{-.05in}
In other words, $\mathcal{C}(\bu,\alpha)$ contains all $\bx \in \mathbb{R}^{F} \setminus\{0\}$ for which the angle between $\bu$ and $\bx$ is not larger than $\alpha$. We say that $\alpha$ and $\bu$ are  respectively the \textit{size angle}  and \textit{basis vector} of the circular cone. In addition, the corresponding truncated circular cone in nonnegative orthant is $\mathcal{C}(\bu,\alpha) \cap \mathcal{P}$.

\begin{figure}
\centering
\includegraphics[width = .875\columnwidth]{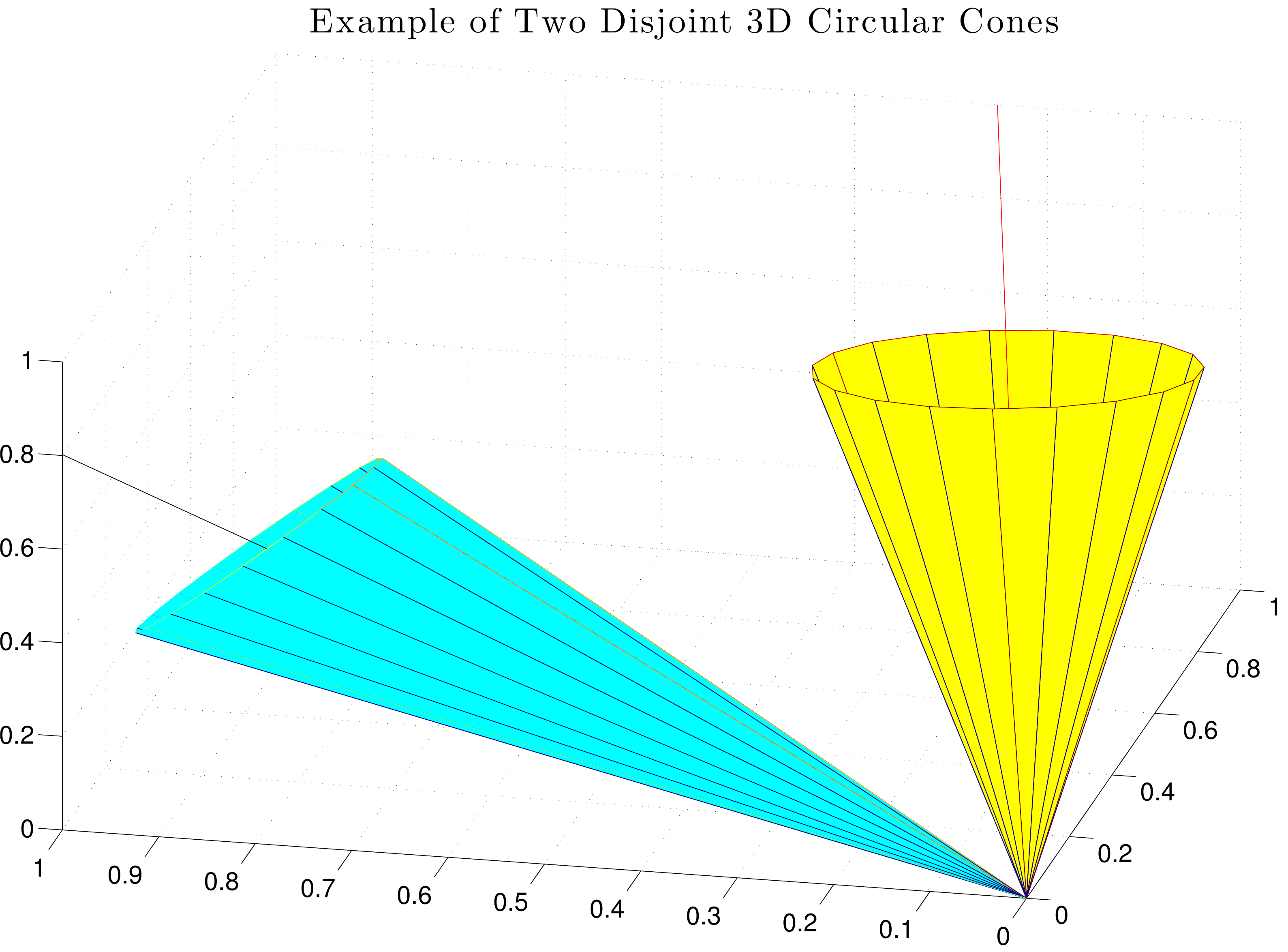}\vspace{-.15in}
\caption{Illustration of the geometric assumption in \eqref{eq:geo_ass}. Here  $\alpha_{1}=\alpha_{2}=0.2$ and $\beta_{12}=0.9>3\alpha_{1}+\alpha_{2}$.}
\label{fig:geo}
\end{figure}

We assume that there are $K$  truncated  circular cones $C_{1}\cap\calP,\ldots,C_{K}\cap\calP$ with corresponding basis vectors and size angles, i.e., $C_{k}:=\mathcal{C}\left(\bu_{k},\alpha_{k}\right)$ for $k \in [K]$. Let $\beta_{ij}:=\arccos \left(\bu_{i}^{T}\bu_{j}\right)$. We make the geometric assumption that  the columns of our data matrix $\bV$ lie in $K$ truncated circular cones which satisfy
\begin{equation} \label{eq:geo_ass}
  \min_{i,j \in [K], i\neq j}\beta_{ij}>\max_{i,j \in [K],i\neq j} \{\max\{\alpha_{i}+3\alpha_{j},3\alpha_{i}+\alpha_{j}\}\}.
\end{equation}
If we sort $\alpha_{1},\ldots,\alpha_{K}$ as $\hat{\alpha}_{1},\ldots,\hat{\alpha}_{K}$ such that $\hat{\alpha}_{1}\geq \hat{\alpha}_{2}\geq \ldots \geq \hat{\alpha}_{K}$, \eqref{eq:geo_ass} is equivalent to 
\begin{equation}
  \min_{i,j \in [K], i\neq j}\beta_{ij}> 3\hat{\alpha}_{1}+\hat{\alpha}_{2}
\end{equation}
The size angle $\alpha_{k}$ is a measure of perturbation in $k$-th circular cone and $\beta_{ij}, i\neq j$ is a measure of distance between the $i$-th basis vector and the $j$-th basis vector. Thus, \eqref{eq:geo_ass} is similar to the second idea in \cite{Arora_12} (cf.\ Section~\ref{sec:near-sep}), namely, that the largest perturbation of the rows of $\bV$ is bounded by a function of the smallest distance from a row of $\bH$ to the convex hull of all other rows. This assumption is realistic for datasets whose samples can be clustered into distinct types; for example, image datasets in which images either contain a distinct foreground (e.g., a face) embedded on a background, or they only comprise a background. See Figure \ref{fig:geo} for an illustration of the geometric assumption in~\eqref{eq:geo_ass} and refer to Figure 1 in \cite{Bittorf_12} for an illustration of the separability condition. 

Now we discuss the relation between our geometric assumption and the separability and near-separability \cite{Arora_12, Bittorf_12} conditions that have appeared in the literature (and discussed in Section~\ref{sec:intro}). Consider a data matrix $\bV$  generated under the extreme case of our geometric assumption that all the size angles of the $K$ circular cones are zero.  Then every column of $\bV$ is a nonnegative multiple of a basis vector of a circular cone. This means that all the columns of $\bV$ can be represented as nonnegative linear combinations of $K$ columns, i.e., the $K$ basis vectors $\bu_1,\ldots,\bu_K$. This can be considered as a special case of separability assumption. When the size angles are not all  zero, our geometric assumption can be considered as a special case of the near-separability assumption.

In Lemma~\ref{lem:correct_cl}, we show that  Algorithm \ref{algo:clustering}, which has time complexity $O(KFN)$, correctly clusters the columns of $\bV$ under the geometric assumption. 

%In this section we first provide several useful lemmas and demonstrate our algorithm for approximate NMF under our geometric assumption. Subsequently, we present our main theorems that present  upper bounds on the relative errors of the NMF approximation.
\begin{lemma}
\label{lem:correct_cl}
Under the geometric assumption on $\bV$, if Algorithm~\ref{algo:clustering} is applied to $\bV$, then the columns of $\bV$ are partitioned  into  $K$ subsets, such that the data points in the same subset are generated from the same  truncated  circular cone.   
\end{lemma}
\begin{IEEEproof}
We normalize $\bV$ to obtain $\bV'$, such that all the columns of $\bV'$ have unit $\ell_{2}$ norm. From the definition, we know if a data point is in a truncated circular cone, then the normalized data point is also in the truncated circular cone. Then for any two columns $\bx$, $\by$ of $\bV'$ that are in the same truncated circular cone $C_{k}\cap \mathcal{P}, k\in [K]$, the largest possible angle between them is $\min\{2\alpha_{k},\pi/2\}$, and thus the distance $\|\bx-\by\|_{2}$ between these two data points is not larger than $\sqrt{2\left(1-\cos \left(2\alpha_{k}\right)\right)}$. On the other hand, for any two columns $\bx$, $\by$ of $\bV'$ that are in two truncated circular cones $C_{i}\cap \mathcal{P}, C_{j}\cap \mathcal{P}, i\neq j$,   the smallest possible angle between them is $\beta_{ij}-\alpha_{i}-\alpha_{j}$, thus the smallest possible distance between them is $\sqrt{2\left(1-\cos\left(\beta_{ij}-\alpha_{i}-\alpha_{j}\right)\right)}$. Then under the geometric assumption \eqref{eq:geo_ass}, the distance between any two unit data points in distinct truncated circular cones is larger than the distance between any two unit data points in the same truncated circular cone.  Hence, Algorithm~\ref{algo:clustering} returns the correct clusters.  
%To obtain the correct $K$-clustering for $\bV'$, we can apply Algorithm~\ref{algo:clustering}. 
\end{IEEEproof}

%\iffalse
\begin{algorithm}[t]
\caption{Greedy clustering method with geometric assumption in~\eqref{eq:geo_ass}}%\label{algo:clustering}
\label{algo:clustering}
\begin{algorithmic} 
\State {\bf Input}: Data matrix $\bV \in \mathbb{R}_{+}^{F\times N}$, $K \in \mathbb{N}$
\State {\bf Output}: A set of non-empty, pairwise disjoint index sets $\mathscr{I}_{1},\mathscr{I}_{2},\ldots,\mathscr{I}_{K} \subseteq [N]$ such that their union is $[N]$
%\State {\bf for} $t$ = 1 to $N$ {\bf do}
\State 1) Normalize $\bV$ to obtain $\bV'$, such that all the columns of $\bV'$ have unit $\ell_{2}$ norm.
\State 2) Arbitrarily pick a point $\bz_1 \in \bV'$ (i.e., $\bz_{1}$ is a column in $\bV'$) as the first centroid.
\State 3) {\bf for} $k$ = 1 to $K-1$ {\bf do}
\begin{equation}
  %\bz_{k+1}:=\argmax_{\bz\in \bV'}\{\min\{\|\bz-\bz_{i}\|,i\in \left[k\right]\}\},
  \bz_{k+1}:=\mathrm{argmin}_{\bz\in \bV'}\{\max\{\bz_{i}^{T}\bz,i\in [k]\}\}
\end{equation}
and set $\bz_{k+1}$ be the $(k+1)$-st centroid.
\State 4) $\mathscr{I}_{k}:=\{n\in [N]: k = \argmax_{j\in[K]}\bz_{j}^{T}\bV'(:,n)\}$ for all  $k\in [K]$.
\end{algorithmic}
\end{algorithm} 

Now we present the following two useful lemmas. Lemma \ref{lem:svd_pert} provides an upper bound for perturbations of singular values. Lemma \ref{lem:rank_one} shows that we can directly obtain the best rank-one nonnegative matrix factorization from the best rank-one SVD.

\begin{lemma}[Perturbation of singular values~\cite{Golub_89}]
\label{lem:svd_pert}
If $\bA$ and $\bA+\bE$ are in $\mathbb{R}^{F\times N}$,  then
\begin{equation}
  \sum_{p=1}^{P}\left(\sigma_{p}(\bA+\bE)-\sigma_{p}(\bA)\right)^{2} \leq \|\bE\|_{\rmF}^{2},
\end{equation}
where $P=\min\{F,N\}$ and $\sigma_{p}(\bA)$ is the $p$-th largest singular value of $\bA$. In addition, we have 
\begin{equation}
  |\sigma_{p}(\bA+\bE)-\sigma_{p}(\bA)| \leq \sigma_{1}(\bE)=\|\bE\|_{2}
\end{equation}
for any $p\in [P]$.
\end{lemma}

\begin{lemma}[Rank-One Approximate NMF~\cite{Gon_09}]
\label{lem:rank_one}
Let $\sigma \bu \bv^{T}$ be the rank-one singular value decomposition of a matrix $\bV \in \mathbb{R}_{+}^{F\times N}$. Then $\bu':=\sigma |\bu|$, $\bv':=|\bv|$ solves 
\begin{equation}
  \min_{\bx \in \mathbb{R}_{+}^{F},\by \in \mathbb{R}_{+}^{N}} \|\bV-\bx\by^{T}\|_{\rmF}.
\end{equation}
\end{lemma}

\section{Non-Probabilistic Theorems}\label{sec:non-prob}   
In this section, we first present a  deterministic theorem concerning an upper bound for the relative error of NMF. Subsequently, we provide several extensions of this  theorem. 
\begin{theorem}\label{thm:main}
 Suppose all the data points in data matrix $\bV\in \mathbb{R}^{F\times N}_{+}$ are drawn from $K$ truncated circular cones $C_{1}\cap \mathcal{P},\ldots,C_{K}\cap \mathcal{P}$, where $C_{k}:=\mathcal{C}\left(\bu_{k},\alpha_{k}\right)$ for $k \in [K]$. Then there is a pair of factor matrices $\bW^{*}\in \mathbb{R}^{F\times K}_{+}$, $\bH^{*}\in \mathbb{R}^{K\times N}_{+}$, such that 
 \begin{equation}
\frac{\|\bV-\bW^{*}\bH^{*}\|_{\rmF}}{\|\bV\|_{\rmF}} \leq \max_{k\in [K]}\{\sin\alpha_{k}\}. \label{eqn:bd_non_prob0}
\end{equation}
\end{theorem}
\begin{IEEEproof}
Define $\mathscr{I}_{k}:=\{n\in [N]: \bv_{n}\in C_{k}\cap \mathcal{P}\}$ (if a data point $\bv_{n}$ is contained in more than one truncated circular cones, we arbitrarily assign any one it is contained in). Then $\mathscr{I}_{1},\mathscr{I}_{2},\ldots,\mathscr{I}_{K} \subseteq [N]$ are disjoint index  sets and their union is $[N]$. Any two data points $\bV\left(:,j_{1}\right)$ and $\bV\left(:,j_{2}\right)$ are in the same circular cones if $j_{1}$ and $j_{2}$ are in the same index set. Let $\bV_{k}=\bV\left(:,\mathscr{I}_{k}\right)$ and without loss of generality, suppose that $\bV_{k}\in C_{k}$ for $k\in [K]$. For any $k\in [K]$ and any column $\bz$ of $\bV_{k}$, suppose the angle between $\bz$ and $\bu_{k}$ is $\beta$, we have $\beta \leq \alpha_{k}$ and $\bz=\|\bz\|_{2} (\cos\beta) \bu_{k}+\by$, with $\|\by\|_{2}=\|\bz\|_{2}(\sin\beta )\leq \|\bz\|_{2}(\sin\alpha_{k})$. Thus $\bV_{k}$ can be written as the sum of a rank-one matrix $\bA_{k}$ and a perturbation matrix $\bE_{k}$. By Lemma \ref{lem:rank_one}, we can find the best rank-one approximate NMF of $\bV_{k}$ from the singular value decomposition of $\bV_{k}$. Suppose $\bw_{k}^{*} \in \mathbb{R}_{+}^{F}$ and $\bh_{k} \in \mathbb{R}_{+}^{|\mathscr{I}_{k}|}$ solve  the best rank-one approximate NMF. Let $\bS_{k}:=\bw_{k}^{*}\bh_{k}^{T}$ be the best rank-one approximation matrix of $\bV_{k}$. Let $P_{k}=\min\{F,|\mathscr{I}_{k}|\}$, then by Lemma \ref{lem:svd_pert}, we have  
\begin{equation} 
  \|\bV_{k}-\bS_{k}\|_{\rmF}^{2}=\sum_{p=2}^{P_{k}} \sigma_{p}^{2}\left(\bV_{k}\right)=\sum_{p=2}^{P_{k}} \sigma_{p}^{2}\left(\bA_{k}+\bE_{k}\right) \leq \|\bE_{k}\|_{\rmF}^{2}.  \label{eq:main_ineq0}
\end{equation}
From the previous result, we know that
\begin{equation}  
  \frac{\|\bE_{k}\|_{\rmF}^{2}}{\|\bV_{k}\|_{\rmF}^{2}} = \frac{\sum_{\bz \in \bV_{k}} \|\bz\|_{2}^{2}\sin^{2}\beta_{\bz}}{\sum_{\bz \in \bV_{k}} \|\bz\|_{2}^{2}} \leq \sin^{2}\alpha_{k},\label{eq:main_ineq}
\end{equation}
where $\beta_{\bz}$ denotes the angle between $\bz$ and $\bu_{k}$, $\beta_{\bz} \leq \alpha_{k}$, and $\bz\in \bV_{k}$ runs over all columns of the matrix $\bV_k$.

Define $\bh_{k}^{*} \in \mathbb{R}^{N}_{+}$ as $\bh_{k}^{*}(n)=\bh_{k}(n)$, if $n\in \mathscr{I}_{k}$ and $\bh_{k}^{*}(n)=0$ if $n \notin \mathscr{I}_{k}$. Let $\bW^{*}:=\big[\bw_{1}^{*},\bw_{2}^{*},\ldots,\bw_{K}^{*}\big]$ and $\bH^{*}:=\big[\left(\bh_{1}^{*}\right)^{T};\left(\bh_{2}^{*}\right)^{T}\ldots;\left(\bh_{K}^{*}\right)^{T}\big]$, then we have 
\begin{align}   \frac{\|\bV-\bW^{*}\bH^{*}\|_{\rmF}^{2}}{\|\bV\|_{\rmF}^{2}}=\frac{\sum_{k=1}^{K} \|\bV_{k}-\bw^{*}_{k}\bh_{k}^{T}\|_{\rmF}^{2}}{\|\bV\|_{\rmF}^{2}} & \\
  \leq \frac{\sum_{k=1}^{K} \|\bV_{k}\|_{\rmF}^{2}\sin^{2}\alpha_{k}}{\sum_{k=1}^{K} \|\bV_{k}\|_{\rmF}^{2}}. &
\end{align} 
Thus we have \eqref{eqn:bd_non_prob0} as desired.  
\end{IEEEproof}

  In practice, to obtain the tightest possible upper bound for~\eqref{eqn:bd_non_prob0}, we need to solve the following optimization problem: 
\begin{equation}
\min \max_{k\in [K]} \alpha  (\bV_{k} ),
\end{equation}
where $\alpha  (\bV_{k}  )$ represents the smallest possible size angle corresponding to $\bV_{k}$  (defined in \eqref{eqn:defVk}) and   the minimization is taken over all possible clusterings of  the columns of $\bV$. We consider finding an optimal size angle and a corresponding basis vector for any data matrix, which we hereby write as $\bX:=\left[\bx_{1},\ldots,\bx_{M}\right] \in \mathbb{R}_{+}^{F\times M}$ where  $M \in \mathbb{N}_{+}$.  
This is solved by the following optimization problem:
\begin{align}
\text{minimize}\qquad & \alpha \nn
%\label{green} 
\\
\text{subject to}\qquad & \bx_{m}^{T}\bu \geq \cos \alpha, \quad m\in [M],   
%\label{green-constraint-1} 
\\
& \bu\geq 0, \quad \|\bu\|_{2}=1, \quad \alpha\geq 0,\nn
%\label{green-constraint-2}
%\label{green-constraint-3}
\end{align}
where the decision variables are $(\alpha,\bu)$. 
Alternatively,   consider
\begin{align}
\text{maximize}\qquad & \cos \alpha \nn
%\label{green} 
\\
\text{subject to}\qquad & \bx_{m}^{T}\bu \geq \cos \alpha, \quad m\in [M],      \label{eqn:svm2}
%\label{green-constraint-1} 
\\
& \bu\geq 0, \quad \|\bu\|_{2}=1.\nn
%\label{green-constraint-2}
%\label{green-constraint-3}
\end{align}
Similar to the primal optimization problem for linearly separable support vector machines~\cite{Boser_92}, we can obtain the optimal $\bu$ and $\alpha$ for \eqref{eqn:svm2} by solving  
\begin{align}
\text{minimize}\qquad & \frac{1}{2}\|\bu\|_{2}^{2} \nn
%\label{green} 
\\
\text{subject to}\qquad & \bx_{m}^{T}\bu \geq 1 , \quad m\in [M],  \quad \bu\geq 0,   \label{eqn:svm3}
%\label{green-constraint-1} 
%\\
%& \bu\geq 0 \nn
%\label{green-constraint-2}
%\label{green-constraint-3}
\end{align}
where the decision variable here is only $\bu$. Note that  \eqref{eqn:svm3}
 is a quadratic programming problem and can be easily solved by standard convex optimization software. Suppose $\hat{\bu}$ is the optimal solution of \eqref{eqn:svm3}, then $\bu^{*}:=\hat{\bu}/\|\hat{\bu}\|_{2}$ and $\alpha^{*}:=\arccos \left(1/\|\hat{\bu}\|_{2}\right)$ is the optimal basis vector and size angle. 

We now state and prove a relative error bound of the proposed approximate NMF algorithm detailed in Algorithm~\ref{algo:approx} under our geometric assumption. We see that if the size angles of all circular cones are small compared to the angle between the basis vectors of any two circular cones, then exact clustering is possible, and thus the relative error of the best approximate NMF of an arbitrary nonnegative matrix generated from these circular cones can be appropriately controlled by these size angles. Note that rank-one SVD can be implemented by the power method efficiently~\cite{Golub_89}. Recall that as mentioned in Section~\ref{sec:geom}, Theorem \ref{thm:non-prob} is similar to the corresponding theorem for the near-separable case in \cite{Arora_12} in terms of  the geometric condition imposed.
\begin{theorem} \label{thm:non-prob}
  Under the geometric assumption given in Section \ref{sec:geom} for generating $\bV\in \mathbb{R}^{F\times N}_{+}$, Algorithm~\ref{algo:approx} outputs $\bW^{*}\in \mathbb{R}^{F\times K}_{+}$, $\bH^{*}\in \mathbb{R}^{K\times N}_{+}$, such that 
\begin{equation}
\frac{\|\bV-\bW^{*}\bH^{*}\|_{\rmF}}{\|\bV\|_{\rmF}} \leq \max_{k\in [K]}\{\sin\alpha_{k}\}. \label{eqn:bd_non_prob}
\end{equation}
\end{theorem}
\begin{IEEEproof}
From Lemma \ref{lem:correct_cl}, under the geometric assumption in Section \ref{sec:geom}, we can obtain a set of non-empty, pairwise disjoint index sets $\mathscr{I}_{1},\mathscr{I}_{2},\ldots,\mathscr{I}_{K} \subseteq [N]$ such that their union is $[N]$ and two data points $\bV\left(:,j_{1}\right)$ and $\bV\left(:,j_{2}\right)$ are in the same circular cones if and only if $j_{1}$ and $j_{2}$ are in the same index set. Then from Theorem \ref{thm:main}, we can obtain $\bW^{*}$ and $\bH^{*}$ with the same upper bound on the relative error.  
\end{IEEEproof}

\begin{algorithm}[t]
\caption{Clustering and Rank One NMF (\texttt{cr1-nmf})}\label{algo:approx}
\begin{algorithmic} 
\State {\bf Input}: Data matrix $\bV \in \mathbb{R}_{+}^{F\times N}$, $K \in \mathbb{N}$
\State {\bf Output}: Factor matrices $\bW^{*}\in \mathbb{R}_{+}^{F\times K}$, $\bH^{*}\in \mathbb{R}_{+}^{K\times N}$
%\State {\bf for} $t$ = 1 to $N$ {\bf do}
\State 1) Use Algorithm~\ref{algo:clustering} to find a set of non-empty, pairwise disjoint index sets $\mathscr{I}_{1},\mathscr{I}_{2},\ldots,\mathscr{I}_{K} \subseteq  [N]$.
\State 2) {\bf for} $k$ = 1 to $K$ {\bf do}
\begin{align}
  & \bV_{k} := \bV\left(:,\mathscr{I}_{k}\right) ;  \label{eqn:defVk}\\
  & \left[\bU_{k},\bSigma_{k},\bX_{k}\right]  := \mathrm{svd}\left(\bV_{k}\right); \\
  &  \bw_{k}^{*}  :=|\bU_{k}(:,1)|, \quad \bh_{k}:=\bSigma_{k}(1,1)|\bX_{k}(:,1)|;  \\
  &   \bh_{k}^{*} :=\mathrm{zeros}(1,N), \bh_{k}^{*}\left(\mathscr{I}_{k}\right)=\bh_{k}. 
\end{align}
\State 3) $\bW^{*}:=\big[\bw_{1}^{*},\ldots,\bw_{K}^{*}\big]$, $\bH^{*}:=\big[\left(\bh_{1}^{*}\right)^{T};\ldots;\left(\bh_{K}^{*}\right)^{T}\big]$.
\end{algorithmic}
\end{algorithm}

\section{Probabilistic Theorems}\label{sec:prob}  
We now provide a tighter relative error bound by assuming a probabilistic model. For simplicity, we assume a straightforward and easy-to-implement statistical model for the sampling procedure. We first present the proof of the tighter relative error bound corresponding to the probabilistic model in Theorem \ref{thm:prob} to follow, then we show that the upper bound for relative error is tight if we assume all the circular cones are contained in nonnegative orthant in Theorem \ref{thm:prob_ext}. 

We assume the following generating process for each column $\bv$ of $\bV$ in Theorem~\ref{thm:prob} to follow.
\begin{enumerate}
 \item sample $k\in [K]$ with equal probability $1/K$;
 \item sample the squared length $l$ from the exponential distribution\footnote{$\mathrm{Exp} (\lambda)$ is the function $x\mapsto \lambda\exp(- \lambda x)1\{x\ge 0\}$.} $\mathrm{Exp} (\lambda_k)$  with parameter (inverse of the expectation)  $\lambda_{k}$;
 \item uniformly sample a unit vector $\bz \in C_{k}$ w.r.t.\ the angle between $\bz$ and $\bu_{k}$;\footnote{This means we first uniformly sample an angle $\beta \in [0,\alpha_k]$ and subsequently uniformly sample a vector $\bz$ from the set $\{\bx \in \mathbb{R}^{F}: \|\bx\|_2=1, \bx^T \bu_k = \cos \beta\}$}
 \item if $\bz \notin \mathbb{R}^{F}_{+}$, set all negative entries of $\bz$ to zero, and rescale $\bz$ to be a unit vector;
 \item let $\bv=\sqrt{l}\bz$;
\end{enumerate}

\begin{theorem} 
\label{thm:prob}
  Suppose the $K$  truncated  circular cones $C_{k}\cap \mathcal{P}$ with $C_{k}:=\mathcal{C}(\bu_{k},\alpha_{k}) \in \mathbb{R}^{F}$  for $k \in [K]$ satisfy the geometric assumption given by \eqref{eq:geo_ass}. Let $\blambda:=(\lambda_{1};\lambda_{2};\ldots;\lambda_{K}) \in \mathbb{R}_{++}^{K}$. We generate the columns of a data matrix $\bV \in \mathbb{R}^{F\times N}_{+}$ from the above generative process. Let $f(\alpha):=\frac{1}{2}- \frac{\sin 2\alpha}{4\alpha}$, then for a small $\epsilon >0$, with probability at least $1-8\exp(-\xi N\epsilon^{2})$, one has 
\begin{equation}
\frac{\|\bV-\bW^{*}\bH^{*}\|_{\rmF}}{\|\bV\|_{\rmF}} \leq \sqrt{\frac{\sum_{k=1}^{K}f(\alpha_{k})/\lambda_{k}}{\sum_{k=1}^{K}1/\lambda_{k}}}+\epsilon , \label{eqn:prob1}
\end{equation}
where the constant $\xi>0$ depends only on $\lambda_{k}$ and $f(\alpha_{k})$ for all $k \in [K]$.
\end{theorem}
\begin{remark}
 The assumption in Step 1 in the generating process that the data points are generated from $K$ circular cones with equal probability can be easily generalized to unequal probabilities. The assumption in Step 2 that the square of the length of a data point is sampled from an exponential distribution can be easily extended  any nonnegative sub-exponential distribution (cf.~Definition \ref{def:sub-exp} below), or equivalently, the length of a data point is sampled from a nonnegative sub-gaussian distribution (cf.~Definition \ref{def:sub-gauss} in Appendix \ref{app:prf_prob}).
\end{remark}

The relative error bound produced by Theorem~\ref{thm:prob} is better than that of Theorem \ref{thm:non-prob}, i.e., the former is more conservative. This can be seen from~\eqref{eq:int} to follow, or from the inequality $\alpha \leq \tan \alpha$ for  $\alpha \in [0,\pi/2)$. We also observe this in the experiments in Section \ref{sec:syn0}.

Before proving Theorem \ref{thm:prob}, we define  sub-exponential random variables and present a useful lemma.

\begin{definition}\label{def:sub-exp}
A {\em sub-exponential random variable} $X$ is one that satisfies one of the following equivalent properties \\
1. Tails: $\mathbb{P}(|X|>t)\leq \exp(1-t/K_{1})$ for all $t\geq 0$;\\
2. Moments: $\left(\mathbb{E}|X|^{p}\right)^{1/p} \leq K_{2} p$ for all $p\geq 1$;\\
3. $\mathbb{E}\left[\exp(X/K_{3})\right]\leq e$;\\
where $K_{i}, i=1,2,3$ are positive constants. The {\em sub-exponential norm} of $X$, denoted $\|X\|_{\Psi_{1}}$, is defined to be
\begin{equation}
  \|X\|_{\Psi_{1}}:=\sup_{p\geq 1} p^{-1}\left(\mathbb{E}|X|^{p}\right)^{1/p}.
\end{equation} 
\end{definition}

\begin{lemma} (Bernstein-type inequality)\cite{Ver_10} 
\label{lem:large_dev}
Let $X_{1}, \ldots , X_{N}$ be independent sub-exponential random variables with zero expectations, and $M = \max_{i} \|X_{i} \|_{\Psi_{1}}$. Then for every $\epsilon \geq 0$, we have
\begin{equation}
\!\!\mathbb{P}\bigg( \Big|\sum_{i=1}^{N}X_{i}\Big|\!\geq\! \epsilon N\bigg)\! \leq \! 2  \exp \left[-c \cdot \mathrm{min}\left(\frac{\epsilon^{2}}{M^{2}},\frac{\epsilon}{M}\right)N\right], \label{eqn:subexp}
\end{equation}
where $c>0$ is an absolute constant.
\end{lemma}

Theorem~\ref{thm:prob} is proved by combining the large deviation bound in Lemma \ref{lem:large_dev} with the deterministic bound on the relative error in Theorem~\ref{thm:non-prob}.%, we can  prove of. 

\begin{IEEEproof}[Proof of Theorem \ref{thm:prob}]
%\noindent \textbf{of Theorem \ref{thm:prob}}. 
From \eqref{eq:main_ineq0} and \eqref{eq:main_ineq} in the proof of Theorem \ref{thm:non-prob}, to obtain an upper bound for the square of the relative error, we   consider the following random variable
\begin{equation}
  D_{N}:=\dfrac{\sum_{n=1}^{N}L_{n}^{2}\sin^{2}B_{n}}{\sum_{n=1}^{N}L_{n}^{2}},
\end{equation}
where $L_{n}$ is the random variable corresponding to the length of the $n$-th point, and $B_{n}$ is the random variable corresponding to the angle between the $n$-th point and $\bu_{k}$ for some $k \in [K]$ such that the point is in  $C_{k} \cap \mathcal{P}$. We first consider estimating the above random variable with the assumption that all the data points are generated from a single truncated circular cone $C \cap \mathcal{P}$ with $C:=\mathcal{C}(\bu,\alpha)$ (i.e., assume $K=1$), and the square of lengths are generated according to the exponential distribution $\mathrm{Exp}(\lambda)$. Because we assume each angle $\beta_{n} $ for $n\in [N]$ is  sampled from a uniform distribution on $[0,\alpha]$, the expectation of $\sin^{2}B_{n}$ is
\begin{equation}
\label{eq:int}
  \mathbb{E}\left[\sin^{2}B_{n}\right]=\int_{0}^{\alpha} \frac{1}{\alpha}\sin^{2}\beta \,\mathrm{d}\beta=\frac{1}{2}-\frac{\sin 2\alpha}{4\alpha}=f(\alpha).
\end{equation}
Here we only need to consider vectors $\bz \in \bbR_+^F$ whose angles with $\bu$ are not larger than $\alpha$. Otherwise, we have $\mathbb{E} [\sin^{2}B_{n} ]\leq f(\alpha)$. Our probabilistic upper bound also holds in this case. 

Since the length and the angle are independent, we have 
\begin{equation}
  \mathbb{E}\left[D_{N}\right]=\mathbb{E}\left[\mathbb{E}\left[D_{N}|L_{1},\ldots,L_{N}\right]\right]=f(\alpha),
\end{equation}
and we also have
\begin{equation}
  \mathbb{E}\left[L_{n}^{2}\sin^{2}B_{n}\right]=\mathbb{E}\left[L_{n}^{2}\right]\mathbb{E}\left[\sin^{2}B_{n}\right]=\frac{f(\alpha)}{\lambda}.
\end{equation}
Define $X_{n}:=L_{n}^{2}$ for all $n\in [N]$. Let 
\begin{equation}
H_{N}:= \frac{\sum_{n=1}^{N}X_{n}}{N},\;\;\mbox{and}\;\; G_{N}:=\frac{\sum_{n=1}^{N}X_{n}\sin^{2}B_{n}}{N}.
\end{equation}
 We have for all $n \in [N]$,
 \begin{equation}
   \mathbb{E}[X_{n}^{p}]=\lambda^{-p}\Gamma(p+1) \leq \lambda^{-p}p^{p}, \qquad \forall \, p\geq 1,
 \end{equation}
where $\Gamma(\cdot)$ is the gamma function. Thus $\|X_{n}\|_{\Psi_{1}} \leq \lambda^{-1}$, and $X_{n}$ is sub-exponential. By  the triangle inequality, we have $\|X_{n}-\mathbb{E}X_{n}\|_{\Psi_{1}} \leq \|X_{n}\|_{\Psi_{1}}+\|\mathbb{E}X_{n}\|_{\Psi_{1}} \leq 2\|X_{n}\|_{\Psi_{1}}$. Hence, by Lemma \ref{lem:large_dev}, for all $\epsilon>0$, we have \eqref{eqn:subexp} 
where $M$ can be taken as $M=2/\lambda$. Because
\begin{equation}
  \left(\mathbb{E}\big[\big(X_{n}\sin^{2} B_{n}\right)^{p}\big]\big)^{1/p} \leq \lambda^{-1}p \sin^{2} \alpha \leq \lambda^{-1}p,
\end{equation}
we have a similar large deviation result for $G_{N}$. 

On the other hand, for all $\epsilon>0$
\begin{align} 
&\mathbb{P}\left(|D_{N}-f(\alpha)|\geq \epsilon\right)  =\mathbb{P}\left( \Big|\frac{G_{N}}{H_{N}}-f(\alpha)\Big|\geq \epsilon\right)\\
&\le   \mathbb{P}\left(|\lambda G_{N}\!-\! f(\alpha)|\!\geq\! \frac{\epsilon}{2}\right)\!+\!\mathbb{P}\left( \Big|\frac{G_{N}}{H_{N}}\!-\!\lambda G_{N}\Big|\!\geq\! \frac{\epsilon}{2}\right).
\end{align}
For the second term, by fixing small constants $\delta_{1},\delta_{2}>0$, we have 
\begin{align}
&\mathbb{P}\left( \Big|\frac{G_{N}}{H_{N}}-\lambda G_{N}\Big|\geq \frac{\epsilon}{2}\right)=\mathbb{P}\left(\frac{|1-\lambda H_{N}|G_{N}}{H_{N}}\geq \frac{\epsilon}{2}\right) \label{eq:typo1}\\
&\leq \mathbb{P}\left(\frac{|1\! -\!\lambda H_{N}|G_{N}}{H_{N}}\geq \frac{\epsilon}{2}, H_{N}\geq \frac{1}{\lambda}-\delta_{1},G_{N}\! \leq\! \frac{f(\alpha)}{\lambda}\!+\!\delta_{2}\right)\nonumber\\
&\qquad +\mathbb{P}\left(H_{N} < \frac{1}{\lambda}-\delta_{1}\right)+\mathbb{P}\left(G_{N} > \frac{f(\alpha)}{\lambda}+\delta_{2}\right).
\end{align}
Combining the large deviation bounds for $H_{N}$ and $G_{N}$ in \eqref{eqn:subexp} with the above inequalities, if we set $\delta_{1}=\delta_{2}=\epsilon$ and take $\epsilon$ sufficiently small, 
\begin{equation} \label{ineq:final}
  \mathbb{P}\left(|D_{N}-f(\alpha)|\geq \epsilon\right) \leq 8\exp\left(-\xi  N\epsilon^{2}\right),
\end{equation}
where $\xi$ is a positive constant depending on $\lambda$ and $f(\alpha)$.

Now we turn to the  general  case in which $K \in \mathbb{N}$. We have 
\begin{align}
  \mathbb{E}\left[X_{n}\right]&=\frac{\sum_{k=1}^{K}1/\lambda_{k}}{K}, \quad\mbox{and}\\ 
  \mathbb{E}\left[X_{n}\sin^{2}B_{n}\right]&=\frac{\sum_{k=1}^{K}f(\alpha_{k})/\lambda_{k}}{K}, 
\end{align}
and for all $p\ge 1$, 
\begin{equation}
  \left(\mathbb{E}  [X_{n}^{p} ]\right)^{1/p}=\left(\frac{\sum_{k=1}^{K}\lambda_{k}^{-p}\Gamma(p+1)}{K}\right)^{1/p}\leq \frac{p}{\min_{k} \lambda_{k}}.
\end{equation}
Similar to \eqref{ineq:final}, we have 
\begin{equation}
  \mathbb{P}\left( \bigg|D_{N}\!-\!\frac{\sum_{k=1}^{K} f(\alpha_{k} /\lambda_{k})}{\sum_{k=1}^{K}1/\lambda_{k}}\bigg|\!\geq\! \epsilon\right) \leq 8\exp\left(-\xi N\epsilon^{2}\right), 
\end{equation}
and thus, if we let $\Delta:=\sqrt{\frac{\sum_{k=1}^{K}f(\alpha_{k})/\lambda_{k}}{\sum_{k=1}^{K}1/\lambda_{k}}}$, we have
\begin{align} 
  \mathbb{P}\left(\big|\sqrt{D_{N}}-\Delta\big|\leq \epsilon\right)&\geq \mathbb{P}\left(\big|D_{N}-\Delta^{2}\big|\leq \Delta\epsilon\right)\\
  & \geq 1-8\exp\left(-\xi N\Delta^{2}\epsilon^{2}\right).   
\end{align}
This completes the proof of \eqref{eqn:prob1}.  
%\end{IEEEproof} 
\end{IEEEproof}
%\vspace{-.1in}

Furthermore, if the $K$ circular cones $\mathcal{C}_{1},\ldots,\mathcal{C}_{K}$ are contained in the nonnegative orthant $\mathcal{P}$, we do not need to project the data points not in $\mathcal{P}$ onto $\mathcal{P}$. Then we can prove that the upper bound in Theorem \ref{thm:prob} is  asymptotically tight, i.e., 
\begin{equation}
  \frac{\|\textbf{V}-\textbf{W}^{*}\textbf{H}^{*}\|_{\rmF}}{\|\textbf{V}\|_{\rmF}} \xrightarrow{\rmp} \sqrt{\frac{\sum_{k=1}^{K}f(\alpha_{k})/\lambda_{k}}{\sum_{k=1}^{K}1/\lambda_{k}}},\; \mbox{as } N\to\infty.
\end{equation} 
\begin{theorem}\label{thm:prob_ext}
  Suppose the data points of $\bV \in \mathbb{R}^{F\times N}_{+}$ are generated as given in Theorem \ref{thm:prob} with all the circular cones being contained in the nonnegtive orthant, then  Algorithm~\ref{algo:approx} produces $\bW^{*}\in\mathbb{R}^{F\times K}_{+}$ and $\bH^{*} \in \mathbb{R}^{K\times N}_{+}$ with the property that  for any $\epsilon \in (0,1)$ and $t\geq 1$, if $N\geq c(t/\epsilon)^{2}F$, then with probability at least $1-6K\exp(-t^{2}F)$ one has 
  \begin{equation}
   \left|\frac{\|\bV-\bW^{*}\bH^{*}\|_{\rmF}}{\|\bV\|_{\rmF}}-\sqrt{\frac{\sum_{k=1}^{K}f(\alpha_{k})/\lambda_{k}}{\sum_{k=1}^{K}1/\lambda_{k}}}\right| \leq c\epsilon \label{eqn:convergence_prob}
  \end{equation}
  where $c$ is a   constant depending on $K$ and $\alpha_{k}$, $\lambda_{k}$ for $k\in[K]$.
\end{theorem}
\begin{IEEEproof}
 Since the proof of Theorem \ref{thm:prob_ext} is somewhat similar to that of Theorem \ref{thm:prob}, we defer it to Appendix \ref{app:prf_prob}.
\end{IEEEproof}

\section{Automatically Determining $K$}\label{sec:num of clusters}

Automatically determining the latent dimensionality $K$  is an important problem in NMF. Unfortunately, the usual and popular approach for determining the latent dimensionality of nonnegative data matrices based on Bayesian automatic relevance determination by Tan and F\'evotte~\cite{Tan13} does not  work well for data matrices generated under the geometric assumption given in Section \ref{sec:geom}.  This is because in \cite{Tan13}, $\bW$ and $\bH$  are assumed to be generated from the {\em same} distribution. Under the geometric assumption, $\bV$ has well clustered columns and the corresponding coefficient matrix $\bH$ can be approximated by a clustering membership indicator matrix with columns that are $1$-sparse (i.e., only contains one non-zero entry). Thus, $\bW$ and $\bH$ have very different statistics. While there are many approaches \cite{Rou87,Moo00,Has01} to learn the number of clusters in clustering problems, most methods lack strong theoretical guarantees.

By assuming the generative procedure for $\bV$ proposed in Theorem~\ref{thm:prob}, we consider a simple approach for determining $K$ based on the maximum of the ratios between adjacent singular values. We provide a theoretical result for the correctness of this approach. Our method consists in estimating the correct number of circular cones  $\hatK$  as follows:
\begin{equation}\label{eq:estK}
\hatK :=\argmax_{k\in\left \{ K_{\mathrm{min}}, \ldots, K_{\mathrm{max}}\right \}} \frac{\sigma_{k}(\bV)}{\sigma_{k+1}(\bV)}.
\end{equation}
Here $K_{\mathrm{min}}>1$ and $K_{\mathrm{max}} <\mathrm{rank}(\bV)$ are selected based on domain knowledge. 
 The main ideas that underpin~\eqref{eq:estK} are (i) the approximation error for the best rank-$k$ approximation of a data matrix in the Frobenius norm and (ii) the so-called elbow method~\cite{Thorndike53whobelongs} for determining the number of clusters. More precisely, let $\bV_{k}$ be the best rank-$k$ approximation of $\bV$. Then  $\|\bV-\bV_{k}\|_{\rmF}^{2}=\sum_{j=k+1}^{r} \sigma_{j}^{2}(\bV)$, where $r$ is the rank of $\bV$. If we increase $k$ to $k+1$, the square of the best approximation error decreases by $\sigma_{k+1}^{2}(\bV)$. The elbow method chooses a number of clusters $k$ so that the decrease in the objective function value from $k$ clusters to $k+1$ clusters is small compared to the decrease in the objective function value from $k-1$ clusters to $k$ clusters. %This helps us to consider maximizing the ratio $\sigma_{k}^{2}(\bV)/\sigma_{k+1}^{2}(\bV)$ to determine the proper number of clusters. 
Although this approach seems to be simplistic, interestingly, the following theorem tells that under appropriate assumptions, we can correctly find the number of circular cones with high probability. %, when the number of samples $N$ is large.

\begin{theorem}\label{thm:deterK}
  Suppose that  the data matrix $\bV \in \mathbb{R}^{F\times N}_{+}$ is generated according to  the generative process given in Theorem~\ref{thm:prob} where $K$ is the true number of circular cones. Further assume that the size angles for $K$ circular cones are all equal to $\alpha$, the angles between distinct basis vectors of the circular cones are all equal to $\beta$, and the parameters (inverse expectations) for the exponential distributions are all equal to  $\lambda$. In addition, we assume all the circular cones are contained in the nonnegative orthant $\mathcal{P}$ (cf.~Theorem~\ref{thm:prob_ext}) and $K \in \left\{K_{\mathrm{min}}, \ldots,K_{\mathrm{max}}\right \}$ with $K_{\mathrm{min}}>1$ and $K_{\mathrm{max}}<\mathrm{rank}(\bV)$. Then, for any $t\geq 1$, and sufficiently small $\epsilon$ satisfying \eqref{eq:epsilon} in Appendix \ref{app:prf_K}, 
     if  $N\geq c(t/\epsilon)^{2}F$ (for a constant $c>0$ depending only on $\lambda$, $\alpha$ and $\beta$), with probability at least $1-2\left(K_{\max}-K_{\min}+1\right)\exp\left(-t^{2}F\right)$, 
  \begin{equation}
    \frac{\sigma_{K} (\bV)}{\sigma_{K+1}(\bV)}=\max_{j \in\left \{ K_{\mathrm{min}}, \ldots, K_{\mathrm{max}}\right \}  } \frac{\sigma_{j}(\bV)  }{\sigma_{j+1}(\bV)}. \label{eqn:sigma_max}
  \end{equation} 
%  where $c$ is a positive constant depending on $\lambda$, $\alpha$ and $\beta$.
\end{theorem}
\begin{IEEEproof}
Please refer to Appendix \ref{app:prf_K} for the proof.
\end{IEEEproof}
In Section \ref{sec:estK}, we show numerically that the proposed method in \eqref{eq:estK} works well even when the geometric assumption is only  {\em approximately satisfied} (see Section \ref{sec:estK} for a formal definition)  assuming that  $N$ is sufficiently large. This shows that the determination of the correct number of clusters is robust to noise.% as predicted by the theory. 
\begin{remark}
The conditions of Theorem~\ref{thm:deterK} may appear to be rather restrictive. However, we make them only for the sake of convenience in presentation. We do not need to assume that the parameters of the exponential distribution are equal if, instead of $\sigma_j(\bV)$, we consider the singular values of a normalized version of $\bV$. The assumptions that all the size angles are the same and the angles between distinct basis vectors are the same can also be relaxed.  The theorem continues to hold even when the geometric assumption in~\eqref{eq:geo_ass} is not satisfied, i.e., $\beta \leq 4\alpha$. However, we empirically observe in Section~\ref{sec:estK}  that if $\bV$ satisfies the geometric assumption (even approximately), the results are superior compared to the scenario when the assumption is significantly violated. 
\end{remark}
\begin{remark}
We may replace the assumption that the circular cones are contained in the nonnegative orthant by removing  Step 4 in the generating process (projection onto $\calP$) in the generative procedure in Theorem \ref{thm:prob}. Because we are concerned with finding the number of clusters (or circular cones) rather than determining the true latent dimensionality of an NMF problem (cf.~\cite{Tan13}), we can discard the nonnegativity constraint. The number of clusters serves as a proxy for the latent dimensionality of NMF.
\end{remark}

\section{Numerical Experiments}\label{sec:experiment}%\vspace{-.05in}
\subsection{Experiments on Synthetic Data} \label{sec:syn}
To verify the correctness of our bounds, to observe the computational efficiency of the proposed algorithm,  and to check if the procedure for estimating $K$ is effective, we first perform numerical simulations on synthetic datasets. All the experiments  were executed on a Windows machine whose processor is an Intel(R) Core(TM) i5-3570, the CPU speed is 3.40 GHz, and the installed memory (RAM) is 8.00 GB. The Matlab version is 7.11.0.584 (R2010b). The Matlab codes for running the experiments can be found at \url{https://github.com/zhaoqiangliu/cr1-nmf}.

\subsubsection{Comparison of Relative Errors and Running Times} \label{sec:syn0}
\begin{figure}[t]
\subfloat{\includegraphics[width=.5\columnwidth,height=.4\columnwidth]{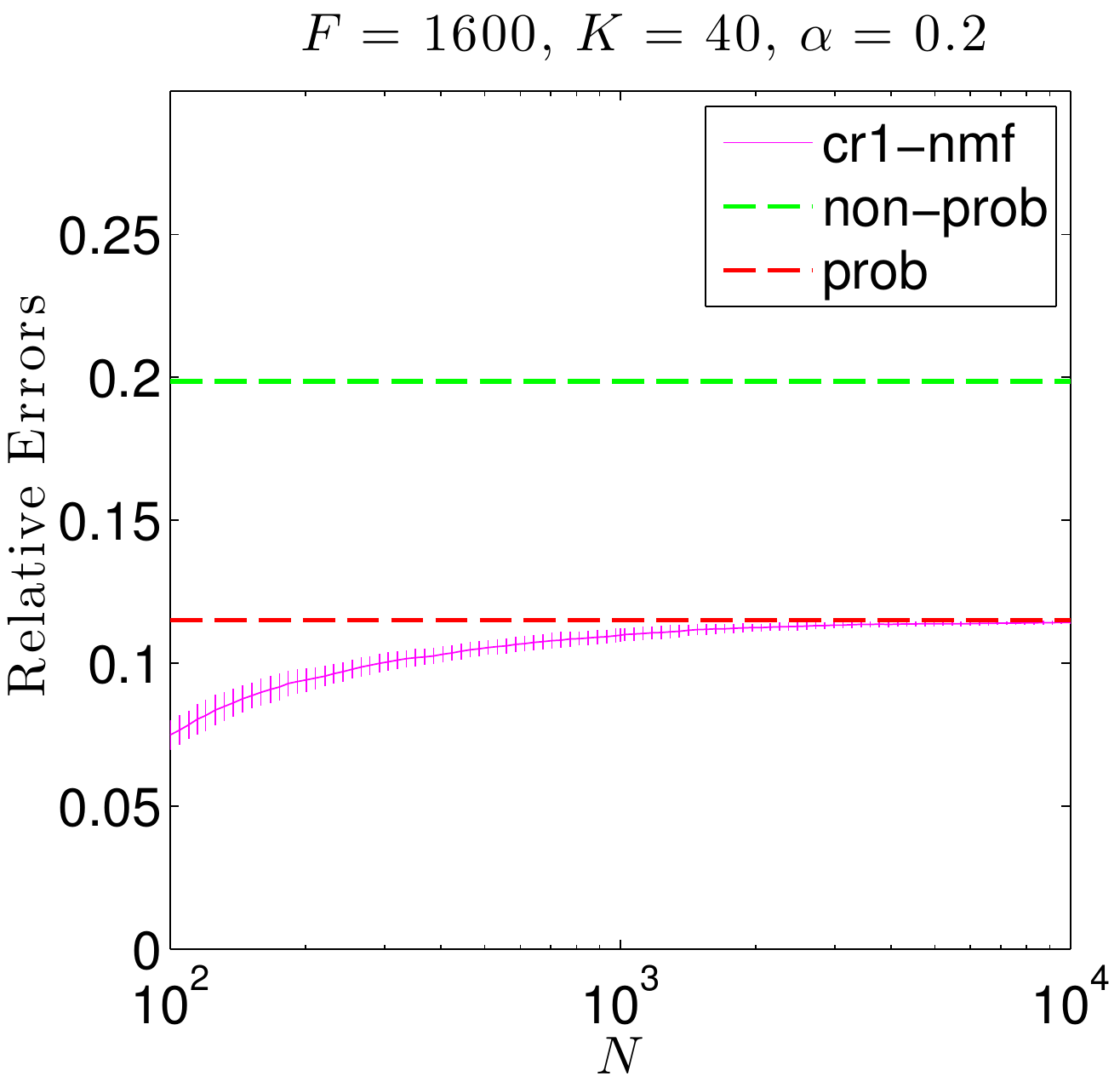}}\hfill
\subfloat{\includegraphics[width=.475\columnwidth,height=.4\columnwidth]{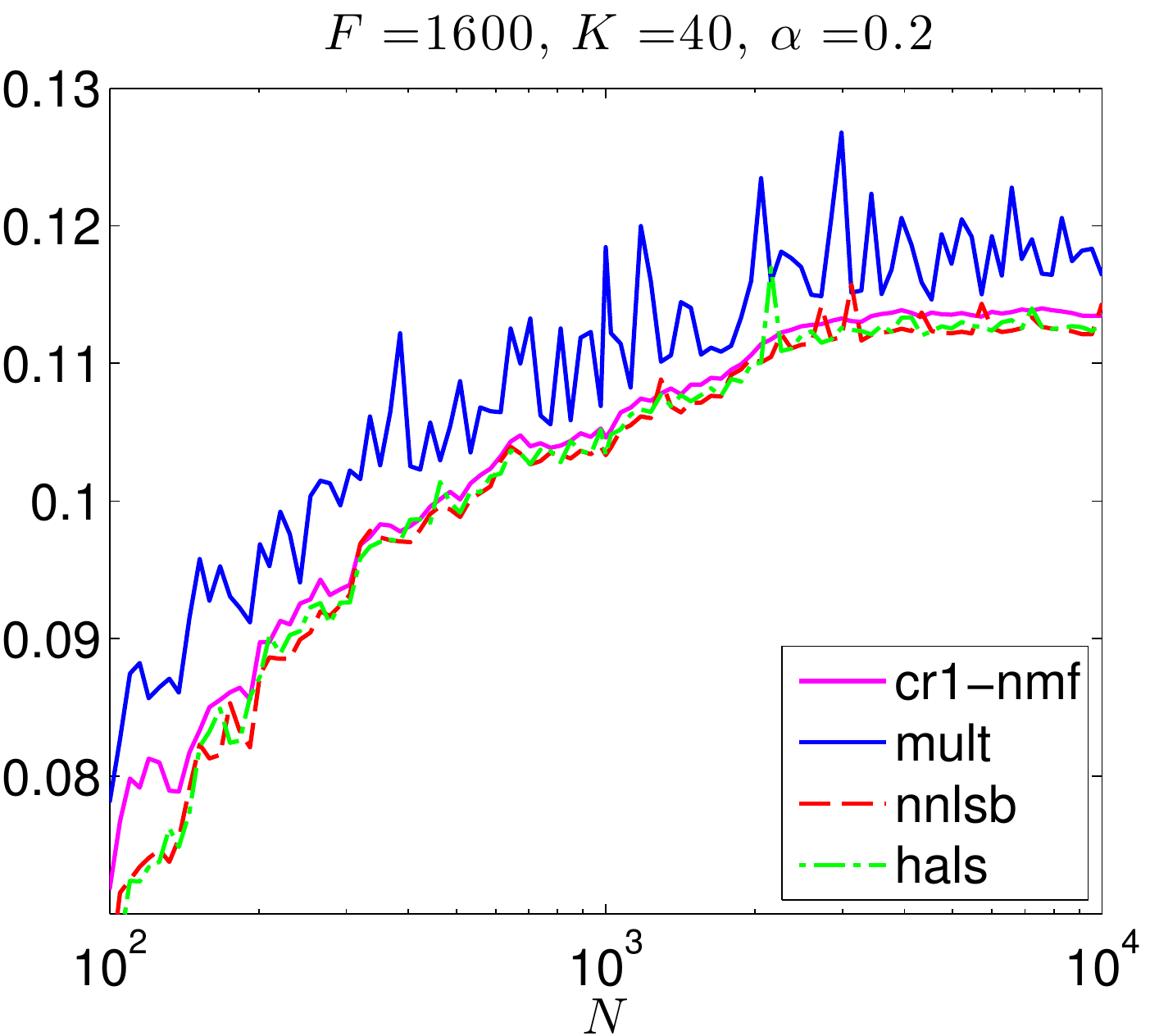}}\\
\subfloat{\includegraphics[width=.5\columnwidth,height=.4\columnwidth]{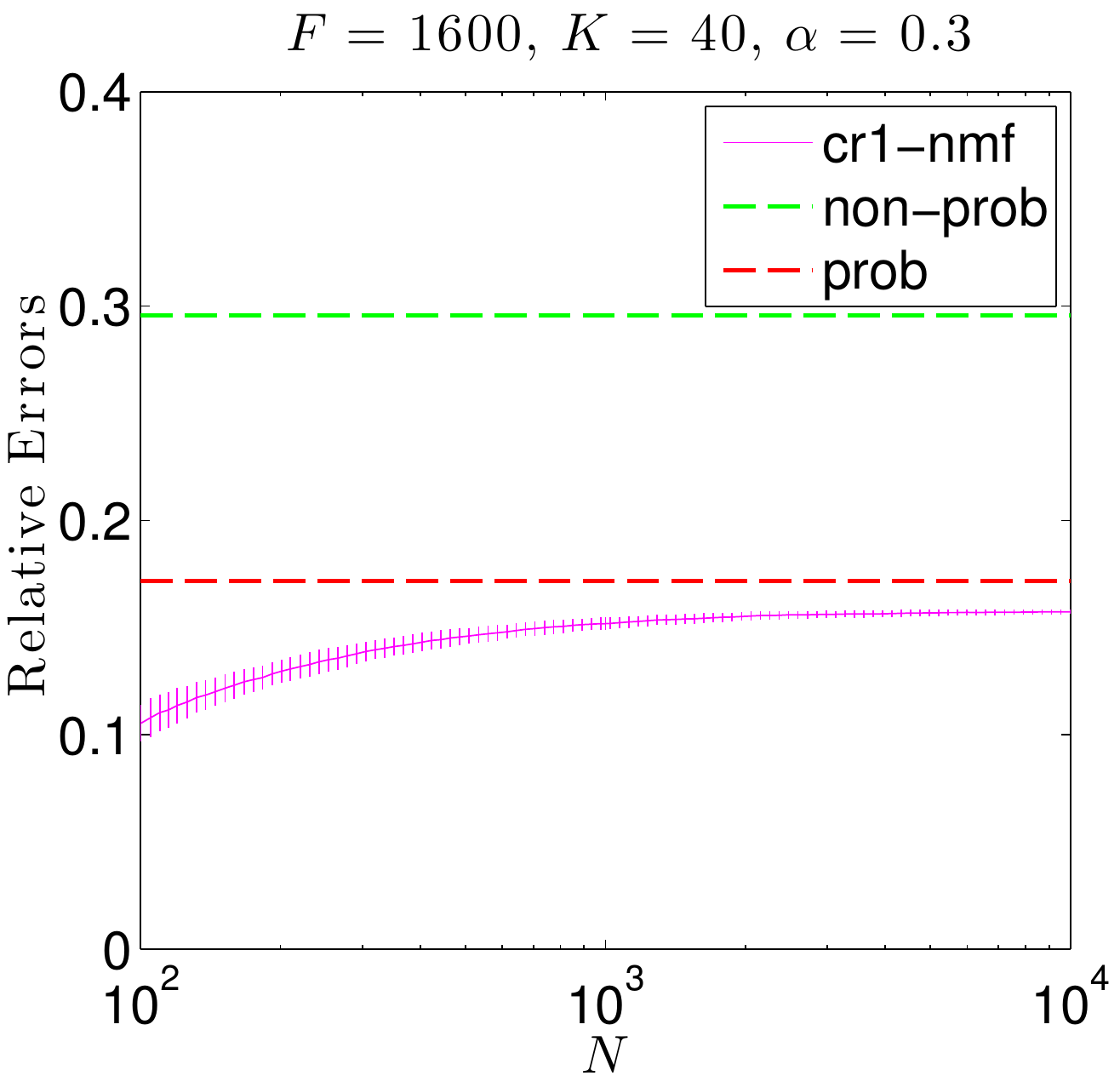}}\hfill
\subfloat{\includegraphics[width=.475\columnwidth,height=.4\columnwidth]{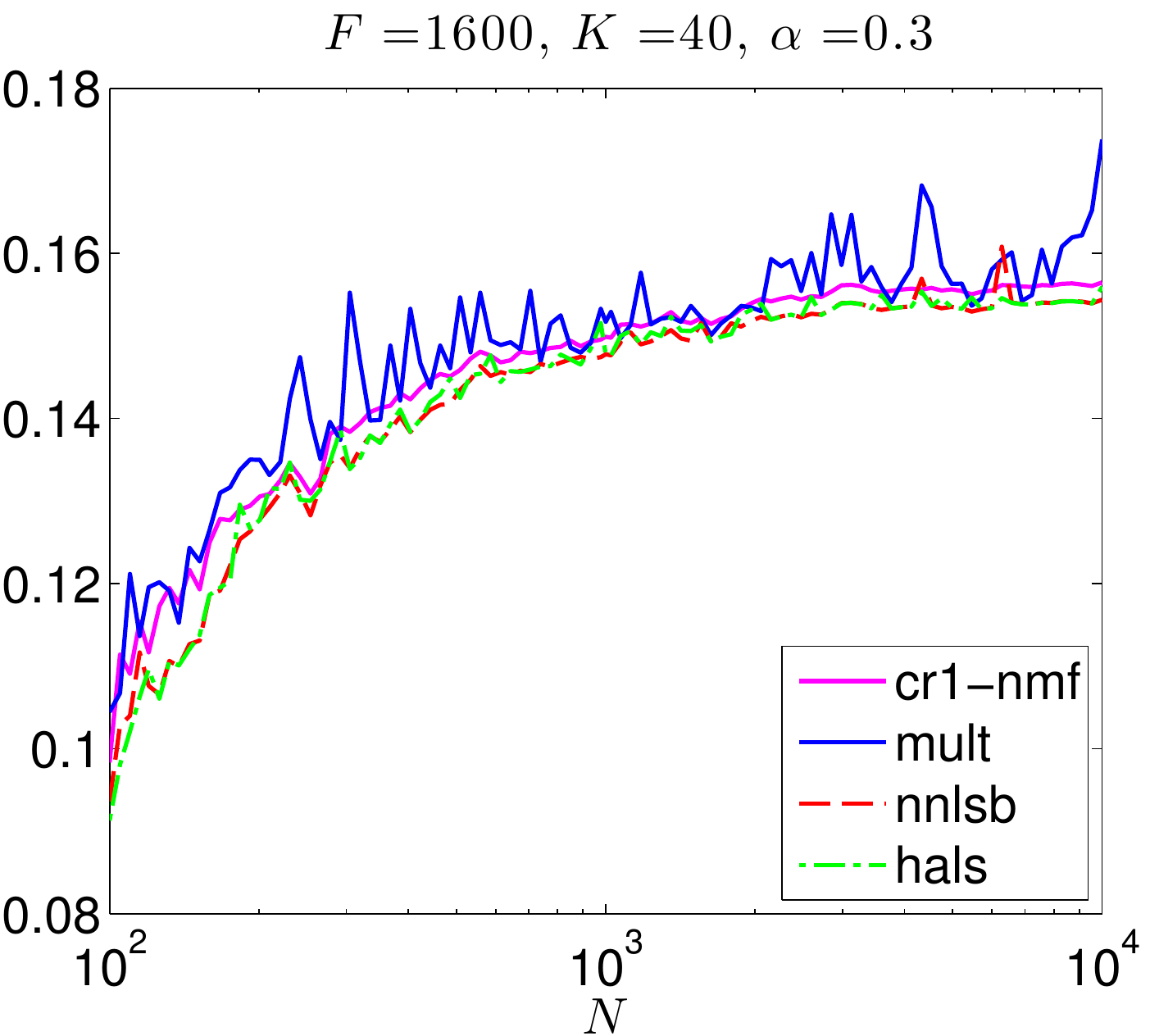}}
\caption{Errors and performances of various algorithms. On the left plot, we compare the empirical performance to the theoretical non-probabilistic and probabilistic bounds given by Theorems  \ref{thm:non-prob} and \ref{thm:prob} respectively. On the right plot, we compare the  empirical performance  to other NMF algorithms.}\label{fig:correct_comp}
\end{figure}

To generate the columns of $\bV$, given an integer $k \in [K]$ and an  angle $\beta \in \left[0,\alpha_{k}\right]$, we uniformly sample a vector $\bz$ from $\{\bx: \bx^{T}\bu_{k}=\cos \beta\}$, 
i.e., $\bz$ is a unit vector such that the angle between $\bz$ and $\bu_{k}$ is $\beta$. To achieve this, note that if $\bu_{k}=\be_{f}, f\in [F]$ ($\be_{f}$ is the  vector with only the $f$-th entry being $1$), this  uniform  sampling can easily  be achieved. For example, we can take $\bx=(\cos \beta) \be_{f}+(\sin \beta ) \by$, where $y(f)=0$, $y(i)=s(i)/\sqrt{\sum_{j\neq f}s(j)^{2}}, i\neq f$, and $s(i) \sim \mathcal{N}(0,1), i \neq f$. We can then use a Householder transformation \cite{Burden05} to map the unit vector generated from the circular cone with basis vector $\be_{f}$ to the unit vector generated from the circular cone with basis vector $\bu_{k}$. The corresponding Householder transformation matrix is (if $\bu_{k}=\be_{f}$, $\bP_{k}$ is set to be the identity matrix $\bI$)
\begin{equation}
  \bP_{k}=\bI-2\bz_{k}\bz_{k}^{T},\quad \mbox{where}\quad\bz_{k}=\frac{\be_{f}-\bu_{k}}{\|\be_{f}-\bu_{k}\|_{2}}. \label{eqn:house}
\end{equation}
In this set of experiments, we set the size angles $\alpha$ to be the same for all the circular cones. The angle between any two basis vectors is set to be $4\alpha+\Delta \alpha$ where $\Delta \alpha:=0.01$. The parameter for the exponential distribution $\blambda:=1./(1:K)$. We increase $N$ from $10^2$ to $10^4$ logarithmically. We fix the parameters $F=1600$, $K=40$ and $\alpha=0.2$ or $0.3$. The results shown in Figure~\ref{fig:correct_comp}. In the left plot of Figure~\ref{fig:correct_comp}, we compare the relative errors  of  Algorithm~\ref{algo:approx} (\texttt{cr1-nmf}) with the derived relative error bounds. In the right plot, we compare the relative errors of our algorithm with the relative errors of three classical algorithms: (i)  the  multiplicative update algorithm \cite{Lee_00} (\texttt{mult}); (ii)  the alternating nonnegative least-squares algorithm with block-pivoting   (\texttt{nnlsb}), which is reported to be one of the best alternating nonnegative least-squares-type algorithm for NMF in terms of both running time and approximation error \cite{Kim_08a}; (iii) and the hierarchical alternating least squares   algorithm~\cite{Cichocki07} (\texttt{hals}).  %We use an implementation of \texttt{mult} that is at least as fast as the corresponding one implemented in the \texttt{nnmf} function in Matlab. 
In contrast to these three algorithms, our algorithm is not iterative. The iteration numbers for \texttt{mult} and \texttt{hals} are set to   100, while the iteration number for \texttt{nnlsb} is set to   $20$, which is sufficient (in our experiments) for approximate  convergence. For statistical soundness of the results of the plots on the  left, $50$ data matrices $\bV\in \mathbb{R}_{+}^{F\times 10000}$ are independently generated  and for each data matrix $\bV$, we run our algorithm for $20$ runs. For the plots on the right, $10$ data matrices $\bV$ are independently generated  and all the algorithms are run for $10$ times for each $\bV$. We also compare the running time for these algorithms when they first achieve the approximation error smaller than or equal the approximation error of Algorithm~\ref{algo:approx}. The running times are shown in Table \ref{tab:compareTime}. Because the running times for $\alpha=0.2$ and $\alpha=0.3$ are similar, we only present the running times for the former.  

 From Figure~\ref{fig:correct_comp}, we observe that the relative errors obtained from Algorithm~\ref{algo:approx} are   smaller than the theoretical relative error bounds. When $\alpha=0.2$, the relative error of Algorithm~\ref{algo:approx} appears to converge to the probabilistic relative error bound as $N$ becomes large, but when $\alpha=0.3$, there is a gap between the relative error and the probabilistic relative error bound. From Theorems \ref{thm:prob} and \ref{thm:prob_ext}, we know that this difference is due to the projection of the cones to the nonnegative orthant. If there is no projection (this may violate the nonnegative constraint), the probabilistic relative error bound is tight as $N$ tends to infinity. We conclude that   when the size angle $\alpha$ is large, the projection step causes a  larger gap between the relative error and the probabilistic relative error bound. 
We   observe from Figure \ref{fig:correct_comp} that there are large oscillations for \texttt{mult}. Other algorithms achieve similar approximation errors. Table \ref{tab:compareTime} shows that classical NMF algorithms require significantly more time (at least an order of magnitude for large $N$) to achieve the same relative error compared to  our algorithm. 

\begin{table}[t]
\centering
\caption{Running times in seconds  of various algorithms ($\alpha=0.2$)}\label{tab:compareTime}
 \begin{tabular}{|c|c|c|c|c|} 
\hline
$N$ 	& \texttt{cr1-nmf} 			& \texttt{mult} & \texttt{nnlsb} 		& \texttt{hals} \\ \hline
$10^2 $ & $\mathbf{0.03} \!\pm\! 0.03$ 	& $1.56 \!\pm\! 0.76$  	& $5.82 \pm 1.15$  	& $0.46 \!\pm\! 0.20$ \\ \hline
$10^3 $ & $\mathbf{0.26} \!\pm\! 0.10$ 	& $9.54 \!\pm\! 5.91$  	& $6.44 \pm 2.70$  	& $3.01 \!\pm\! 1.85$ \\ \hline
$10^4 $ & $\mathbf{1.85} \!\pm\! 0.22$ 	& $85.92 \!\pm\! 54.51$ 	& $27.84 \pm 8.62$  	& $17.39 \!\pm\! 5.77$ \\ \hline
\end{tabular} %\vspace{-.1in}
\end{table} 

\subsubsection{Automatically Determining $K$} \label{sec:estK}
\begin{figure}[t]
\subfloat{\includegraphics[width=.5\columnwidth]{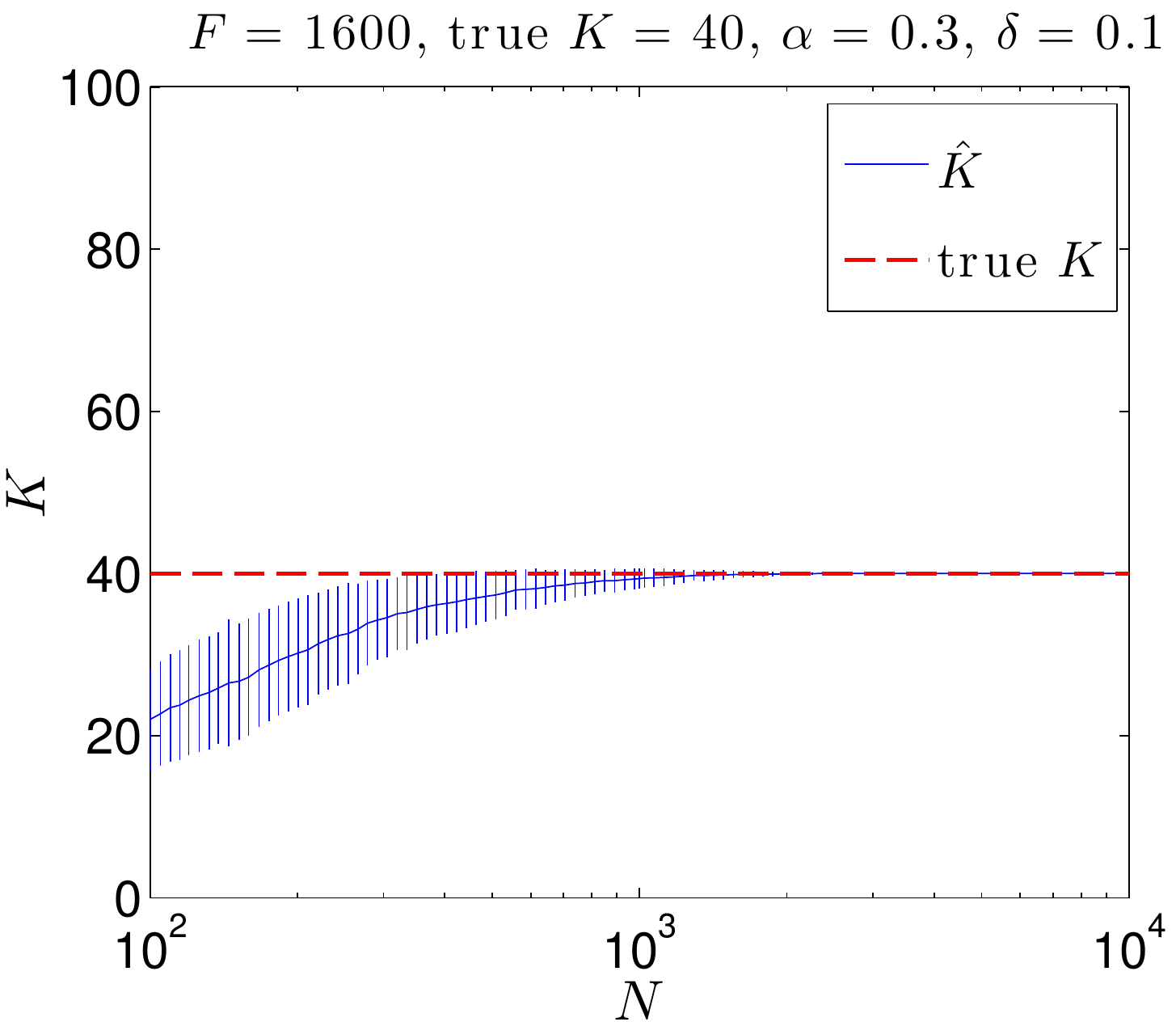}}\hfill
\subfloat{\includegraphics[width=.475\columnwidth]{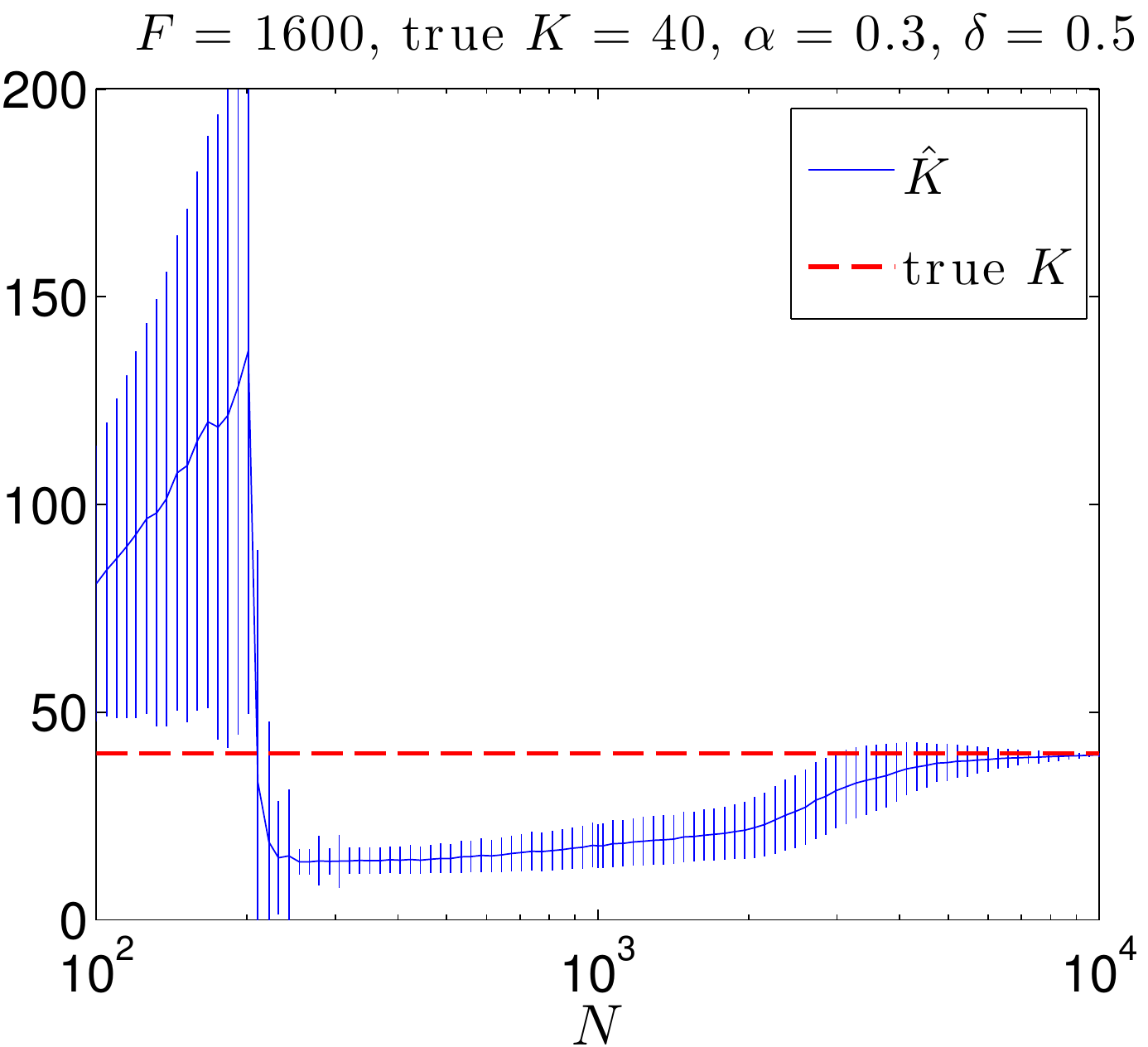}}
\caption{Estimated number of circular cones $K$ with different noise levels. The error bars denote one standard deviation away from the mean.  }\label{fig:est_K}
\end{figure}

We now verify the efficacy and  the robustness of the proposed method  in \eqref{eq:estK} for automatically determining the correct number of circular cones. We generated the data matrix  $\hat{\bV}:=[ \bV +\delta \bE]_{+}$, where each entry of $\bE$ is sampled i.i.d.\ from the standard normal distribution,   $\delta>0$ corresponds to the noise magnitude, and $[\cdot]_{+}$ represents the projection to nonnegative orthant operator. We generated the nominal/noiseless data matrix $\bV$ by setting $\alpha=0.3$, the true number of circular cones $K=40$, and other parameters similarly to the procedure in Section~\ref{sec:syn0}. The noise magnitude $\delta$ is set to be either $0.1$ or $0.5$; the former simulates a relatively clean setting in which the geometric assumption is approximately satisfied, while in the latter, $\hat{\bV}$ is far from a matrix that satisfies the geometric assumption, i.e., a very noisy scenario. We generated $1000$ perturbed data matrices $\hat{\bV}$ independently. From Figure~\ref{fig:est_K} in which the true $K=40$, we observe that, as expected, the method in~\eqref{eq:estK} works well if the noise level is small. Somewhat surprisingly,  it also works well even when the noise level is relatively high (e.g., $\delta=0.5$) if the   number of data points $N$ is also commensurately  large (e.g., $N\ge 5\times 10^3$).   

\subsection{Experiments on Real Datasets}\label{sec:real}
\subsubsection{Initialization Performance in Terms of the Relative Error}\label{sec:real_relerr}
Because real datasets do not, in general, strictly satisfy the geometric assumption, our algorithm \texttt{cr1-nmf}, does not achieve as low a relative error compared to other NMF algorithms.  
However, similar to the popular spherical k-means (\texttt{spkm}; we use $10$ iterations to produce its initial left factor matrix $\bW$) algorithm \cite{Wild04}, our algorithm may be used as {\em initialization method} for NMF. In this section, we compare \texttt{cr1-nmf} to other classical and popular initialization approaches for NMF. These  include random initialization (\texttt{rand}), \texttt{spkm}, and the \texttt{nndsvd} initialization method~\cite{Bou08} (\texttt{nndsvd}). We empirically show that our algorithm, when used as an initializer,  achieves the best performance when combined with classical NMF algorithms. The specifications of the real datasets and the running times for the initialization methods are presented in Tables \ref{tab: real_data}  
and~\ref{tab: runt_init} respectively. 

\begin{table}[t]
\centering
\caption{Information for real datasets used}\label{tab: real_data}
 \begin{tabular}{|c|c|c|c|c|} 
\hline
Dataset Name 			& $F$ 			& $N$ 	& $K$ 	& Description 	\\ \hline
CK\tablefootnote{http://www.consortium.ri.cmu.edu/ckagree/}		& 49$\times$64  	& 8795  & 97  	& face dataset 	\\ \hline
faces94\tablefootnote{http://cswww.essex.ac.uk/mv/allfaces/faces94.html}  	& 200$\times$180  	& 3040  & 152  	& face dataset 	\\ \hline
Georgia Tech\tablefootnote{http://www.anefian.com/research/face\_reco.htm} & 480$\times$640  	& 750 	& 50  	& face dataset 	\\ \hline
PaviaU\tablefootnote{http://www.ehu.eus/ccwintco/index.php?title=Hyperspectral\_Remote\_\\Sensing\_Scenes}  & 207400 & 103 &  9 & hyperspectral \\ \hline
\end{tabular} %\vspace{-.1in}
\end{table} 

\begin{table}
\centering
\caption{Running times for initialization}\label{tab: runt_init}
 \begin{tabular}{|c|c|c|c|} 
\hline
Dataset Name 		& \texttt{cr1-nmf} 			& \texttt{spkm} 			& \texttt{nndsvd}  	\\ \hline
CK  		& $\mathbf{3.30}\pm 0.10$ 	& $6.68 \pm 0.71$ 	& $9.45 \pm 0.12$ 	\\ \hline
faces94  	& $\mathbf{14.50} \pm 0.20$ 	& $32.23 \pm 2.28$ 	& $32.81 \pm 0.29$	\\ \hline
%grimace 	& $\mathbf{0.83} \pm 0.15$ 	& $1.25 \pm 0.33$ 	& $0.89 \pm 0.23$ 	\\ \hline
Georgia Tech 	& $\mathbf{18.90} \pm 1.13$ 	& $24.77 \pm 3.58$ 	& $21.28 \pm 0.35$	\\ \hline
PaviaU  	& $\mathbf{0.73} \pm 0.11$ 	& $2.47 \pm 0.48$ 	& $0.84 \pm 0.12$ 	\\ \hline
\end{tabular} %\vspace{-.1in}
\end{table} 

We use \texttt{mult}, \texttt{nnlsb}, and \texttt{hals} as the classical NMF algorithms that are combined with the initialization approaches. Note that for \texttt{nnlsb}, we only need to initialize the left factor matrix $\bW$. This is because the initial $\bH$ can be obtained from initial $\bW$ using \cite[Algorithm~2]{Kim_08a}.
Also note that   the pair $\left(\bW^{*},\bH^{*}\right)$ produced by   Algorithm \ref{algo:approx} is a fixed point  for \texttt{mult} (see Lemma~\ref{lem:fix_pt} in Appendix~\ref{app:inv}), so we use a small perturbation of $\bH^{*}$ as an  initialization for the right factor matrix. 
For \texttt{spkm}, similarly to~\cite{Wild04,Bou08}, we initialize the right factor matrix randomly. In addition, to ensure a fair comparison between these initialization approaches, we need to shift the iteration numbers appropriately, i.e., the initialization method that takes a longer time should start with a commensurately smaller iteration number when combined one of the three classical NMF algorithms. Table \ref{tab: shifts} reports the  number of shifts. Note that unlike \texttt{mult} and \texttt{hals}, the running times for different iterations of \texttt{nnlsb} can be significantly different. We observe that for most datasets, when run for the same number of iterations, random initialization and \texttt{nndsvd} initialization not only result in larger relative errors, but they also take a much longer time than \texttt{spkm} and our initialization approach. Because initialization methods can also affect the running time of each iteration of \texttt{nnlsb} significantly, we do not report shifts for initialization approaches when  combined with \texttt{nnlsb}. Table \ref{tab: nnlsb} reports running times that  various algorithms first achieve a fixed relative error $\epsilon>0$ for various initialization methods when combined with \texttt{nnlsb}. Our proposed algorithm is clearly superior. 

\iffalse
\begin{table}[t]
\centering
\caption{Shift number for initialization approaches}\label{tab: shifts}
 \begin{tabular}{|c|c|c|c|c|c|} 
\hline
		& CK 	& faces94	& grimace 	& Georgia Tech 	& PaviaU 	\\ \hline
cr1-nmf$+$mult 	& 3 	& 2 		& 4  		& 3  		& 2 		\\ \hline
spkm$+$mult  	& 6	& 5 		& 6  		& 4  		& 7 		\\ \hline
nndsvd$+$mult  	& 8 	& 5 		& 4 		& 3  		& 2 		\\ \hline
cr1-nmf$+$hals 	& 2  	& 2		& 3  		& 2  		& 1 		\\ \hline
spkm$+$hals  	& 5 	& 4 		& 5  		& 3  		& 5 		\\ \hline
nndsvd$+$hals  	& 7 	& 4 		& 3 		& 2  		& 1 		\\ \hline
\end{tabular} %\vspace{-.1in}
\end{table} 
\fi

\begin{table}[t]
\centering
\caption{Shift number for initialization approaches}\label{tab: shifts}
 \begin{tabular}{|c|c|c|c|c|c|} 
\hline
		& CK 	& faces94	& Georgia Tech 	& PaviaU 	\\ \hline
\texttt{cr1-nmf}$+$\texttt{mult} 	& 3 	& 2 		& 3  		& 2 		\\ \hline
\texttt{spkm}$+$\texttt{mult}  	& 6	& 5 		& 4  		& 7 		\\ \hline
\texttt{nndsvd}$+$\texttt{mult}  	& 8 	& 5 		& 3  		& 2 		\\ \hline
\texttt{cr1-nmf}$+$\texttt{hals} 	& 2  	& 2		& 2  		& 1 		\\ \hline
\texttt{spkm}$+$\texttt{hals}  	& 5 	& 4 		& 3  		& 5 		\\ \hline
\texttt{nndsvd}$+$\texttt{hals}  	& 7 	& 4 		& 2  		& 1 		\\ \hline
\end{tabular} %\vspace{-.1in}
\end{table} 

\begin{small}
\begin{table}[t]
\centering
\caption{Running times when algorithm first achieve relative error $\epsilon$ for initialization methods combined with \texttt{nnlsb}}\label{tab: nnlsb}
\begin{tabular}{|c|c|c|c|} 
\hline
CK       	& $\epsilon=0.105$ 		& $\epsilon=0.100$ 		& $\epsilon=0.095$ 		\\ \hline
\texttt{rand} 		& $727.53 \!\pm\! 23.93$  		& $1389.32 \!\pm\! 61.32$  		& -- 				\\ \hline      
\texttt{cr1-nmf} 	& $\mathbf{40.27} \!\pm\! 1.96$  	& $\mathbf{71.77} \!\pm\! 2.83$  	& $\mathbf{129.62} \!\pm\! 5.98$	\\ \hline
\texttt{spkm}  		& $79.37 \!\pm\! 2.52$ 	 	& $91.23 \!\pm\! 2.69$  		& $240.12 \!\pm\! 5.32$ 		\\ \hline
\texttt{nndsvd}   	& $309.25 \!\pm\! 6.24$ 		& $557.34 \!\pm\! 7.59$  		& $1309.51 \!\pm\! 21.97$		\\ \hline\hline
faces94       	& $\epsilon=0.140$ 		& $\epsilon=0.135$ 		& $\epsilon=0.131$ 		\\ \hline
\texttt{rand} 		& $2451.8 \!\pm\! 26.6$  		& $7385.8 \!\pm\! 49.6$  		& -- 				\\ \hline      
\texttt{cr1-nmf} 	& $\mathbf{338.8} \!\pm\! 11.1$  	& $\mathbf{706.3} \!\pm\! 13.3$  	& $\mathbf{3585.2} \!\pm\! 49.4$	\\ \hline
\texttt{spkm}  		& $465.3 \!\pm\! 13.5$  		& $1231.1 \!\pm\! 28.5$  		& $5501.4 \!\pm\! 134.4$ 		\\ \hline
\texttt{nndsvd}   	& $1531.5 \!\pm\! 6.4$ 		& $3235.8 \!\pm\! 12.1$  		& $10588.6 \!\pm\! 35.9$		\\ \hline\hline
%grimace       	& $\epsilon=0.140$ 		& $\epsilon=0.130$ 		& $\epsilon=0.120$ 		\\ \hline
%rand 		& $37.79 \!\pm\! 1.10$  		& $81.03 \!\pm\! 2.33$  		& -- 				\\ \hline      
%cr1-nmf 	& $\mathbf{2.89} \!\pm\! 0.20$  	& $\mathbf{5.85} \!\pm\! 0.21$  	& $15.98 \!\pm\! 0.63$		\\ \hline
%spkm  		& $4.29 \!\pm\! 0.45$  		& $6.46 \!\pm\! 0.56$  		& $\mathbf{15.76} \!\pm\! 0.85$ 	\\ \hline
%nndsvd   	& $12.35 \!\pm\! 0.27$ 		& $18.55 \!\pm\! 0.28$  		& $50.28 \!\pm\! 0.43$		\\ \hline\hline
Georgia Tech   	& $\epsilon=0.185$ 		& $\epsilon=0.18$ 		& $\epsilon=0.175$ 		\\ \hline
\texttt{rand} 		& $3766.7 \!\pm\! 92.8$  		& $5003.7 \!\pm\! 126.8$ 		& $7657.4 \!\pm\! 285.9$ 		\\ \hline      
\texttt{cr1-nmf} 	& $\mathbf{147.3} \!\pm\! 2.8$  	& $\mathbf{308.2} \!\pm\! 7.8$  	& $\mathbf{1565.0} \!\pm\! 59.5$	\\ \hline
\texttt{spkm}  		& $253.2 \!\pm\! 20.1$  		& $537.4 \!\pm\! 43.4$  		& $2139.2 \!\pm\! 142.9$ 		\\ \hline
\texttt{nndsvd}   	& $2027.0 \!\pm\! 7.0$ 		& $2819.4 \!\pm\! 9.5$  		& $4676.4 \!\pm\! 15.3$		\\ \hline\hline
PaviaU        	& $\epsilon=0.0230$ 		& $\epsilon=0.0225$ 		& $\epsilon=0.0220$ 		\\ \hline
\texttt{rand} 		& $192.51 \!\pm\! 16.11$  		& $224.65 \!\pm\! 16.17$ 		& $289.48 \!\pm\! 16.74$ 		\\ \hline      
\texttt{cr1-nmf} 	& $\mathbf{13.30} \!\pm\! 0.40$  	& $\mathbf{16.93} \!\pm\! 0.61$  	& $\mathbf{30.06} \!\pm\! 0.94$	\\ \hline
\texttt{spkm}  		& $32.00 \!\pm\! 3.16$  		& $40.27 \!\pm\! 4.39$  		& $52.40 \!\pm\! 6.29$ 		\\ \hline
\texttt{nndsvd}  	& $79.92 \!\pm\! 0.84$ 		& $106.29 \!\pm\! 0.91$  		& $160.10 \!\pm\! 0.92$		\\ \hline
\end{tabular} %\vspace{-.1in}
\end{table} 
\end{small}

\begin{figure}[t]
\subfloat{\includegraphics[width=.33\columnwidth,height=.37\columnwidth]{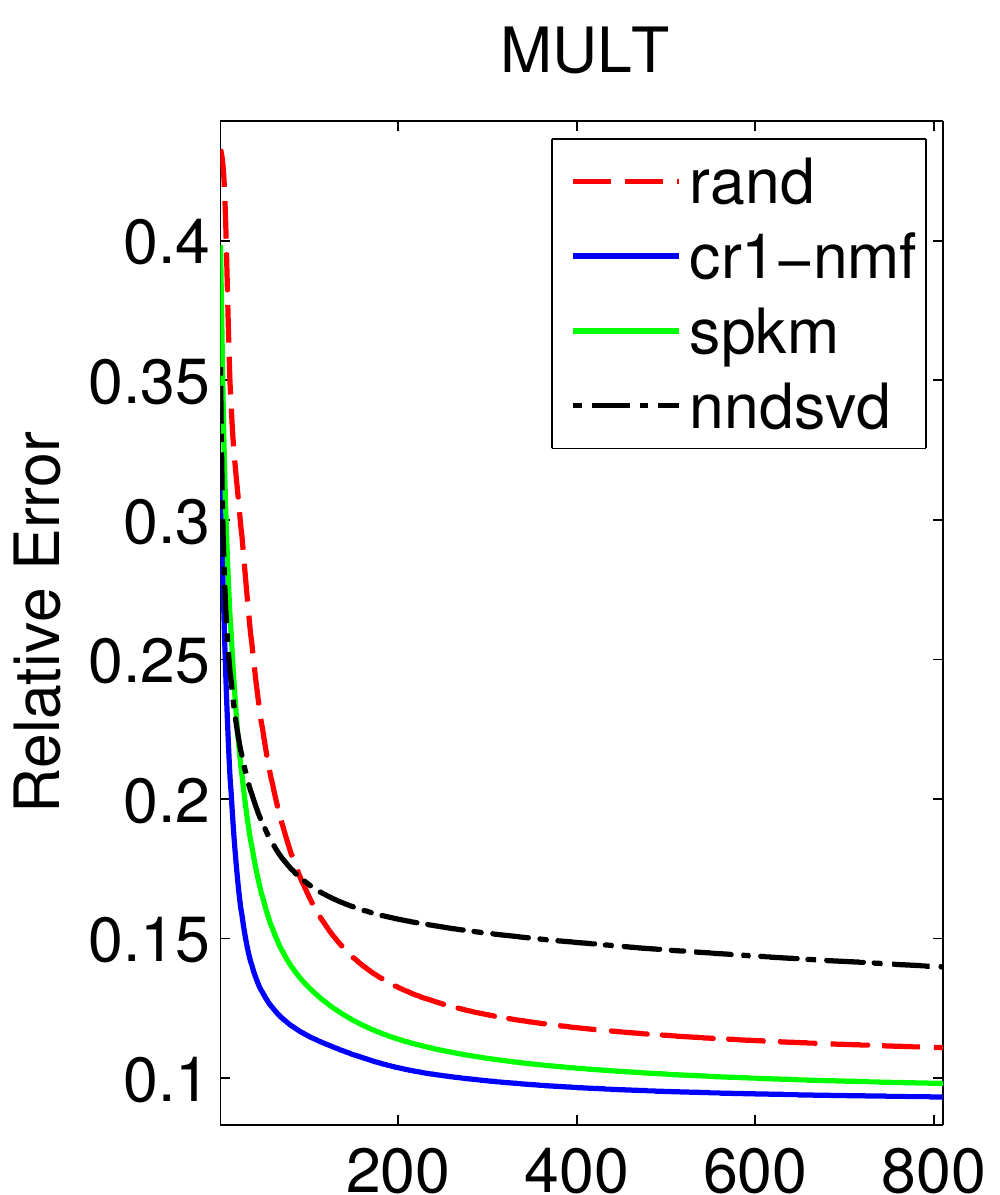}}\hfill
\subfloat{\includegraphics[width=.33\columnwidth,height=.37\columnwidth]{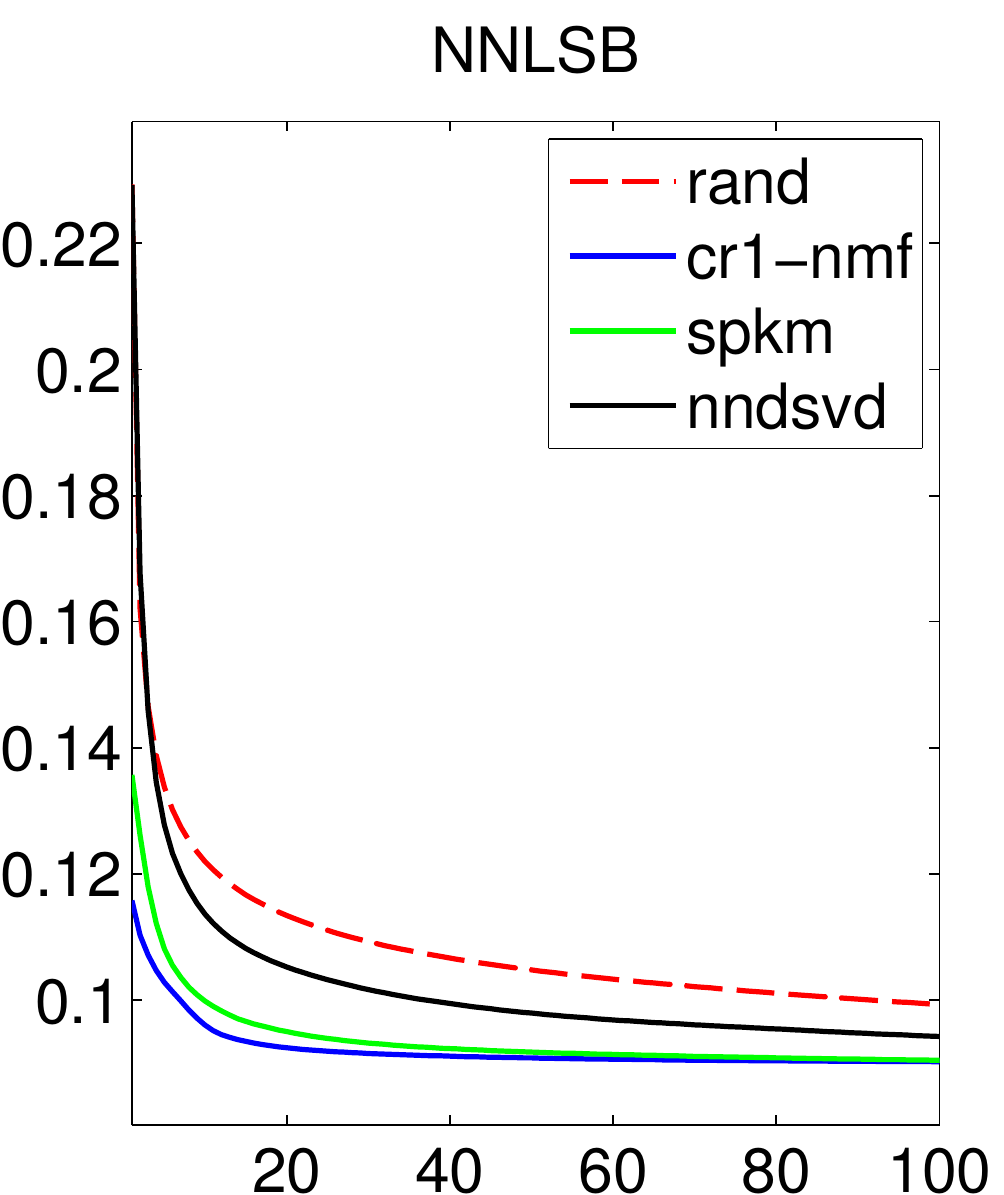}}\hfill
\subfloat{\includegraphics[width=.33\columnwidth,height=.37\columnwidth]{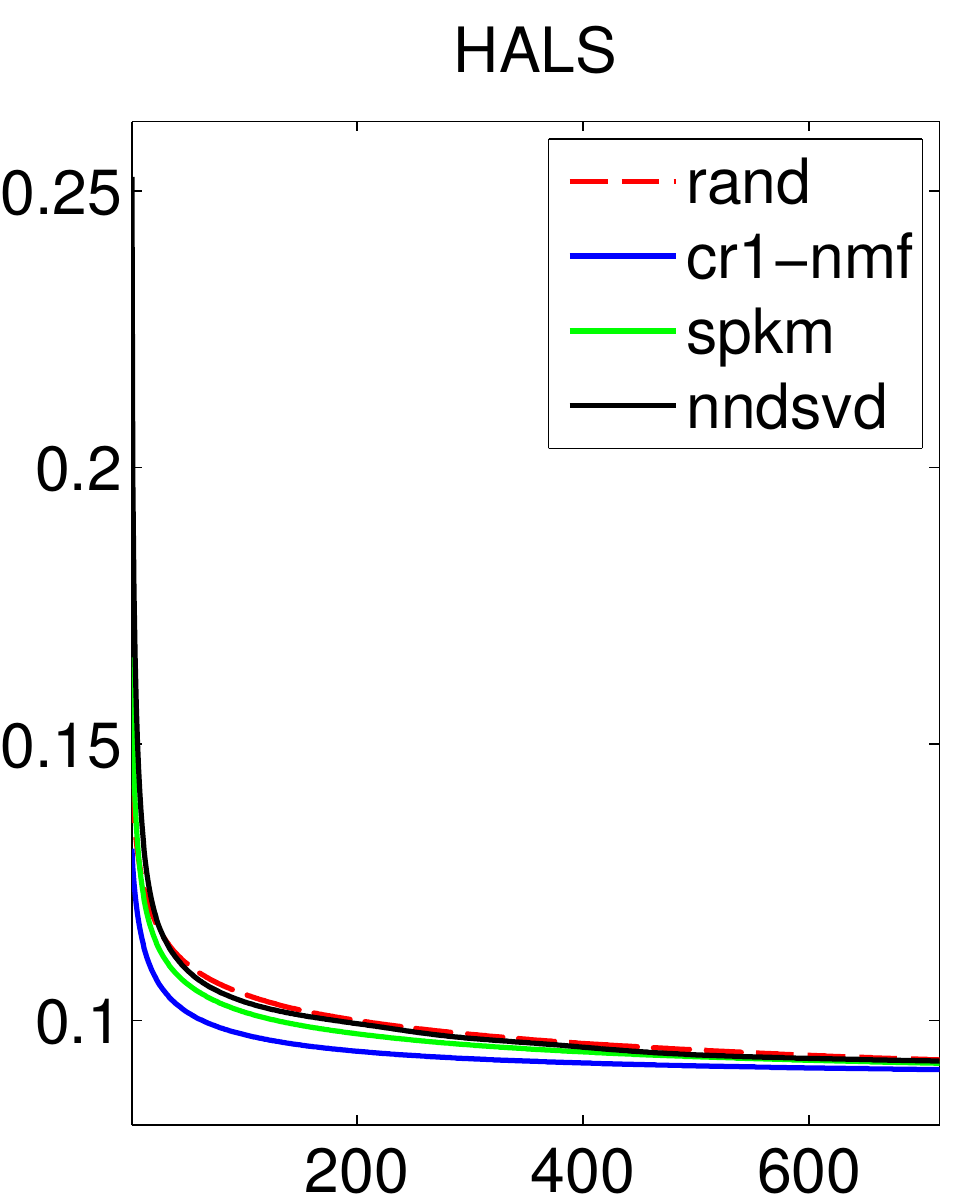}}\\
\subfloat{\includegraphics[width=.33\columnwidth,height=.37\columnwidth]{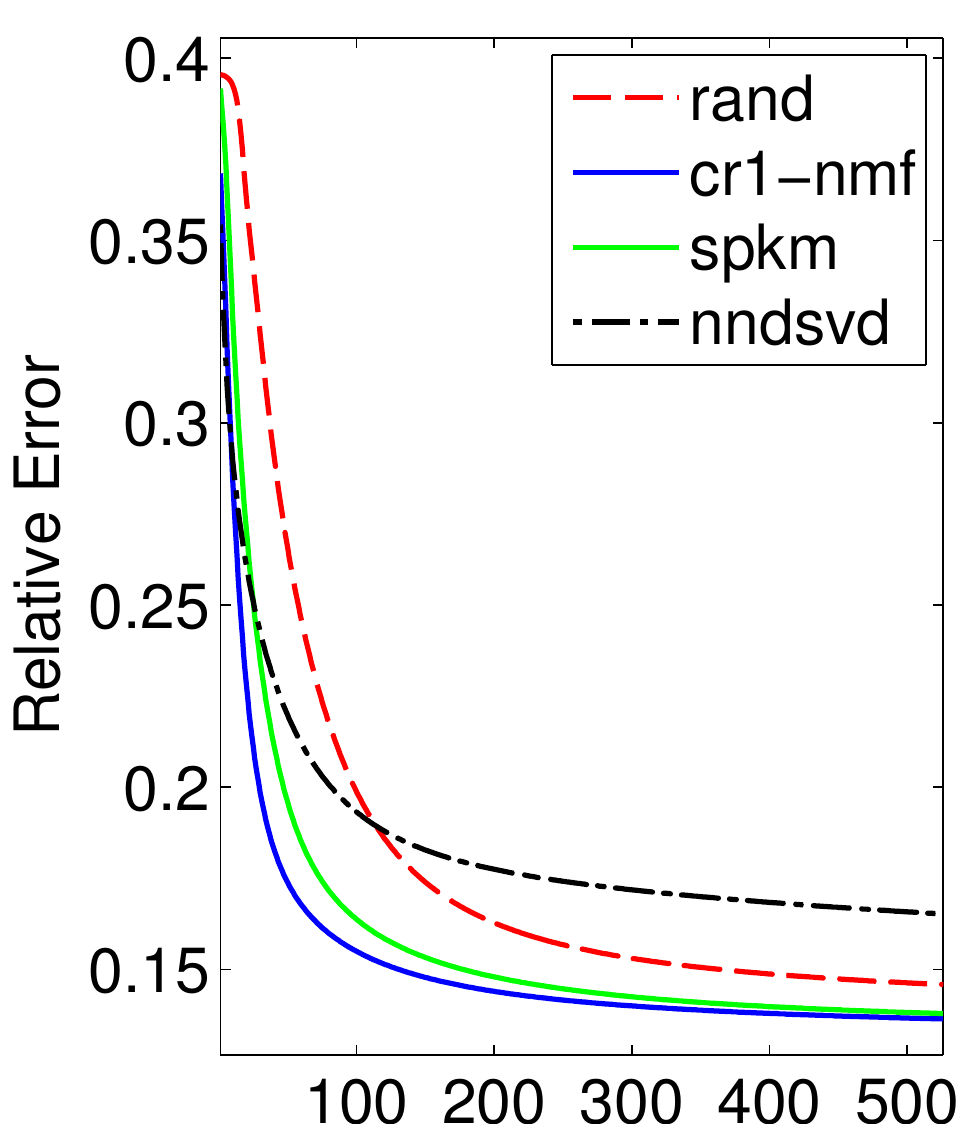}}\hfill
\subfloat{\includegraphics[width=.33\columnwidth,height=.37\columnwidth]{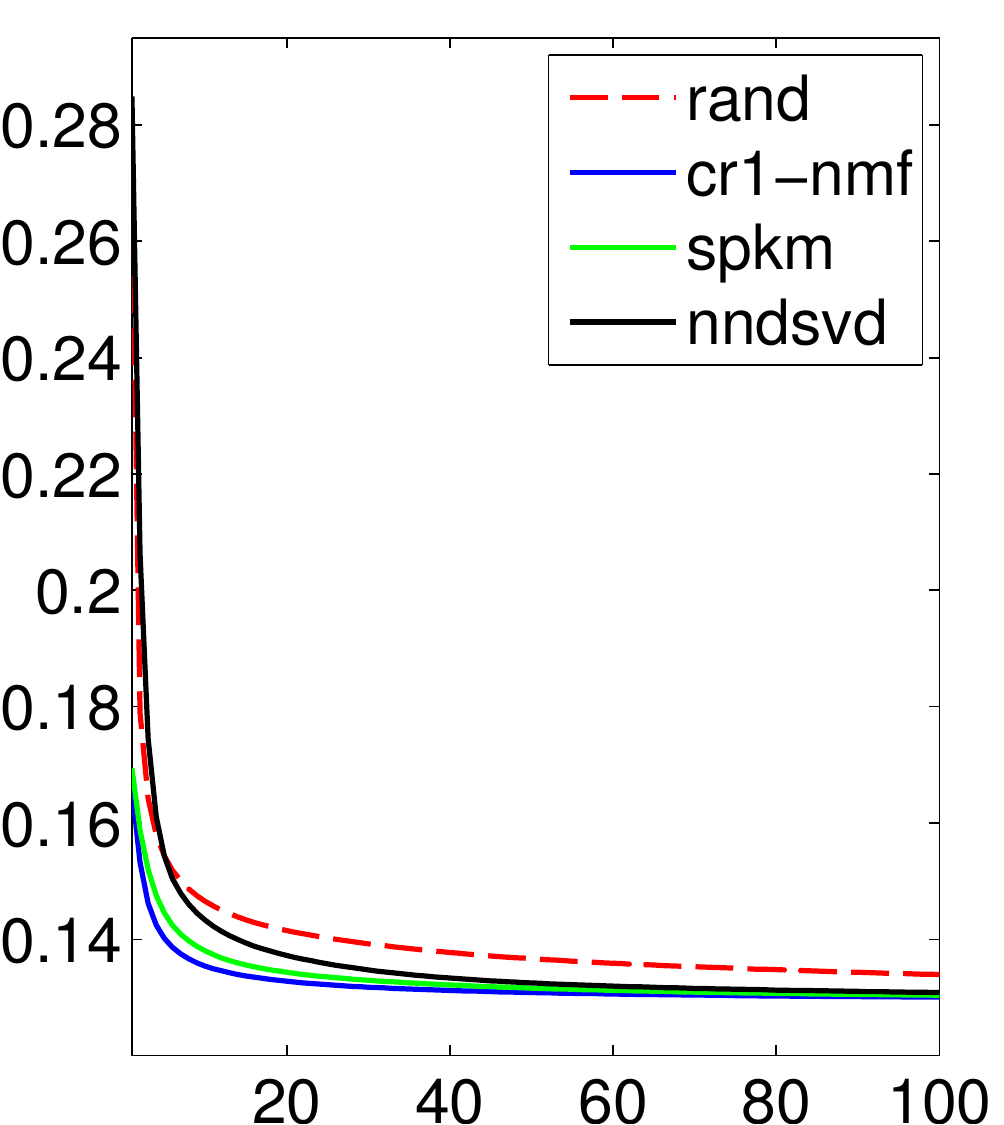}}\hfill
\subfloat{\includegraphics[width=.33\columnwidth,height=.37\columnwidth]{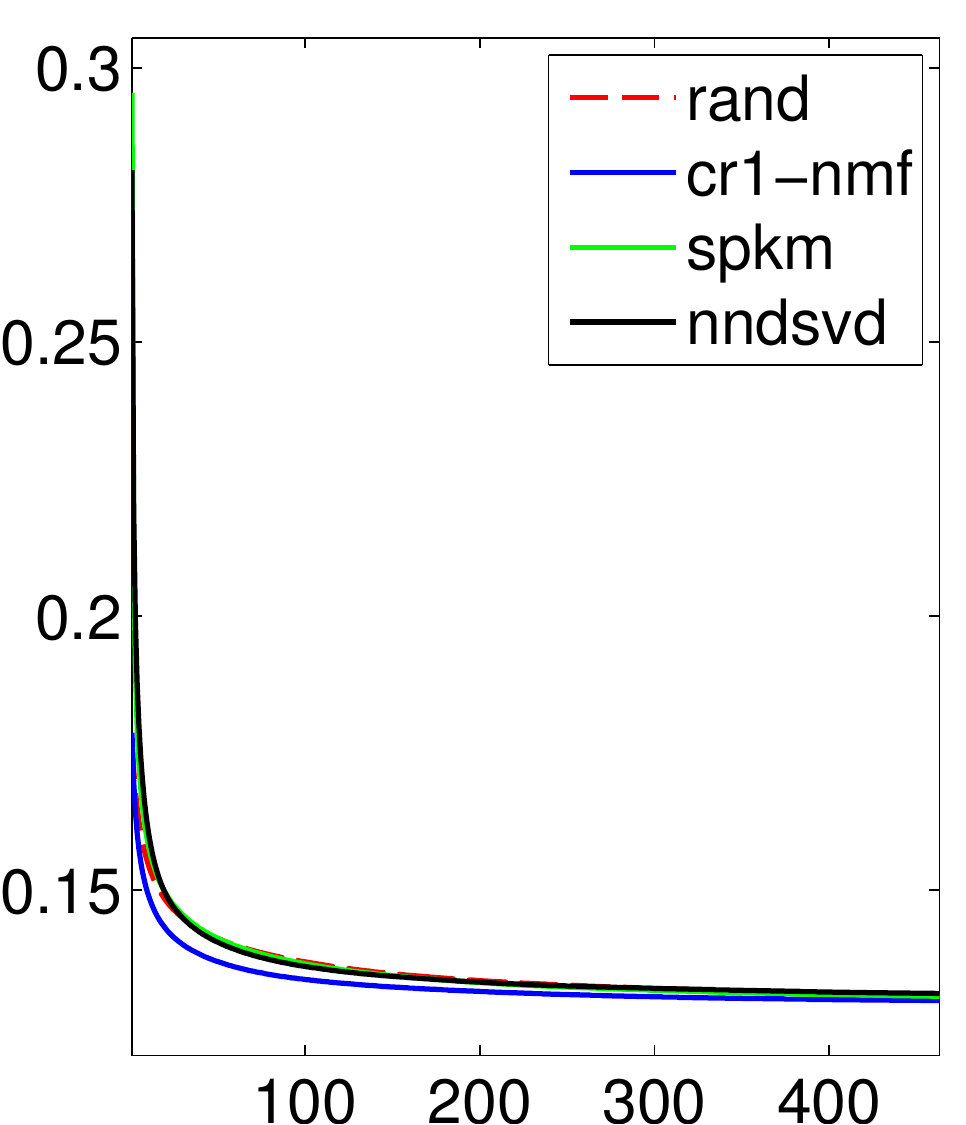}}\\
\subfloat{\includegraphics[width=.33\columnwidth,height=.37\columnwidth]{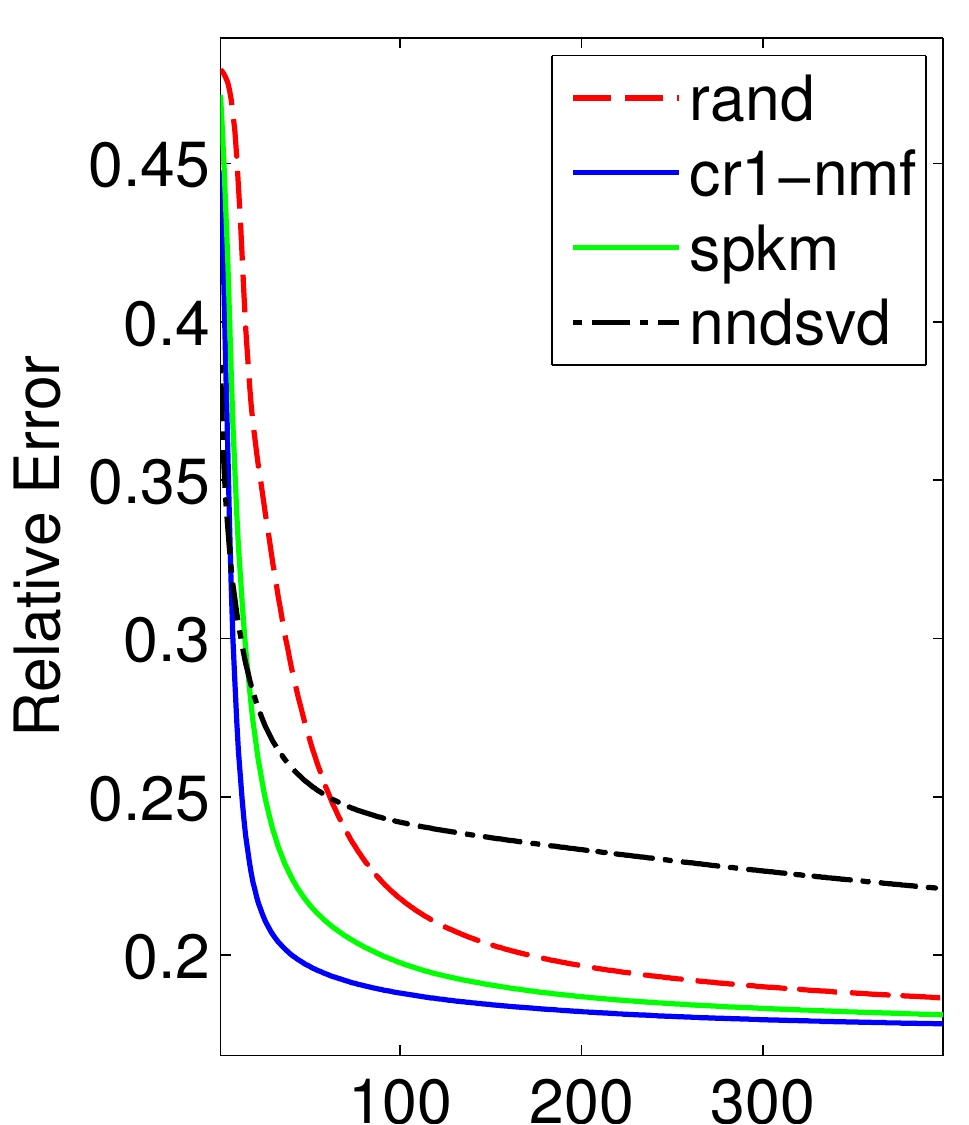}}\hfill
\subfloat{\includegraphics[width=.33\columnwidth,height=.37\columnwidth]{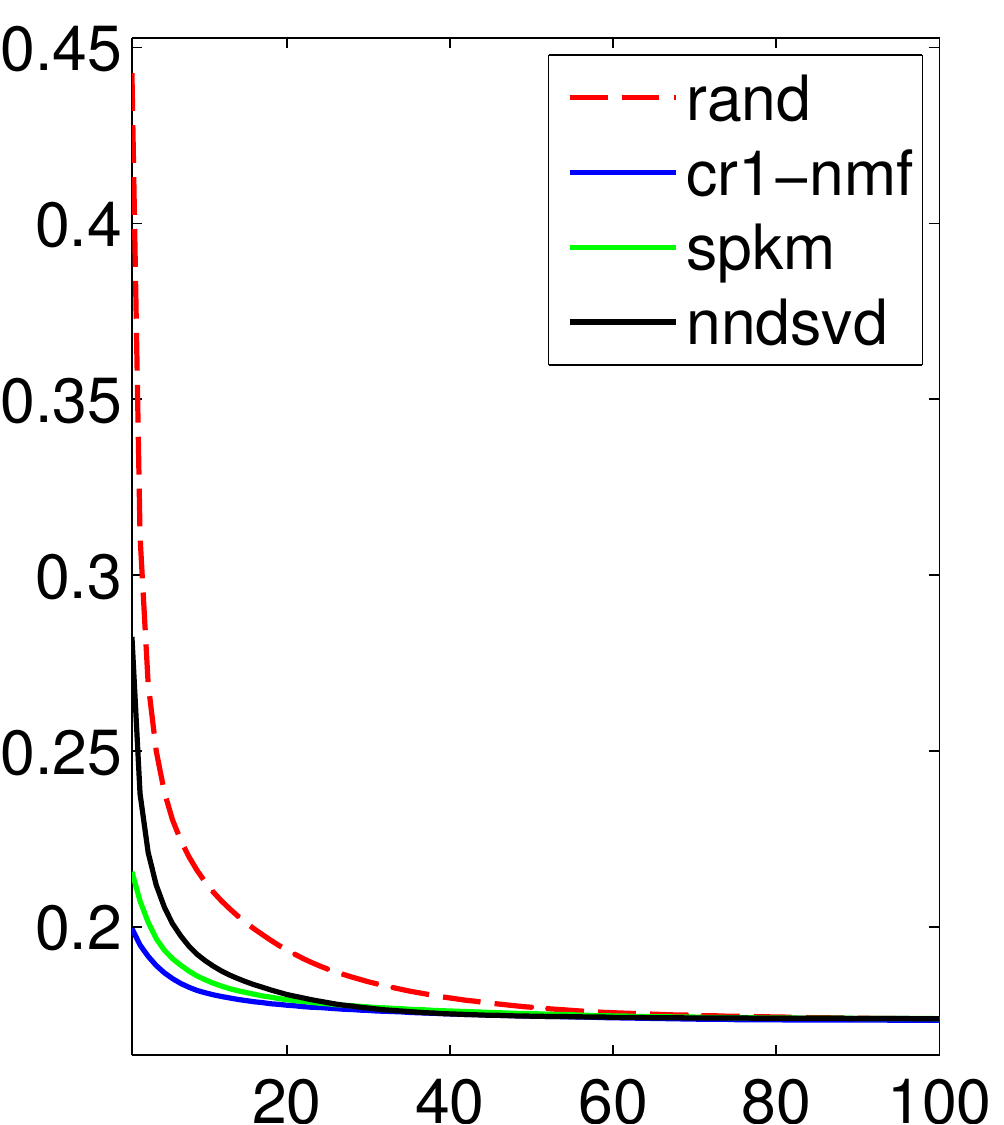}}\hfill
\subfloat{\includegraphics[width=.33\columnwidth,height=.37\columnwidth]{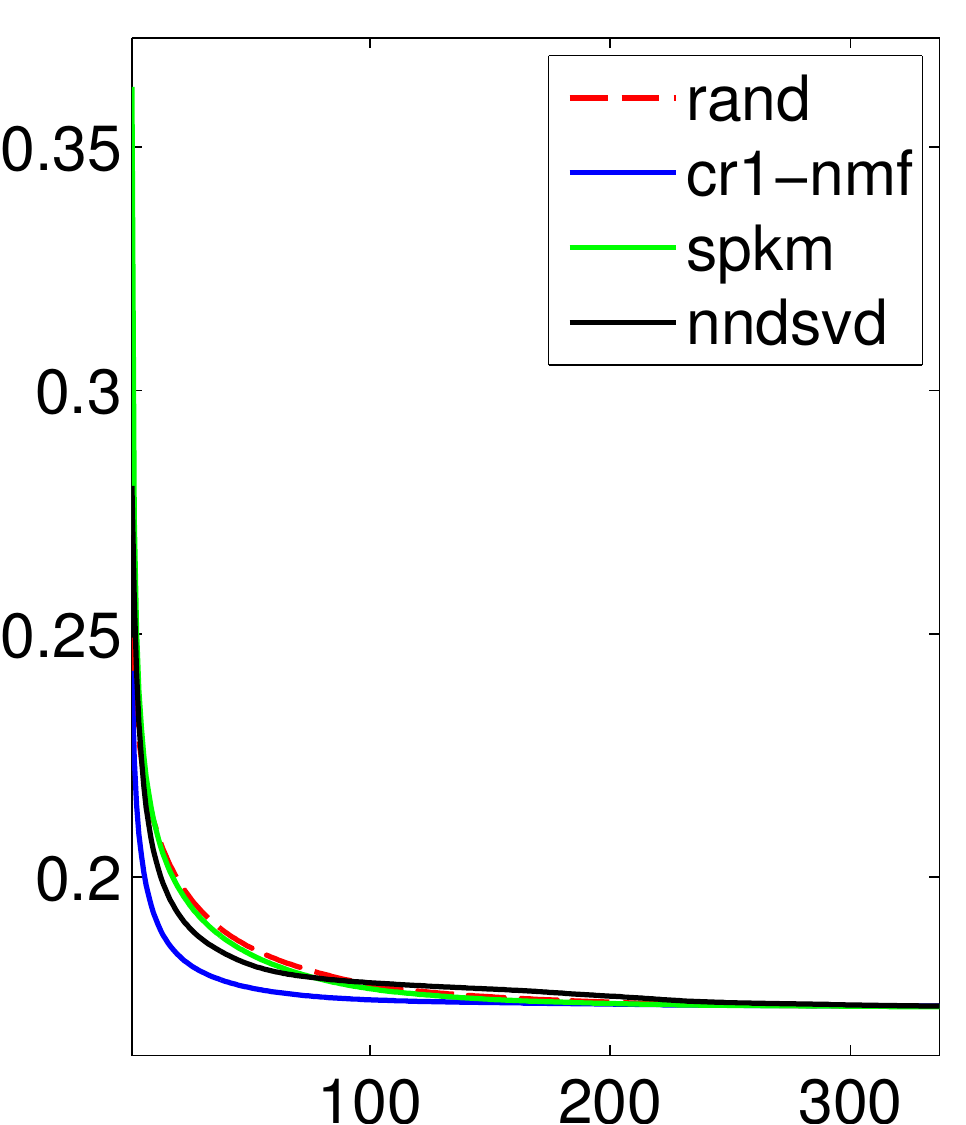}}\\
\subfloat{\includegraphics[width=.33\columnwidth,height=.37\columnwidth]{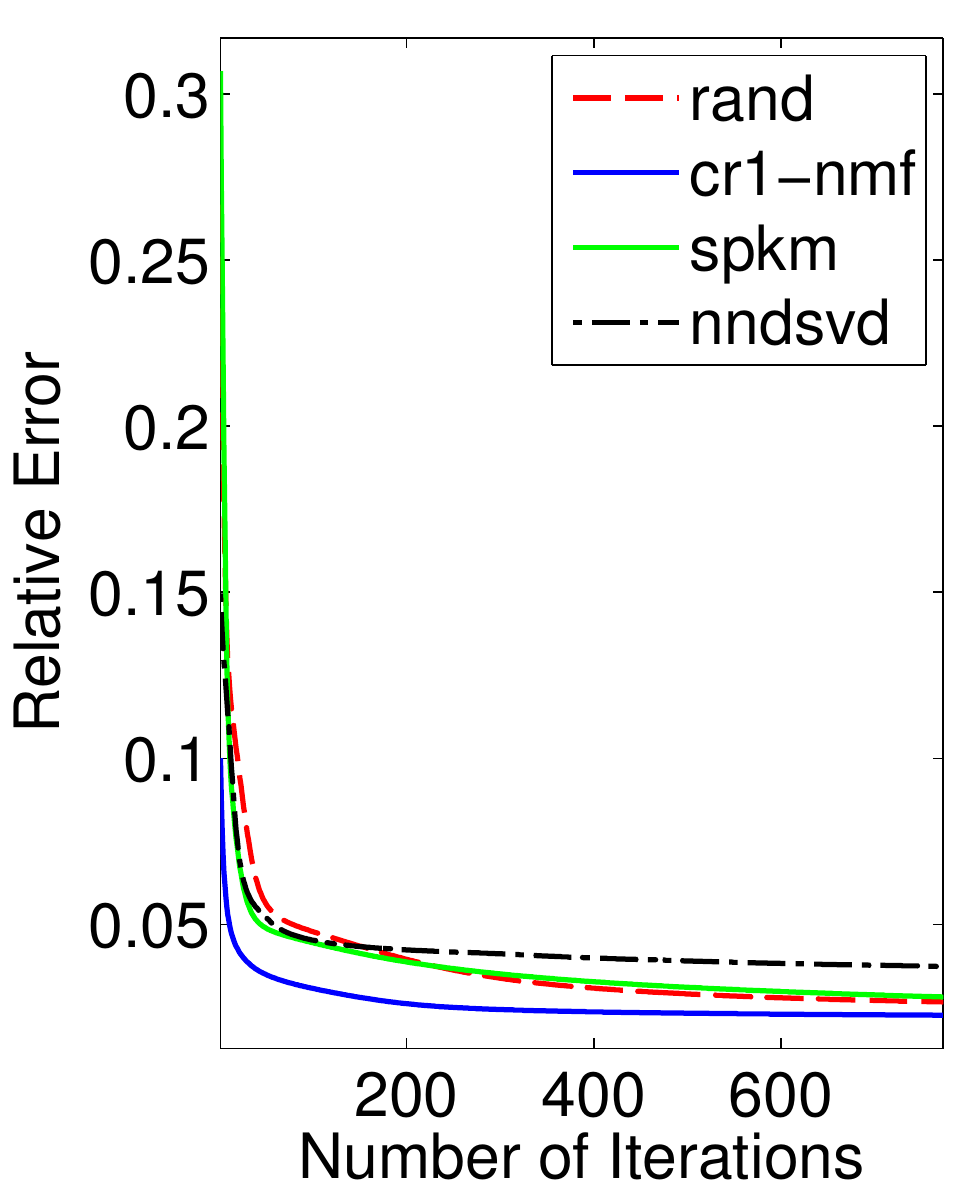}}\hfill
\subfloat{\includegraphics[width=.33\columnwidth,height=.37\columnwidth]{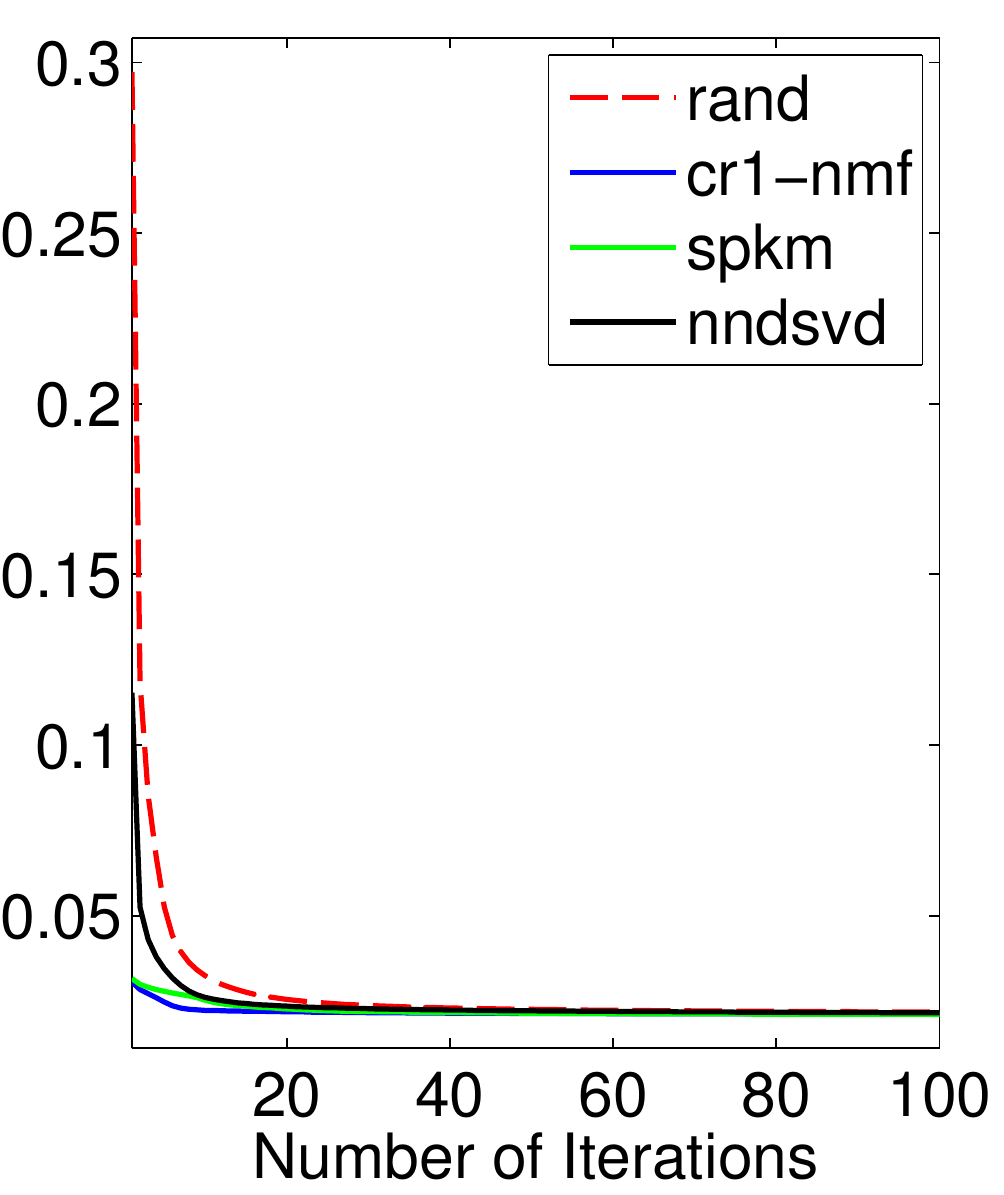}}\hfill
\subfloat{\includegraphics[width=.33\columnwidth,height=.37\columnwidth]{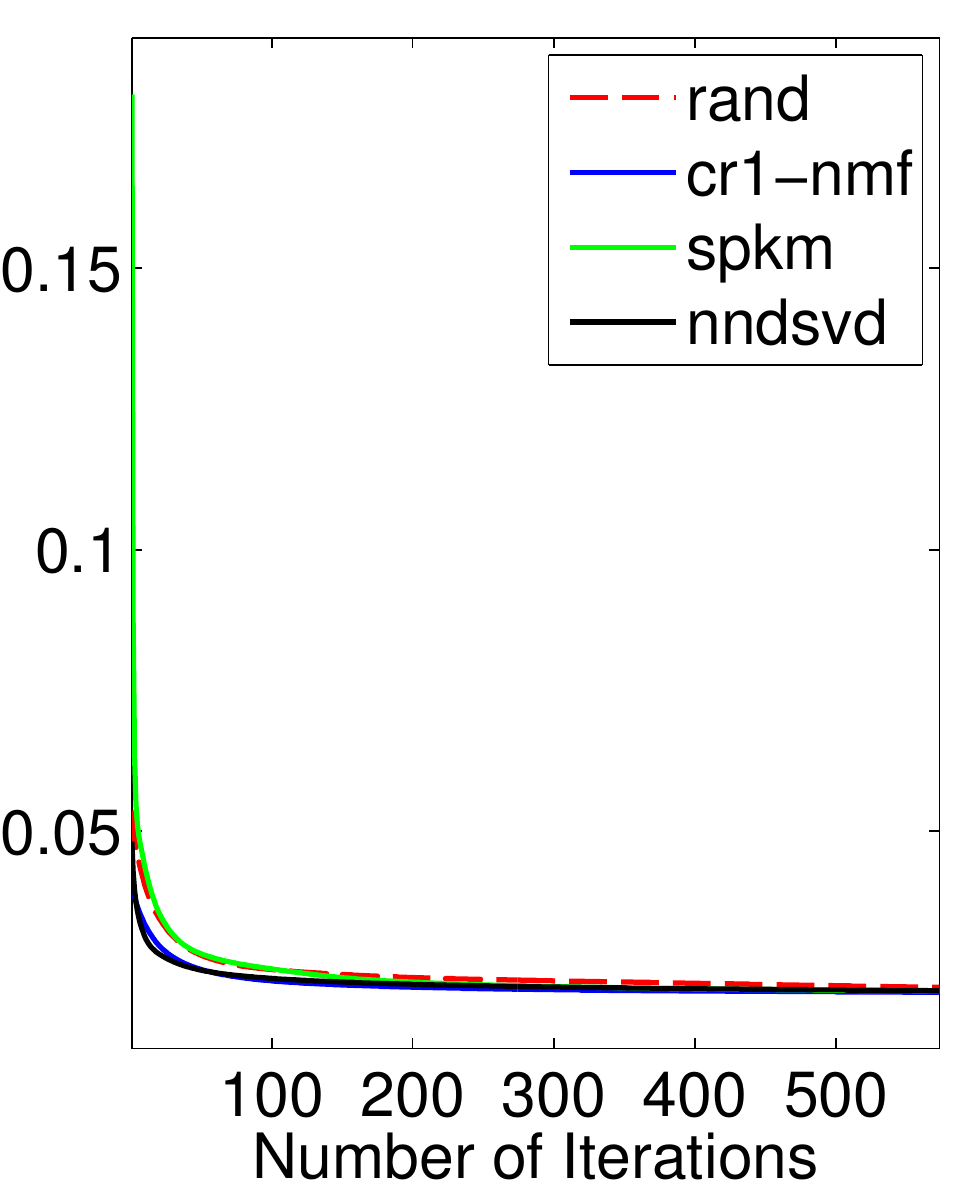}}\\
\caption{The first to fourth rows  are the numerical results for CK, faces94, Georgia Tech, and PaviaU datasets respectively. }\label{fig:init_rel}
\end{figure}

\begin{figure}[h]
\centering
\subfloat{\includegraphics[width=.3\columnwidth,height=.25\columnwidth]{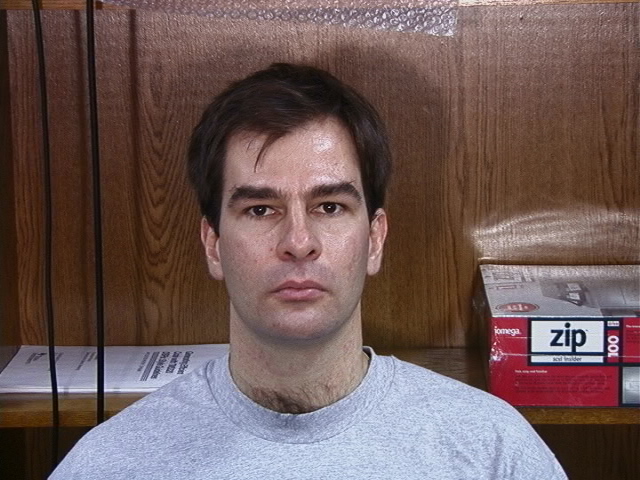}} \hspace{.1in}
\subfloat{\includegraphics[width=.3\columnwidth,height=.25\columnwidth]{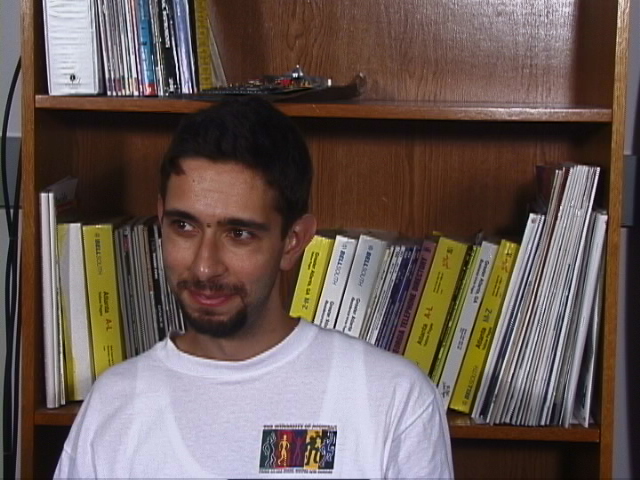}}  \hspace{.1in}
\subfloat{\includegraphics[width=.3\columnwidth,height=.25\columnwidth]{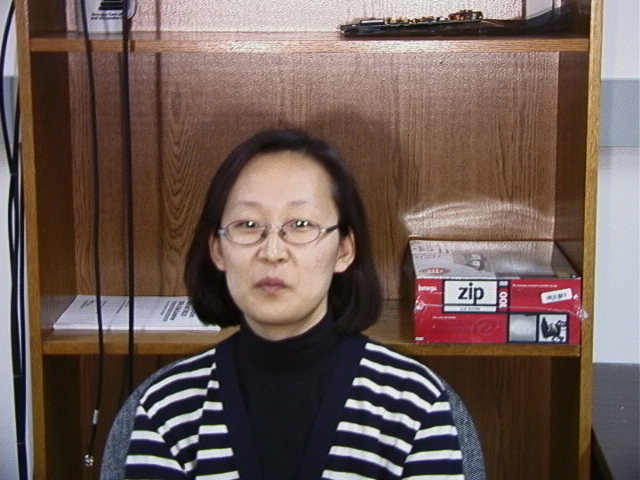}}\\

\subfloat{\includegraphics[width=.3\columnwidth,height=.25\columnwidth]{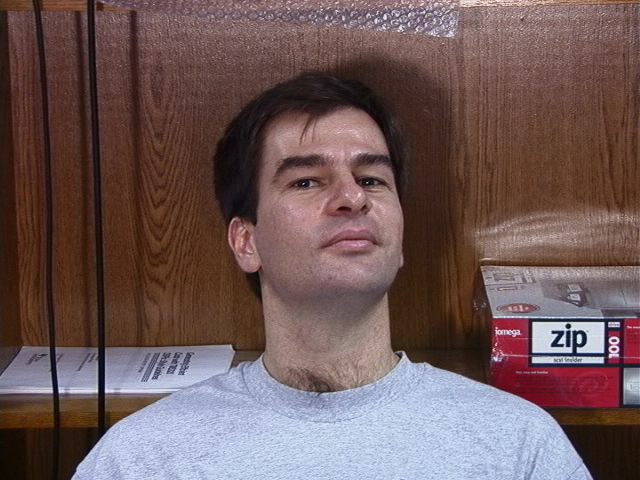}} \hspace{.1in}
\subfloat{\includegraphics[width=.3\columnwidth,height=.25\columnwidth]{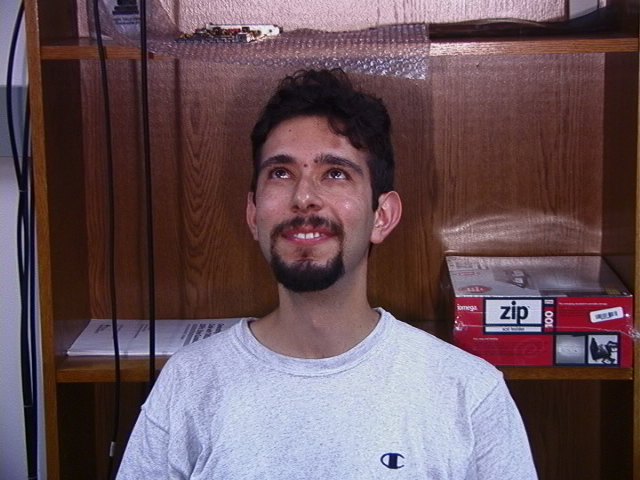}} \hspace{.1in}
\subfloat{\includegraphics[width=.3\columnwidth,height=.25\columnwidth]{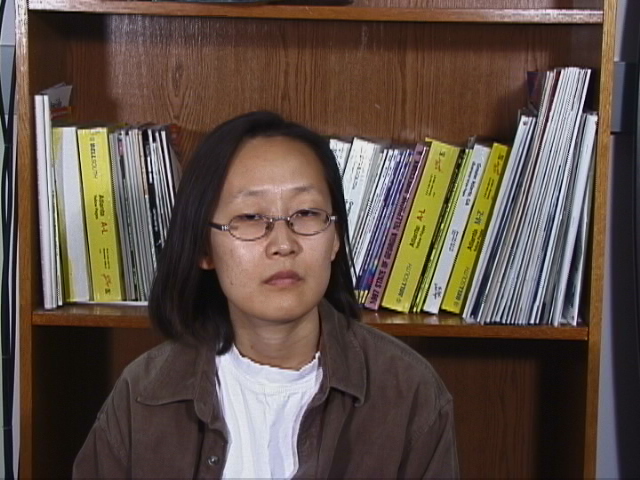}}\\

\subfloat{\includegraphics[width=.3\columnwidth,height=.25\columnwidth]{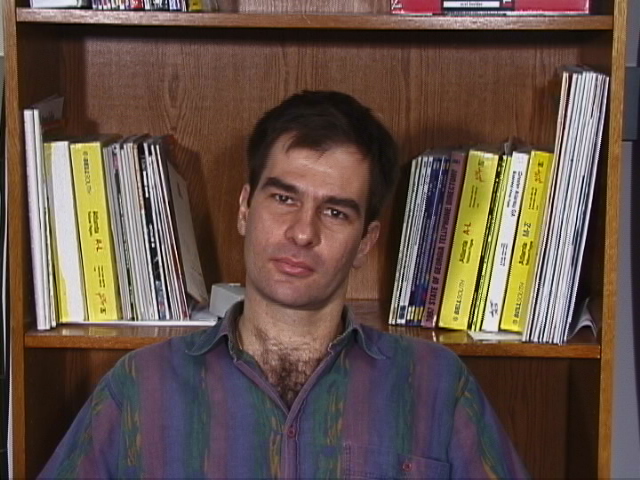}} \hspace{.1in}
\subfloat{\includegraphics[width=.3\columnwidth,height=.25\columnwidth]{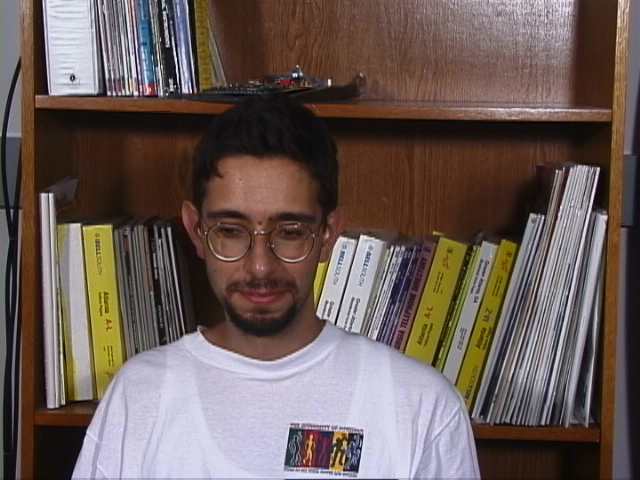}} \hspace{.1in}
\subfloat{\includegraphics[width=.3\columnwidth,height=.25\columnwidth]{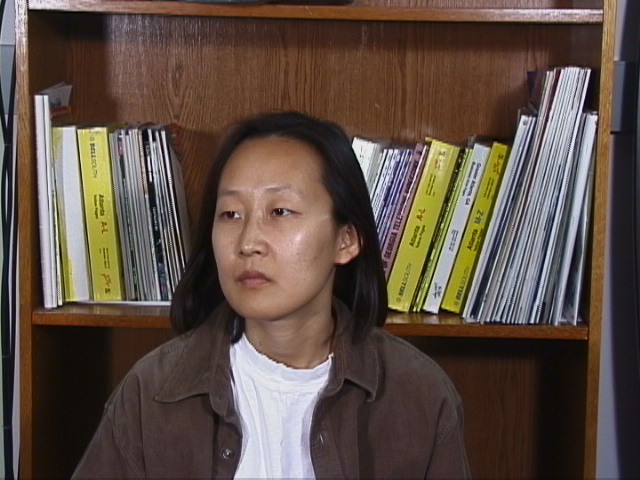}}\\
\caption{Images of 3 individuals in Georgia Tech dataset. }\label{fig: ground_truth_main}
\end{figure}

We observe from Figure \ref{fig:init_rel} that our algorithm almost always outperforms all other initialization approaches in terms of convergence speed  and/or the final relative error when combined with classical NMF algorithms for the selected real datasets (except that \texttt{nndsvd}$+$\texttt{hals} performs the best for PaviaU). In addition, we present the results from the Georgia Tech image dataset. For ease of illustration, we only display the results for $3$ individuals  (there are images for $50$ individuals in total) for the various initialization methods combined with \texttt{mult}. Several images of these $3$ individuals are presented in Figure~\ref{fig: ground_truth_main}. The basis images produced at the 20$^{\mathrm{th}}$ iteration are presented in Figure~\ref{fig: basis_images_20} (more basis images obtained at other iteration numbers are presented in the supplementary material). We observe from the basis images in Figure~\ref{fig: basis_images_20} that our initialization method is clearly superior to  \texttt{rand} and \texttt{nndsvd}. In the supplementary material, we additionally present %\blue{(Delete) two sets of figures: (i) basis images learned from a subset of the Georgia Tech dataset produced over time for various initialization methods combined with \texttt{mult} and (ii)} 
an illustration of Table~\ref{tab: nnlsb} as a figure where the horizontal and vertical axes are the running times (instead of number of iterations) and the relative errors respectively. These additional plots substantiate our conclusion that Algorithm \ref{algo:approx} serves as a good initializer for various other NMF algorithms. 

\begin{figure}[htp]
\subfloat{\includegraphics[width=.3\columnwidth,height=.25\columnwidth]{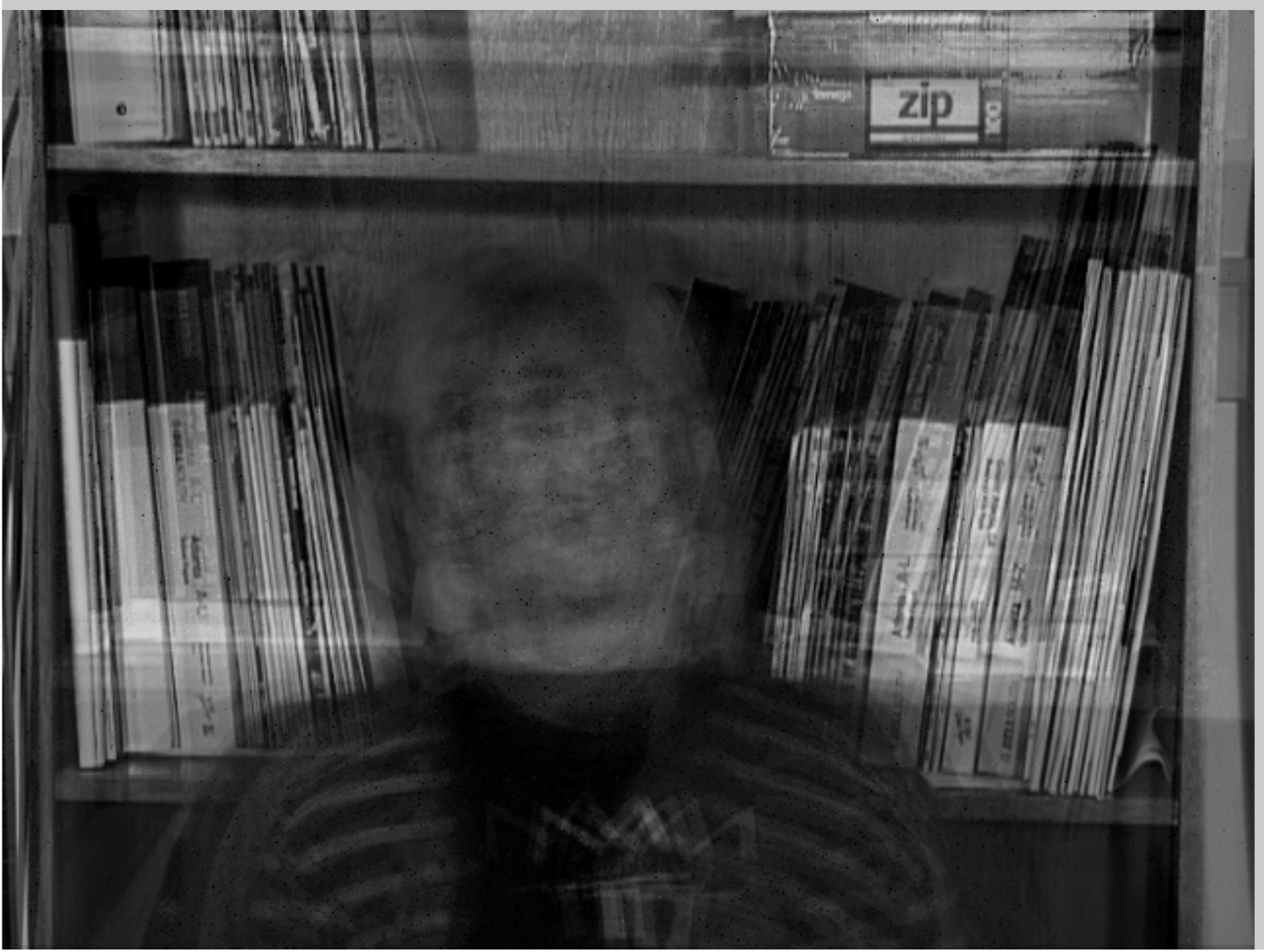}}\hfill
\subfloat{\includegraphics[width=.3\columnwidth,height=.25\columnwidth]{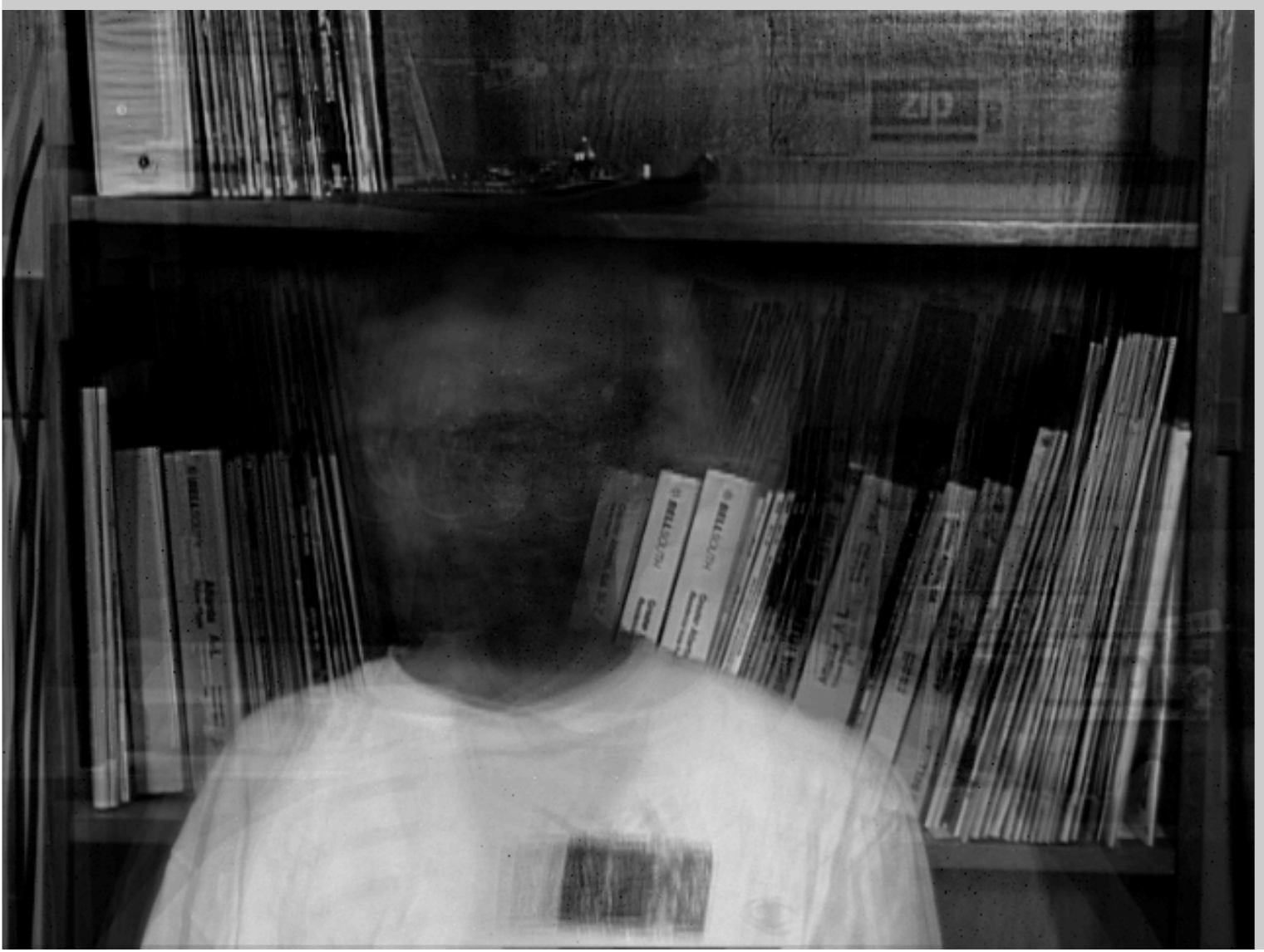}}\hfill
\subfloat{\includegraphics[width=.3\columnwidth,height=.25\columnwidth]{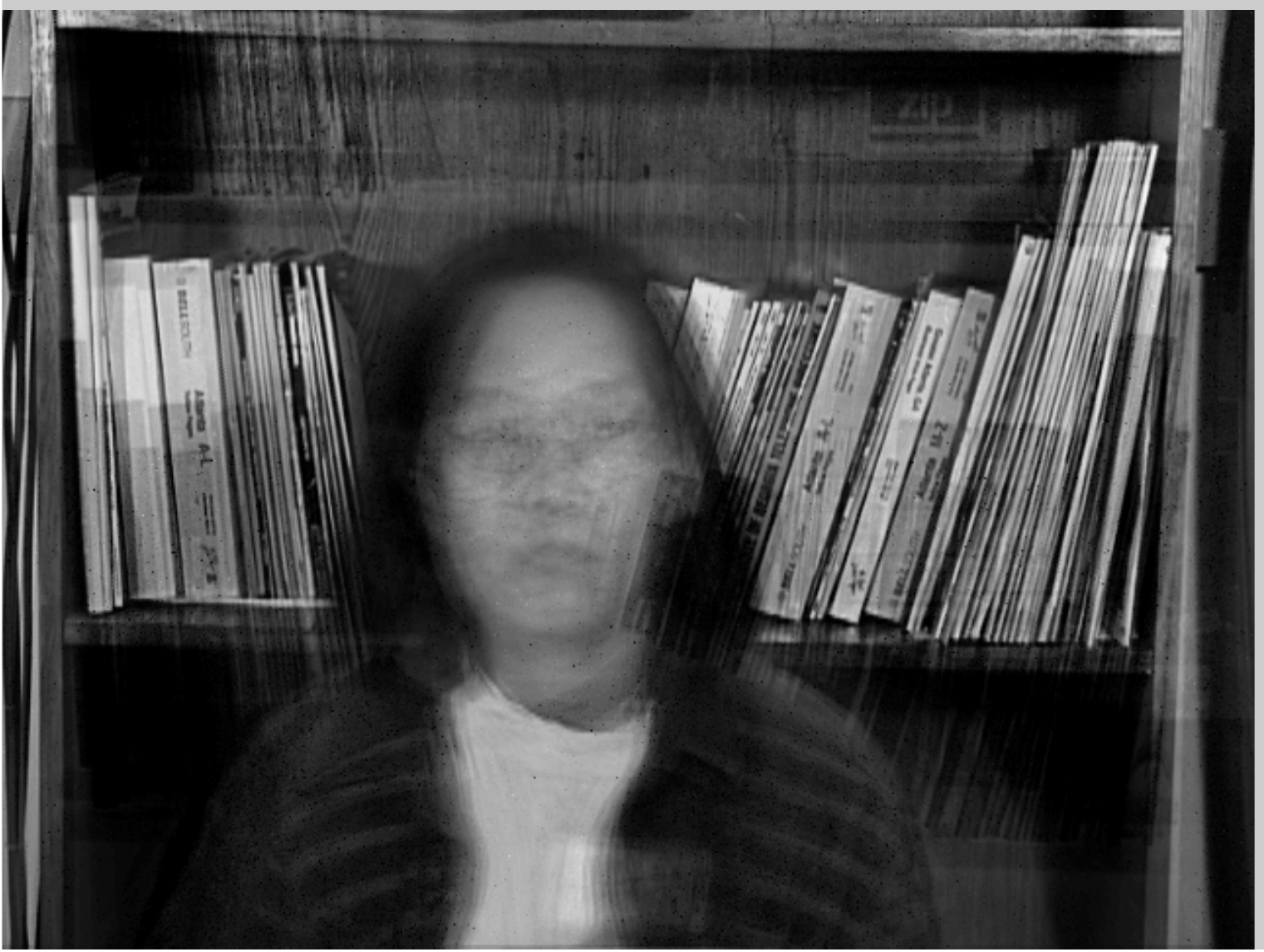}}\\

\subfloat{\includegraphics[width=.3\columnwidth,height=.25\columnwidth]{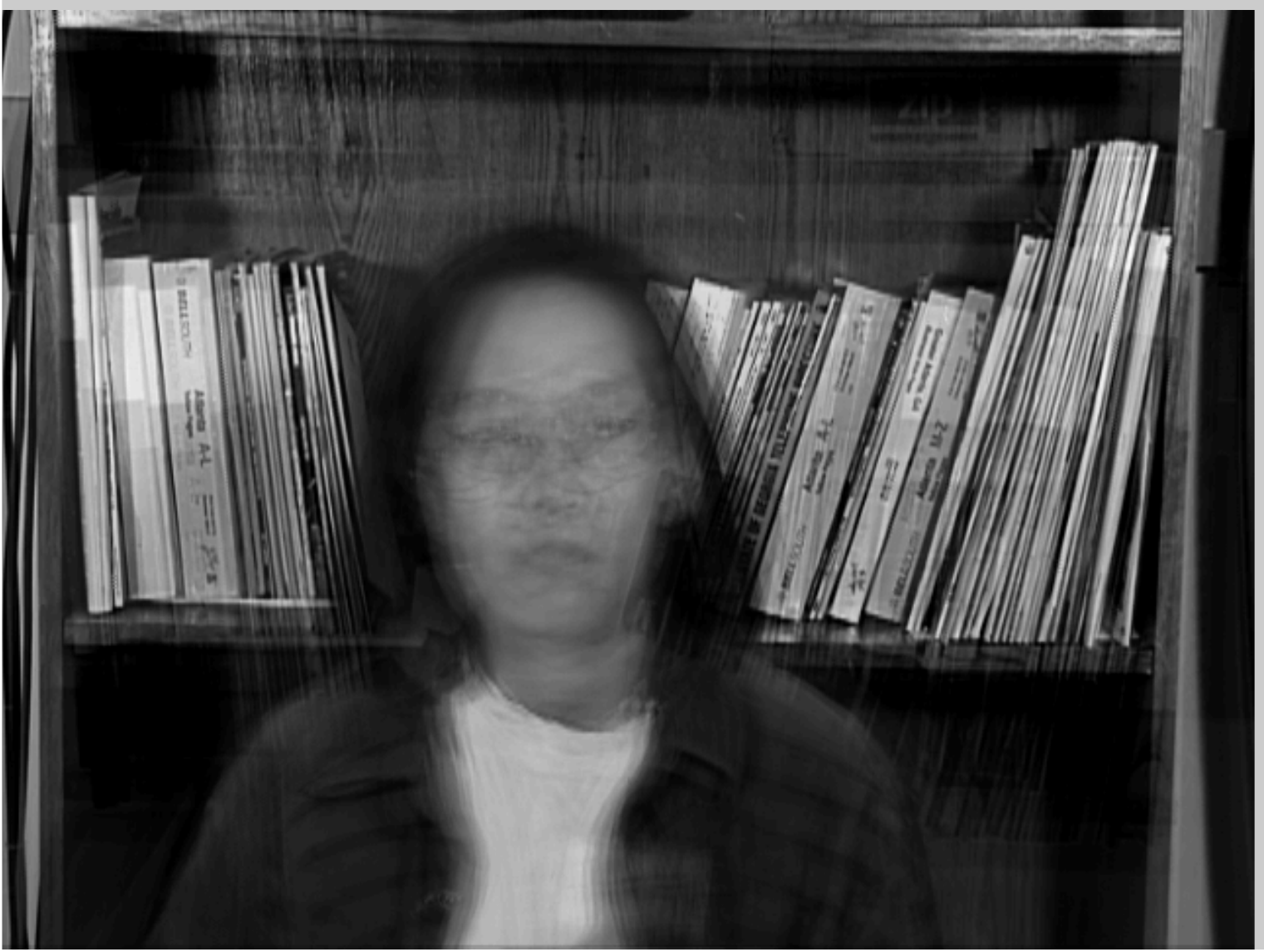}}\hfill
\subfloat{\includegraphics[width=.3\columnwidth,height=.25\columnwidth]{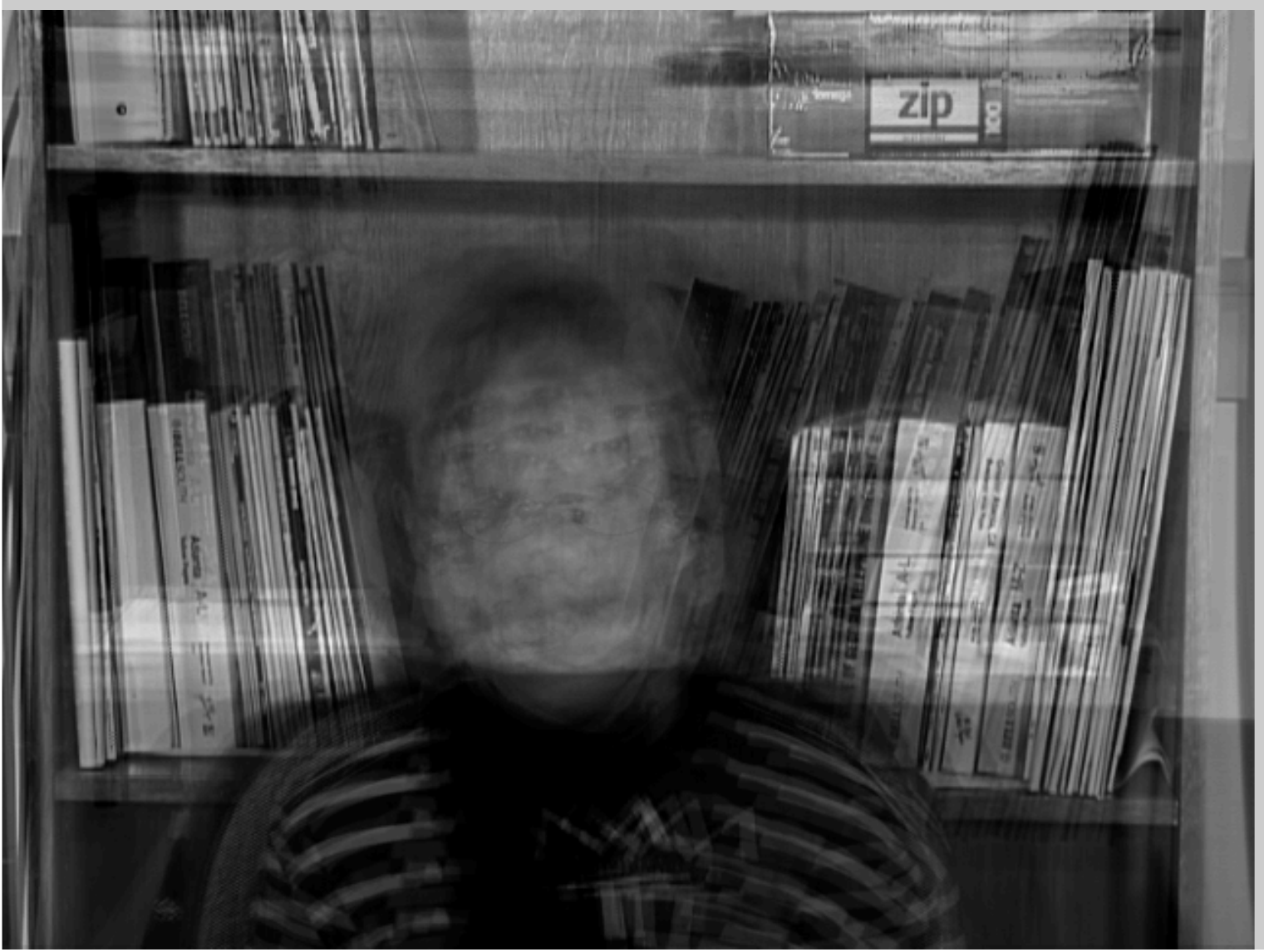}}\hfill
\subfloat{\includegraphics[width=.3\columnwidth,height=.25\columnwidth]{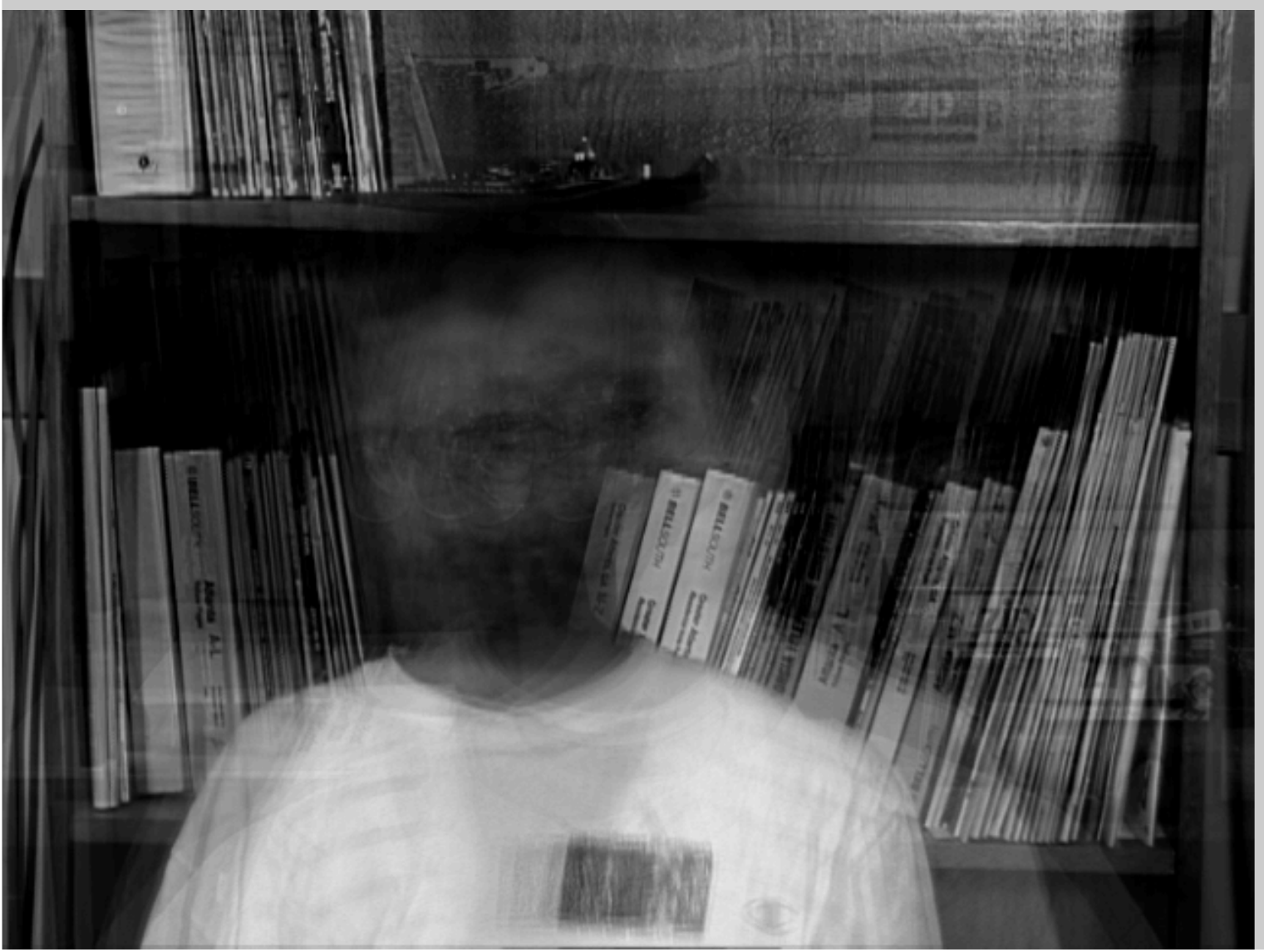}}\\

\subfloat{\includegraphics[width=.3\columnwidth,height=.25\columnwidth]{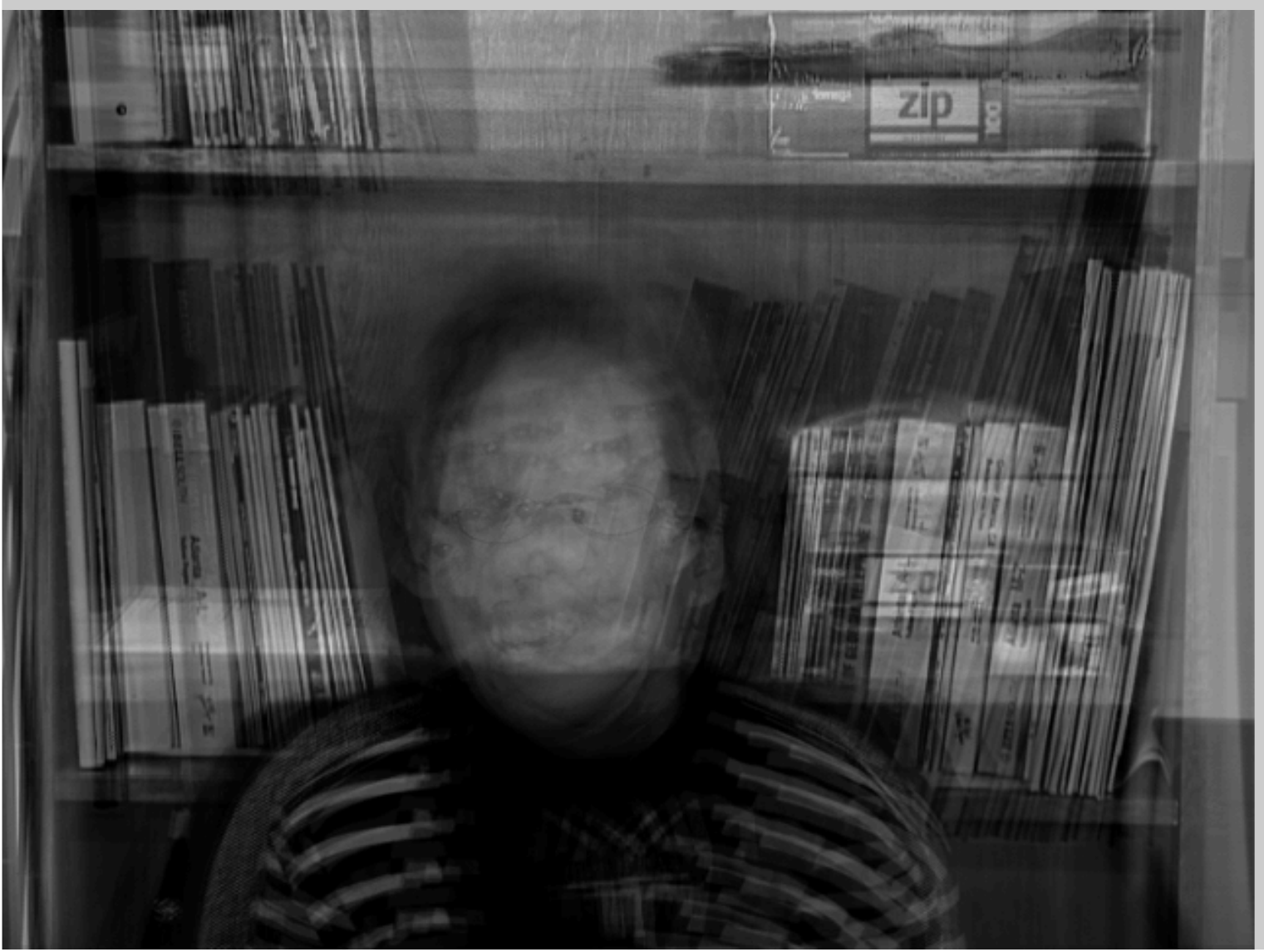}}\hfill
\subfloat{\includegraphics[width=.3\columnwidth,height=.25\columnwidth]{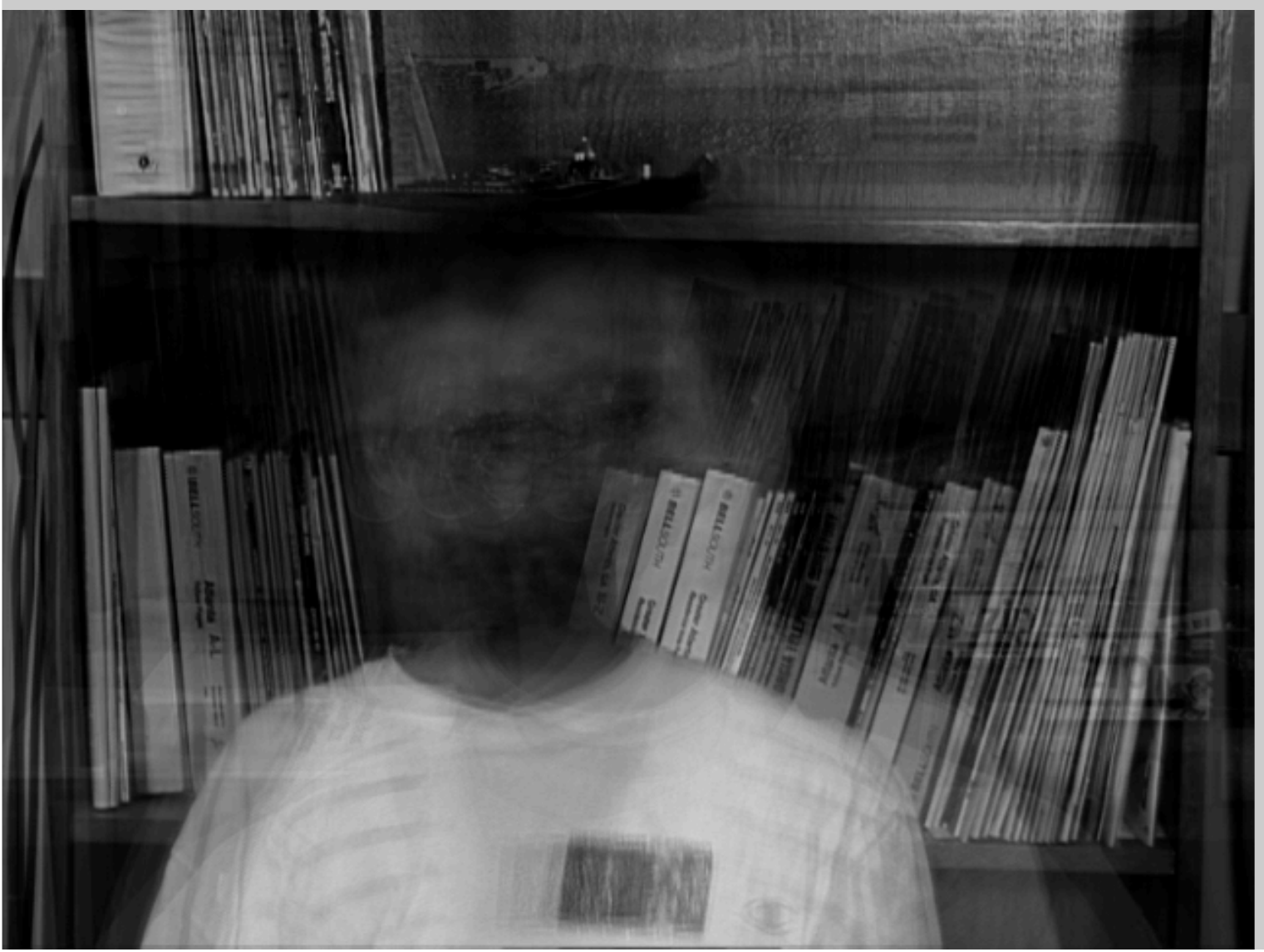}}\hfill
\subfloat{\includegraphics[width=.3\columnwidth,height=.25\columnwidth]{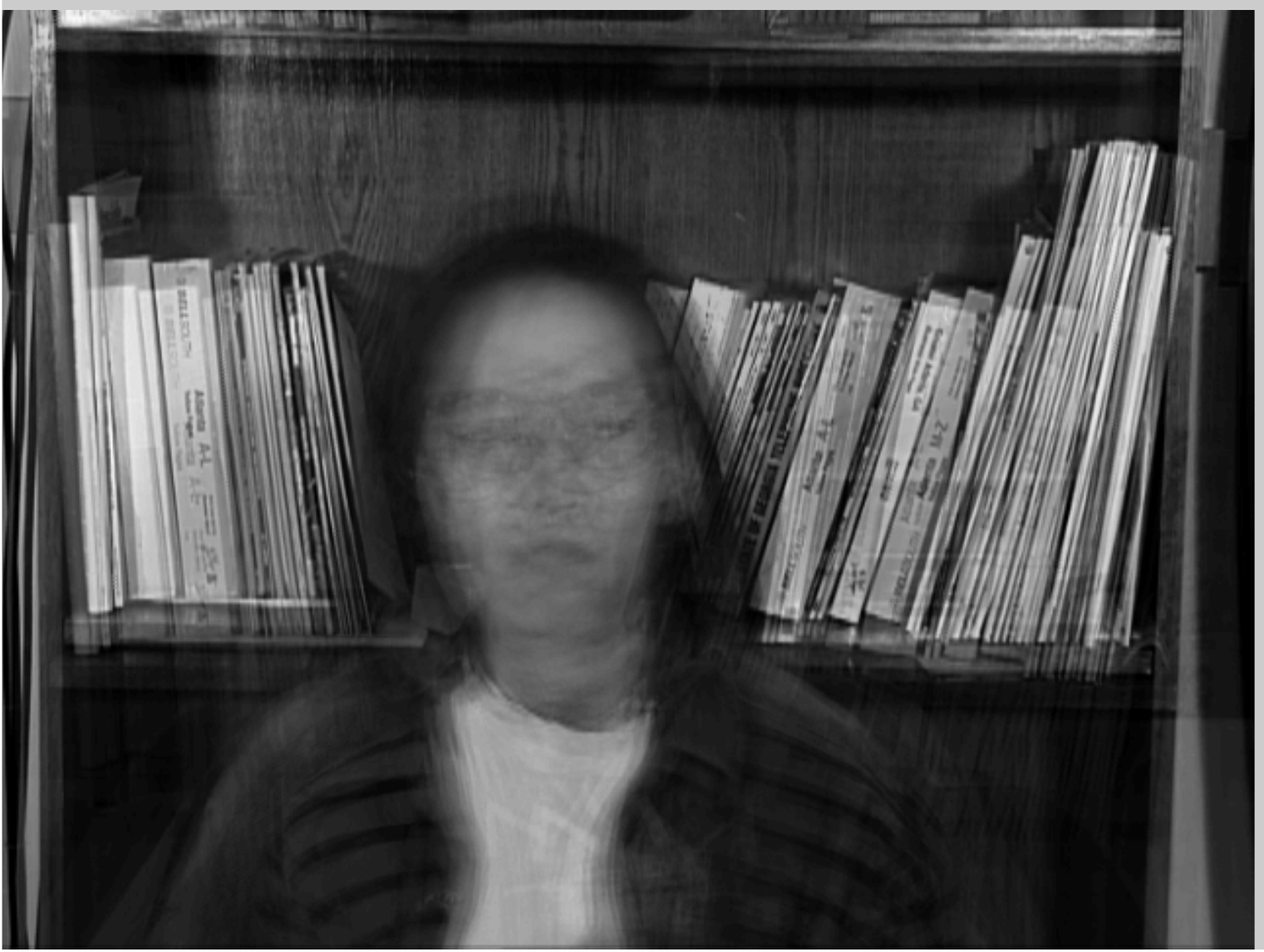}}\\

\subfloat{\includegraphics[width=.3\columnwidth,height=.25\columnwidth]{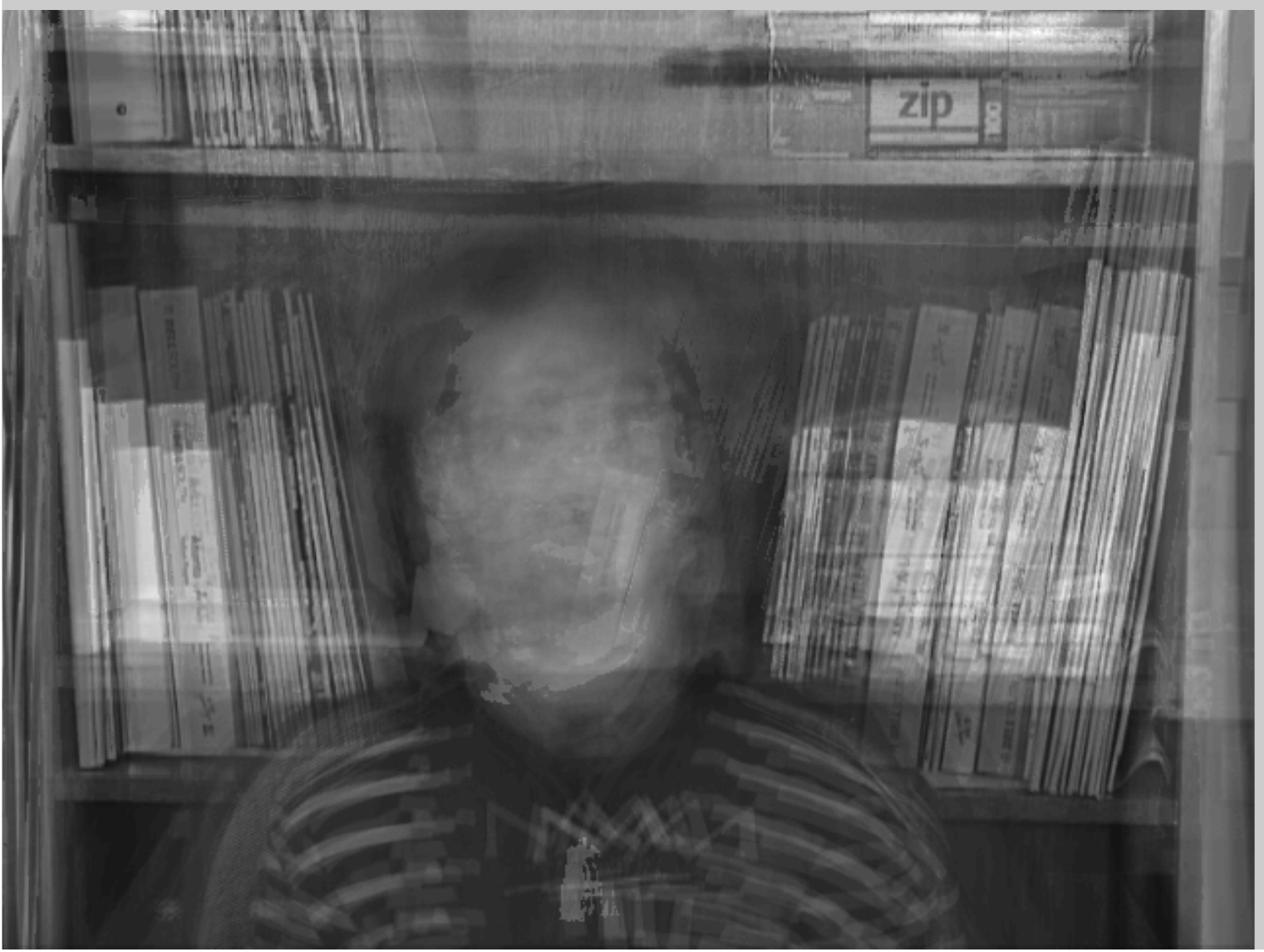}}\hfill
\subfloat{\includegraphics[width=.3\columnwidth,height=.25\columnwidth]{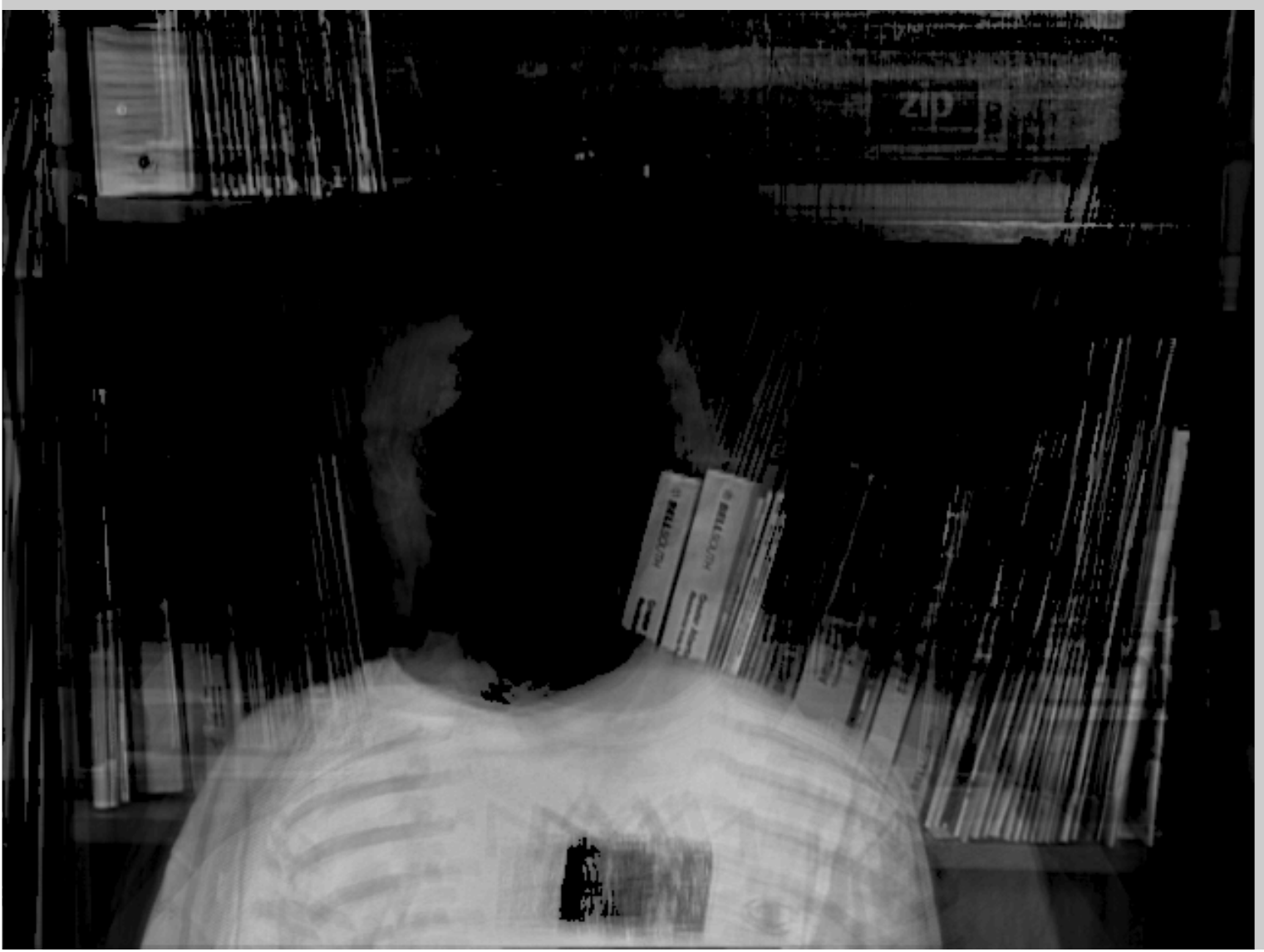}}\hfill
\subfloat{\includegraphics[width=.3\columnwidth,height=.25\columnwidth]{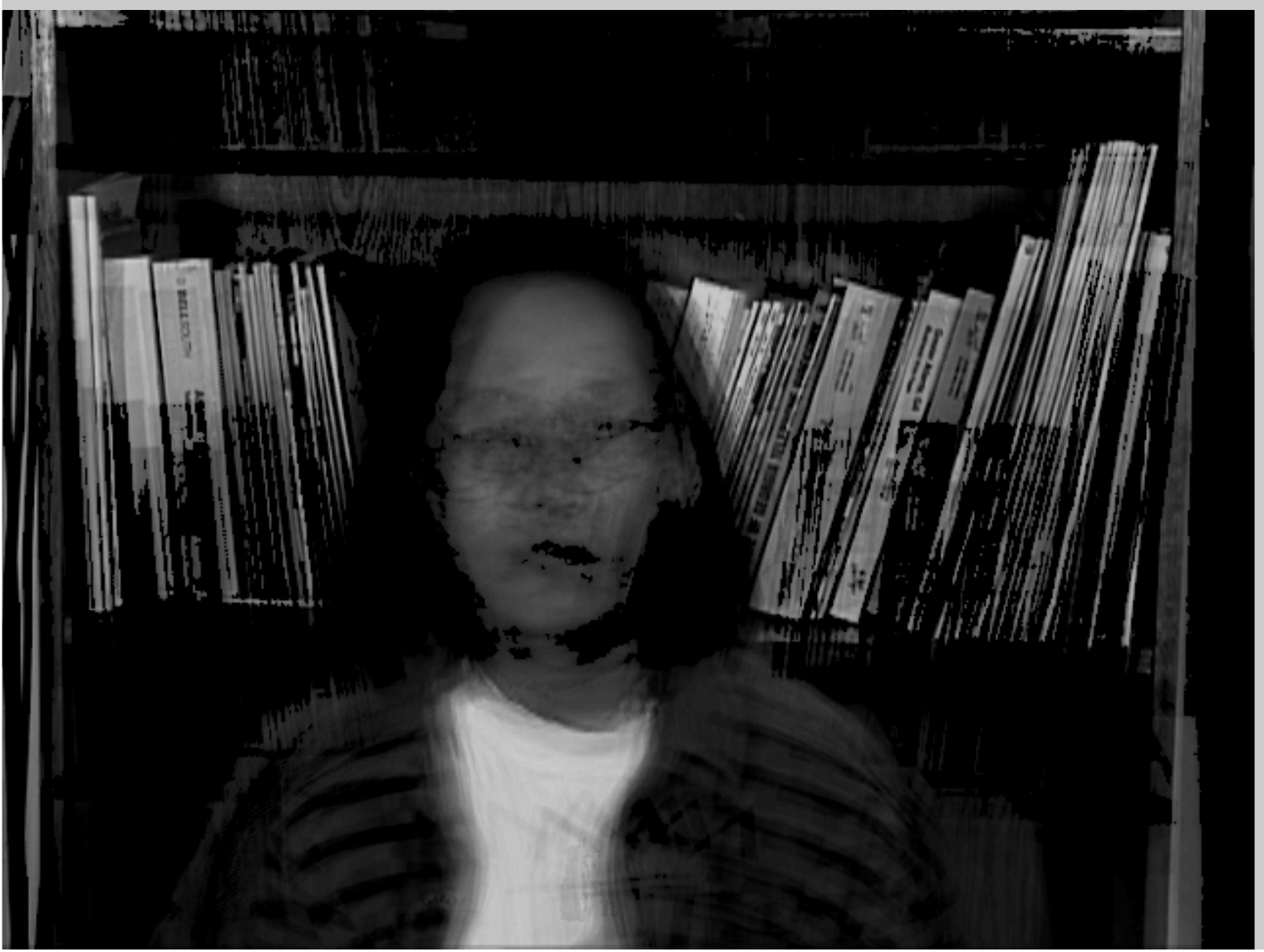}}\\

\caption{Basis images of 3 individuals in Georgia Tech dataset obtained at the $20^{\mathrm{th}}$ iteration. The first to fourth rows pertain to  \texttt{rand}, \texttt{cr1-nmf}, \texttt{spkm}, and \texttt{nndsvd} initializations respectively.} \label{fig: basis_images_20}
\end{figure}

\subsubsection{Intuition for the Advantages of  \texttt{cr1-nmf} over \texttt{spkm}  as an Initializer for NMF Algorithms} \label{sec:intuition}
%Now we provide some {\em intuition} for the effectiveness of \texttt{cr1-nmf} as an initializer for other NMF algorithms. 
From Figure \ref{fig:init_rel}, we see that the difference between the results obtained from using \texttt{spkm} as initialization method and the corresponding results obtained from using our initialization approach appears to be rather insignificant. However, from  Table \ref{tab: nnlsb}, which reports the running time to {\em first} achieve specified relative errors $\epsilon>0$ for the initialization methods combined with \texttt{nnlsb} (note that \texttt{nnlsb} only needs to use the initial left factor matrix, and thus we can compare the initial estimated basis vectors obtained by \texttt{spkm} and \texttt{cr1-nmf} directly), we see  that our initialization approach is clearly faster than \texttt{spkm}.% (when both methods are combined with \texttt{nnlsb}). 

%Why is this so?
 In addition, consider the scenario where there are duplicate or near-duplicate samples. Concretely, assume the data matrix   $\bV := \begin{bmatrix} 1 & 1 & 0 \\ 0 & 0 & 1 \end{bmatrix} \in \mathbb{R}^{2\times 3}_{+}$ and $K=1$. Then   the left factor matrix produced by rank-one NMF is $\bw = [1;0]$ and the normalized mean vector (centroid for \texttt{spkm}) is $\bar{\bu}:= [\frac{2}{\sqrt{5}};\frac{1}{\sqrt{5}}]$. The approximation error w.r.t.\  $\bw$ is $\|\bV - \bw \bw^T \bV\|_{\mathrm{F}} = 1$, while the approximation error w.r.t.\  $\bar{\bu}$ is $\|\bV - \bar{\bu} \bar{\bu}^T \bV\|_{\mathrm{F}} \approx 1.0954$.  Note that \texttt{spkm} is more constrained since it implicitly outputs a binary right factor matrix $\bH \in\{0,1\}^{K\times N}$ while rank-one NMF (cf.\ Lemma \ref{lem:rank_one}) does not impose this stringent requirement. Hence \texttt{cr1-nmf} generally leads to a smaller relative error compared to \texttt{spkm}. 

\subsubsection{Initialization Performance in Terms of Clustering} \label{sec: init_cluster}
We now compare    clustering performances using various initialization methods. To obtain a comprehensive evaluation, we use three widely-used evaluation metrics, namely, the normalized mutual information~\cite{Alex02} (nmi), the Dice coefficient \cite{Salton89} (Dice) and the purity~\cite{Manning08,Li13}. The clustering results for the CK and tr11\footnote{The tr11 dataset can be found at \url{http://glaros.dtc.umn.edu/gkhome/fetch/sw/cluto/datasets.tar.gz}. It is a canonical  example of a text dataset  and contains $6429$ terms and $414$ documents. The   number of clusters/topics is $K=9$.} datasets are shown in Tables~\ref{tab: clustering1} and~\ref{tab: clustering2} respectively. 
%and the sparsity of it is 95.62\%, i.e., 95.62\% entries of its term-document matrix are zero. We do not consider relative error for text datasets because the relative error is close to one even when approximate convergence is achieved. } 
Clustering results for other datasets   are shown in the supplementary material (for space considerations). We run the standard \texttt{k-means} and \texttt{spkm} clustering algorithms for at most   $1000$ iterations and terminate the algorithm if the cluster  memberships do not change.  All the classical NMF algorithms are terminated if the variation of the product of factor matrices is small over $10$ iterations. Note that \texttt{nndsvd} is a deterministic initialization method, so its clustering results are the same across different runs. We observe  from  Tables~\ref{tab: clustering1} and~\ref{tab: clustering2} and those in the supplementary material that our initialization approach almost always outperforms all others  (under   all  the three evaluation metrics). % for the real datasets under consideration.

\section{Conclusion and Future Work}
\subsection{Summary of Contributions}
We proposed a new geometric assumption for the purpose of performing NMF. In contrast to the separability condition~\cite{Donoho_04,Arora_12,Bittorf_12}, under our geometric assumption, we are able to prove several novel deterministic and probabilistic results concerning the relative errors of learning the factor matrices. We are also able to provide a theoretically-grounded method of choosing the number of clusters (i.e., the number of circular cones) $K$. We showed experimentally on synthetic datasets that satisfy the geometric assumption  that our algorithm performs exceedingly well in terms of accuracy and speed. Our method also serves a fast and effective  initializer for running NMF on real datasets. Finally, it outperforms other competing methods on various clustering tasks. 
\subsection{Future Work and Open Problems }\label{sec:open}
We plan to explore the following extensions.
\begin{enumerate}
\item First, we hope to prove theoretical guarantees for the scenario when $\bV$ only satisfies an {\em approximate version} of the geometric assumption, i.e., we only have access to $\hat{\bV}:=[ \bV +\delta \bE]_{+}$ (cf.~Section~\ref{sec:estK}) where $\delta\approx 0$.
\item Second, here we focused on upper bounds on the relative error. To assess the tightness of these bounds,  we hope to prove {\em minimax lower} bounds on the relative error similarly to Jung et al.~\cite{Jung}. 
 \item Third, as   mentioned in Section~\ref{sec:thm_contribution}, our geometric assumption in \eqref{eq:geo_ass} can be considered as a special case of the near-separability assumption for NMF~\cite{Donoho_04}. To the best of our knowledge,  there is no theoretical guarantee for the relative error under the near-separability assumption.
 
 \item For large-scale data, it is often desirable to perform NMF in an {\em online} fashion~\cite{ZhaoTan17, ZhaoTanXu16}, i.e., each data point $\bv_n$ arrives in a sequential manner. We would like to develop online versions of the algorithm herein.
 
 \item It would be fruitful to leverage the theoretical results for $k$-means++ \cite{arthur2007} to provide guarantees for a probabilistic version of   our initialization method.  Note that our method is deterministic while $k$-means++ is probabilistic, so a probabilistic variant of  Algorithm~\ref{algo:approx} may have to be developed for fair comparisons with $k$-means++.
 
 \item We may also extend our Theorem~\ref{thm:deterK} to near-separable data matrices, possibly with additional assumptions. 
\end{enumerate}

\begin{table}[t]
\centering
\caption{Clustering performances for initialization methods combined with classical NMF algorithms for the CK dataset}\label{tab: clustering1}
 \begin{tabular}{|c|c|c|c|} 
\hline
			& nmi 				& Dice 				& purity 			\\ \hline
\texttt{k-means} 		& $0.941 \!\pm\! 0.008$  		& $0.773 \!\pm\! 0.030$  		& $0.821 \!\pm\! 0.023$ 		\\ \hline   
\texttt{spkm} 			& $0.940 \!\pm\! 0.010$  		& $0.765 \!\pm\! 0.036$  		& $0.815 \!\pm\! 0.031$ 		\\ \hline\hline
\texttt{rand}$+$\texttt{mult} 		& $0.919 \!\pm\! 0.009$  		& $0.722 \!\pm\! 0.026$  		& $0.753 \!\pm\! 0.025$ 		\\ \hline      
\texttt{cr1-nmf}$+$\texttt{mult}	& $\mathbf{0.987} \!\pm\! 0.002$  	& $\mathbf{0.944} \!\pm\! 0.006$  	& $\mathbf{0.961} \!\pm\! 0.006$	\\ \hline
\texttt{spkm}$+$\texttt{mult}		& $0.969 \!\pm\! 0.005$  		& $0.875 \!\pm\! 0.020$  		& $0.911 \!\pm\! 0.018$ 		\\ \hline
\texttt{nndsvd}$+$\texttt{mult} 	& $0.870 \!\pm\! 0.000$ 		& $0.614 \!\pm\! 0.000$  		& $0.619 \!\pm\! 0.000$		\\ \hline\hline
\texttt{rand}$+$\texttt{nnlsb} 		& $0.918 \!\pm\! 0.011$  		& $0.727 \!\pm\! 0.026$  		& $0.756 \!\pm\! 0.027$ 		\\ \hline      
\texttt{cr1-nmf}$+$\texttt{nnlsb}	& $\mathbf{0.986} \!\pm\! 0.003$  	& $\mathbf{0.940} \!\pm\! 0.011$  	& $\mathbf{0.959} \!\pm\! 0.010$	\\ \hline
\texttt{spkm}$+$\texttt{nnlsb}		& $0.984 \!\pm\! 0.004$  		& $0.929 \!\pm\! 0.014$  		& $0.956 \!\pm\! 0.012$ 		\\ \hline
\texttt{nndsvd}$+$\texttt{nnlsb}  	& $0.899 \!\pm\! 0.000$ 		& $0.688 \!\pm\! 0.000$  		& $0.724 \!\pm\! 0.000$		\\ \hline\hline
\texttt{rand}$+$\texttt{hals} 		& $0.956 \!\pm\! 0.007$  		& $0.826 \!\pm\! 0.017$  		& $0.859 \!\pm\! 0.022$ 		\\ \hline      
\texttt{cr1-nmf}$+$\texttt{hals}	& $\mathbf{0.974} \!\pm\! 0.006$  	& $\mathbf{0.889} \!\pm\! 0.015$  	& $\mathbf{0.925} \!\pm\! 0.016$	\\ \hline
\texttt{spkm}$+$\texttt{hals}		& $0.964 \!\pm\! 0.005$  		& $0.854 \!\pm\! 0.015$  		& $0.885 \!\pm\! 0.020$ 		\\ \hline
\texttt{nndsvd}$+$\texttt{hals}  	& $0.942 \!\pm\! 0.000$ 		& $0.786 \!\pm\! 0.000$  		& $0.830 \!\pm\! 0.000$		\\ \hline
\end{tabular} %\vspace{-.1in}
\end{table} 
\begin{table}[t]
\centering
\caption{Clustering performances for initialization methods combined with classical NMF algorithms for the  tr11 dataset}\label{tab: clustering2}
 \begin{tabular}{|c|c|c|c|} 
\hline
			& nmi 				& Dice 				& purity 			\\ \hline
\texttt{k-means} 		& $0.520 \!\pm\! 0.061$  		& $0.470 \!\pm\! 0.042$  		& $0.673 \!\pm\! 0.059$ 		\\ \hline   
\texttt{spkm} 			& $0.504 \!\pm\! 0.103$  		& $0.454 \!\pm\! 0.085$  		& $0.664 \!\pm\! 0.091$ 		\\ \hline\hline
\texttt{rand}$+$\texttt{mult} 		& $0.595 \!\pm\! 0.040$  		& $0.540 \!\pm\! 0.050$  		& $0.764 \!\pm\! 0.025$ 		\\ \hline      
\texttt{cr1-nmf}$+$\texttt{mult}	& $\mathbf{0.649} \!\pm\! 0.049$  	& $\mathbf{0.610} \!\pm\! 0.052$  	& $\mathbf{0.791} \!\pm\! 0.023$	\\ \hline
\texttt{spkm}$+$\texttt{mult}		& $0.608 \!\pm\! 0.052$  		& $0.550 \!\pm\! 0.061$  		& $0.773 \!\pm\! 0.031$ 		\\ \hline
\texttt{nndsvd}$+$\texttt{mult} 	& $0.580 \!\pm\! 0.000$ 		& $0.515 \!\pm\! 0.000$  		& $0.761 \!\pm\! 0.000$		\\ \hline\hline
\texttt{rand}$+$\texttt{nnlsb} 		& $0.597 \!\pm\! 0.030$  		& $0.537 \!\pm\! 0.040$  		& $0.765 \!\pm\! 0.018$ 		\\ \hline      
\texttt{cr1-nmf}$+$\texttt{nnlsb}	& $\mathbf{0.655} \!\pm\! 0.046$  	& $\mathbf{0.615} \!\pm\! 0.050$  	& $\mathbf{0.794} \!\pm\! 0.023$	\\ \hline
\texttt{spkm}$+$\texttt{nnlsb}		& $0.618 \!\pm\! 0.052$  		& $0.563 \!\pm\! 0.065$  		& $0.776 \!\pm\! 0.027$ 		\\ \hline
\texttt{nndsvd}$+$\texttt{nnlsb}  	& $0.585 \!\pm\! 0.000$ 		& $0.512 \!\pm\! 0.000$  		& $0.766 \!\pm\! 0.000$		\\ \hline\hline
\texttt{rand}$+$\texttt{hals} 		& $0.609 \!\pm\! 0.044$  		& $0.555 \!\pm\! 0.056$  		& $0.772 \!\pm\! 0.024$ 		\\ \hline      
\texttt{cr1-nmf}$+$\texttt{hals}	& $\mathbf{0.621} \!\pm\! 0.052$  	& $\mathbf{0.580} \!\pm\! 0.062$  	& $\mathbf{0.778} \!\pm\! 0.026$	\\ \hline
\texttt{spkm}$+$\texttt{hals}		& $0.619 \!\pm\! 0.052$  		& $0.567 \!\pm\! 0.061$  		& $0.776 \!\pm\! 0.027$ 		\\ \hline
\texttt{nndsvd}$+$\texttt{hals}  	& $0.583 \!\pm\! 0.000$ 		& $0.511 \!\pm\! 0.000$  		& $0.768 \!\pm\! 0.000$		\\ \hline
\end{tabular} %\vspace{-.1in}
\end{table}

\appendices
\section{Proof of Theorem \ref{thm:prob_ext}}\label{app:prf_prob}
To prove Theorem \ref{thm:prob_ext}, we first provide a few definitions and lemmas. Consider the following  condition that ensures that the circular cone $C(\bu,\alpha)$ is  entirely contained in the non-negative orthant~$\mathcal{P}$.
\begin{lemma}\label{lem:cc_nn}
If $\bu = (u(1),u(2),\ldots, u(F))$ is a positive unit vector and $\alpha>0$ satisfies
\begin{equation}\label{eq:cc_nn}
  \alpha \leq \arccos \sqrt{1-u_{\mathrm{min}}^{2}}, 
\end{equation}
where $u_{\mathrm{min}}:=\min_{f}u(f)$, then $\mathcal{C}(\bu,\alpha) \subseteq \mathcal{P}$.% (i.e., it is entirely contained in the nonnegative orthant). 
\end{lemma}
\begin{IEEEproof}[Proof of Lemma \ref{lem:cc_nn}] %\ref{lem:cc_nn}}\\
Because any nonnegative vector $\bx$ is spanned by basis vectors $\be_{1},\be_{2},\ldots,\be_{F}$, given a positive unit vector $\bu$, to find the largest size angle, we only need to consider the angle between $\bu$ and $\be_{f}, f \in [F]$. Take any $f \in [F]$, if the angle $\beta$ between $\bu$ and $\be_{f}$ is not larger than $\pi/4$, we can obtain the unit vector symmetric to $\be_{f}$ w.r.t.\  $\bu$ in the plane spanned by $\bu$ and $\be_{f}$ is also nonnegative. In fact, the vector is $2(\cos\beta)\bu -\be_{f}$. Because $u(f)=\cos\beta$ and $\beta \leq \pi/4$, we have $2\cos^2 \beta \geq 1$ and the vector is nonnegative. If $\beta > \pi/4$, i.e., $u(f) < 1/\sqrt{2} $, we can take the extreme nonnegative unit vector $\bz$ in the span of $\bu$ and $\be_{f}$, i.e., 
\begin{equation}
  \bz=\frac{\bu-u(f)\be_{f}}{\|\bu-u(f)\be_{f}\|_{2}},
\end{equation}
and it is easy to see $\bu^{T}\bz=\sqrt{1-u(f)^{2}}$. Hence the angle between $\bz$ and $\bu$ is $\pi/2-\beta < \pi/4$. Therefore, the largest size angle $\alpha_{\be_f}$ w.r.t.\  $\be_{f}$ is 
\begin{equation}
\alpha_{\be_f} :=  \left\{
\begin{array}{cc}
\arccos u(f),  & \mbox{if}\;u(f) \geq 1/\sqrt{2} \\
\arccos \sqrt{1-u(f)^{2}},   & \mbox{if}\;u(f) < 1/\sqrt{2} 
\end{array} \right.
\end{equation}
or equivalently, $\alpha_{\be_f}=\min\{\arccos u(f), \arccos \sqrt{1-u(f)^{2}}\}$. Thus, the largest size angle corresponding to $\bu$ is 
\begin{equation}
  \min_{f}   \big\{\min\{\arccos u(f), \arccos \sqrt{1-u(f)^{2}}\} \big\}
\end{equation}
Let $u_{\mathrm{max}}:=\max_{f} u(f)$ and $u_{\mathrm{min}}: =\min_{f} u(f)$. Then the largest size angle corresponding to $\bu$ is 
\begin{equation}
  \min\big\{\arccos u_{\mathrm{max}}, \arccos \sqrt{1-u_{\mathrm{min}}^{2}}\big\}. \label{eqn:arccos}
\end{equation}
Because $u_{\mathrm{max}}^{2}+u_{\mathrm{min}}^{2}\leq 1$ for $F>1$, the expression in \eqref{eqn:arccos} equals $\arccos \sqrt{1-u_{\mathrm{min}}^{2}}$ and this
%\begin{equation}
%  \min\{\arccos \bu_{\mathrm{max}}, \arccos \sqrt{1-\bu_{\mathrm{min}}^{2}}\}=\arccos \sqrt{1-\bu_{\mathrm{min}}^{2}}.
%\end{equation}
  completes the proof.  
\end{IEEEproof}

\begin{lemma} \label{lem:lsv_special}
Define $f(\beta) :=\frac{1}{2}-\frac{\sin(2\beta)}{4\beta}$ and $g(\beta):=\frac{1}{2}+\frac{\sin(2\beta)}{4\beta}$ for $\beta \in \left(0,\frac{\pi}{2}\right]$.  Let $\be_{f}$, $f\in [F]$ be the unit vector with only the $f$-th entry being 1, and $C$ be the circular cone with basis vector $\bu=\be_{f}$, size angle being $\alpha$, and the inverse expectation parameter for the exponential distribution being $\lambda$. Then if the columns of the data matrix $\bV \in \mathbb{R}^{F\times N}$ are generated as in Theorem \ref{thm:prob} from $C$ ($K=1$) and with no projection to the nonnegative orthant (Step 4 in the generating process), we have 
\begin{equation}
  \mathbb{E}\left(\frac{\bV\bV^{T}}{N}\right)=\frac{\bD_{f}}{\lambda}
\end{equation}
where $\bD_{f}$ is a diagonal matrix with the $f$-th diagonal entry being $g(\alpha)$ and other diagonal entries being $f(\alpha)/(F-1)$.
\end{lemma}
\begin{IEEEproof}[Proof of Lemma \ref{lem:lsv_special}]
Each column $\bv_{n}$, $n\in [N]$ can be generated as follows: First, uniformly sample a $\beta_{n} \in [0,\alpha]$ and sample a positive scalar $l_{n}$ from the exponential distribution $\mathrm{Exp}(\lambda)$, then we can write $\bv_{n}=\sqrt{l_{n}}\left[\cos\beta_{n} \be_{f}+\sin\beta_{n}\by_{n}\right]$, where $\by_{n}$ can be generated from sampling $y_{n}(1),\ldots,y_{n}(f-1),y_{n}(f+1),\ldots,y_{n}(F)$ from the standard normal distribution $\mathcal{N}(0,1)$, and setting $y_{n}(j)=y_{n}(j)/\sqrt{\sum_{i\neq f} y_{n}(i)^2}$, $j\neq f$, $y_{n}(f)=0$. Then 
\begin{align}
  & \mathbb{E}\left[ v_{n}(f_{1}) v_{n}(f_{2})\right] \nn\\
&\! \!=\mathbb{E}\big [ l_{n}  ((\cos^{2}\beta) e_{f}(f_{1})e_{f}(f_{2})\!   +\! (\sin^{2}\beta)y_{n}(f_{1})y_{n}(f_{2})  )\big] \\
&\!\! = \left\{
\begin{array}{rcl}
& 0,   & {f_{1}\neq f_{2}},\\
& g(\alpha)/\lambda,   & {f_{1}=f_{2}=f},\\
& f(\alpha)/\left((F-1)\lambda\right),   & {f_{1}=f_{2}\neq f}.
\end{array} \right. ,
\end{align}
where $e_f(f_1) = 1 \{ f=f_1\}$ is the $f_1$-th entry of the vector $\be_f$.
Thus $\mathbb{E}\left(\bV\bV^{T}/N\right)=\mathbb{E}\left(\bv_{n}\bv_{n}^{T}\right)=\bD_{f}/\lambda$.
\end{IEEEproof}

\begin{definition} \label{def:sub-gauss}
A {\em sub-gaussian random variable} $X$ is one that satisfies one of the following equivalent properties \\
1. Tails: $\mathbb{P}(|X|>t)\leq \exp\left(1-t^{2}/K_{1}^{2}\right)$ for all $t\geq 0$;\\
2. Moments: $\left(\mathbb{E}|X|^{p}\right)^{1/p} \leq K_{2} \sqrt{p}$ for all $p\geq 1$;\\
3. $\mathbb{E}\left[\exp\left(X^{2}/K_{3}^{2}\right)\right]\leq e$;\\
where $K_{i}, i=1,2,3$ are positive constants. The {\em sub-gaussian norm} of $X$, denoted $\|X\|_{\Psi_{2}}$, is defined to be
\begin{equation}
  \|X\|_{\Psi_{2}}:=\sup_{p\geq 1} p^{-1/2}\left(\mathbb{E}|X|^{p}\right)^{1/p}.
\end{equation} 
A random vector $X\in \mathbb{R}^{F}$ is called {\em sub-gaussian} if $X^{T}\bx$ is a sub-gaussian random variable for any constant vector $\bx\in \mathbb{R}^{F}$. The {\em sub-gaussian norm} of $X$ is defined as 
\begin{equation}
  \|X\|_{\Psi_{2}}=\sup \limits_{\|\bx\|_{2}=1}\|X^{T}\bx\|_{\Psi_{2}}.
\end{equation}
\end{definition}

\begin{lemma} \label{lem:gau_exp}
A random variable $X$ is sub-gaussian if and only if  $X^2$ is sub-exponential. Moreover, it holds that
\begin{equation}
  \|X\|_{\Psi_{2}}^{2} \leq \|X\|_{\Psi_{1}} \leq 2\|X\|_{\Psi_{2}}^{2}.
\end{equation}
\end{lemma}
See Vershynin~\cite{Ver_10} for the proof.
\begin{lemma}[Covariance  estimation of sub-gaussian distributions \cite{Ver_10}] \label{lem:cov_est}
Consider a sub-gaussian distribution  $\bbP$ in $\mathbb{R}^{F}$ with covariance matrix $\Sigma$. Let $\epsilon \in (0,1)$ and $ t\geq 1$.  If $N\geq c(t/\epsilon)^{2}F$, then with probability at least $1-2\exp(-t^{2}F)$,
\begin{equation}
  \|\Sigma_{N}-\Sigma\|_{2} \leq \epsilon ,
\end{equation}
where  $\|\cdot\|_2=\sigma_1(\cdot)$ is the spectral norm,  $\Sigma_{N}:=\sum_{n=1}^{N}X_{n}X_{n}^{T}/N$ is the empirical covariance matrix, and $X_{n}, n\in [N]$ are independent samples from $\bbP$. The constant $c=c_{g}$ depends only on the sub-gaussian norm $g=\|X\|_{\Psi_{2}}$ of the random vector  $X$ sampled from this distribution.
\end{lemma}

\begin{IEEEproof}[Proof of Theorem \ref{thm:prob_ext}] %\ref{thm:prob_ext}\\
Similar to Theorem \ref{thm:non-prob}, we have 
\begin{align}
   \frac{\|\bV-\bW^{*}\bH^{*}\|_{\rmF}^{2}}{\|\bV\|_{\rmF}^{2}} 
  &=  \frac{\sum_{k=1}^{K}\|\bV_{k}-\bw^{*}_{k}\bh_{k}^{T}\|_{\rmF}^{2}}{\sum_{k=1}^{K}\|\bV_{k}\|^{2}_{F}}\\
&  =  \frac{\sum_{k=1}^{K}\left(\|\bV_{k}\|^{2}_{F}-\sigma_{1}^{2}\left(\bV_{k}\right)\right)}{\sum_{k=1}^{K}\|\bV_{k}\|^{2}_{F}}\\
&  =    1-\frac{\sum_{k=1}^{K} \sigma_{1}^{2}\left(\bV_{k}\right)}{\sum_{k=1}^{K} \|\bV_{k}\|_{\rmF}^{2}}.
\end{align}
Take any $k \in [K]$ and consider $\sigma_{1}^{2}\left(\bV_{k}\right)$. Define the index $f_{k}:=\mathrm{argmin}_{f\in [F]} \bu_{k}$ and the orthogonal matrix $\bP_{k}$  as in~\eqref{eqn:house}. 
%\begin{equation}
%  \bP_{k}=1-2\bz_{k}\bz_{k}^{T}, \bz_{k}=\frac{\be_{f_{k}}-\bu_{k}}{\|\be_{f_{k}}-\bu_{k}\|_{2}} .
%\end{equation}
The columns of $\bV_{k}$ can be considered as Householder transformations of the data points generated from the circular cone $C_{f_{k}}^{0}:=\mathcal{C}\left(\be_{f_{k}},\alpha_{k}\right)$ (the circular cone with basis vector $\be_{f_{k}}$ and size angle $\alpha_{k}$), i.e., $\bV_{k}=\bP_{k}\bX_{k}$, where $\bX_{k}$ contains the corresponding data points in $C_{f_{k}}^{0}$. In addition, denoting $N_{k}$ as the number of data points in $\bV_{k}$, we have
\begin{align}
  \frac{\sigma_{1}^{2}\left(\bV_{k}\right)}{N_{k}} & =\frac{\sigma_{1}^{2}\left(\bV_{k}^{T}\right)}{N_{k}}= \lambda_{\max}\left(\frac{\bV_{k}\bV_{k}^{T}}{N_{k}}\right)
\end{align}
where $\lambda_{\max}\left(\bV_{k}\bV_{k}^{T}/N_{k}\right)$ represents the largest eigenvalue of $\bV_{k}\bV_{k}^{T}/N_{k}$.
Take any $\bv \in \bV_{k}$. Note that $\bv$ can be written as $\bv=\bP_{k}\bx$ with $\bx$ being generated from $C_{f_{k}}^{0}$. Now, for all unit vectors $\bz \in \mathbb{R}^{F}$, we have 
\begin{align}
   \|\bv\|_{\Psi_{2}} &=\|\bP_{k}\bx\|_{\Psi_{2}}=\|\bx\|_{\Psi_{2}}   \\
 & =   \sup \limits_{\|\bz\|_{2}=1} \sup \limits_{p\geq 1} p^{-1/2}\left(\mathbb{E}\left(|\bx^{T}\bz|^{p}\right)\right)^{1/p} &\\
&  \leq  \sup \limits_{p\geq 1} p^{-1/2} \mathbb{E}\left(\|\bx\|_{2}^{p}\right)^{1/p} &\\
&  =   \|\|\bx\|_{2}\|_{\Psi_{2}} \leq \sqrt{\|\|\bx\|_{2}^{2}\|_{\Psi_{1}}}\leq 1/\sqrt{\lambda_{k}} .
\end{align}
That is, all  columns are sampled from a sub-gaussian distribution. By Lemma \ref{lem:lsv_special}, 
\begin{equation}
  \mathbb{E}\left(\bv\bv^{T}\right)=\mathbb{E}\left(\bP_{k}\bx\bx^{T}\bP_{k}^{T}\right)=\bP_{k}\bD_{f_{k}}\bP_{k}^{T}/\lambda_{k}.
\end{equation}
By Lemma \ref{lem:cov_est}, we have for $\epsilon \in (0,1), t\geq 1$ and if $N_{k}\geq \xi_{k}(t/\epsilon)^{2}F$ ($\xi_{k}$ is a positive constant depending on $\lambda_{k}$), with probability at least $1-2\exp(-t^{2}F)$, 
\begin{align}
&  \left|\lambda_{\max}\left(\bV_{k}\bV_{k}^{T}/N_{k}\right)-\lambda_{\max}\left(\mathbb{E}\left(\bv\bv^{T}\right)\right)\right| \nn\\*
& \quad \leq  \|\bV_{k}\bV_{k}^{T}/N_{k}-\mathbb{E}\left(\bv\bv^{T}\right)\|_{2} \leq \epsilon,  \label{eqn:le_eps}
\end{align}
where the first inequality follows from Lemma \ref{lem:svd_pert}. Because $\lambda_{\max}\left(\mathbb{E}\left(\bv\bv^{T}\right)\right)=g(\alpha_{k} )/\lambda_{k}$, we can obtain that with probability at least $1-4K\exp(-t^{2}F)$, 
\begin{align}
&  \left|\sum_{k=1}^{K}  \frac{\sigma_{1}^{2}\left(\bV_{k}\right)}{ N}-\sum_{k=1}^{K}  \frac{g (\alpha_{k})  } { K\lambda_{k} }\right|\nn\\*
%&=  \left|\sum_{k=1}^{K}\frac{\sigma_{1}^{2}\left(\bV_{k}\right)}{N_{k}}\frac{N_{k}}{N}-\sum_{k=1}^{K} \frac{g(\alpha_{k} )}{K\lambda_{k}}\right| \\
&=    \left|\sum_{k=1}^{K}\lambda_{\max}\left(\frac{\bV_{k}\bV_{k}^{T}}{N_{k}}\right)\frac{N_{k}}{N}-\sum_{k=1}^{K} \frac{g(\alpha_{k} )}{K\lambda_{k}}\right|\\
&\leq   2K\epsilon,
\end{align}
where the final inequality follows from the triangle inequality and  \eqref{eqn:le_eps}.
From the proof of Theorem \ref{thm:prob}, we know that with probability at least $1-2\exp(-c_{1}N\epsilon^{2})$, 
\begin{equation}
  \left|\frac{\|\bV\|_{\rmF}^{2}}{N}-\frac{\sum_{k=1}^{K} 1/\lambda_{k} }{K}\right| \leq \epsilon.
\end{equation}
Taking $N$ to be sufficiently large such that $t^{2}F\leq c_{1}N\epsilon^{2}$, we have with probability at least $1-6K\exp(-t^{2}F)$, 
\begin{align}
  \frac{\sum_{k=1}^{K}g(\alpha_{k}  )/\lambda_{k}}{\sum_{k=1}^{K}1/\lambda_{k}}-c_{2}\epsilon &  \leq \frac{\sum_{k=1}^{K} \sigma_{1}^{2}\left(\bV_{k}\right)}{\sum_{k=1}^{K} \|\bV_{k}\|_{\rmF}^{2}}  \\
&  \leq   \frac{\sum_{k=1}^{K}g(\alpha_{k}  )/\lambda_{k}}{\sum_{k=1}^{K}1/\lambda_{k}}+c_{3}\epsilon. 
\end{align}
Note that $g(\alpha_{k}  )+f(\alpha_{k}  )=1$. As a result, we have 
\begin{align}
  \frac{\sum_{k=1}^{K}f(\alpha_{k}  )/\lambda_{k}}{\sum_{k=1}^{K}1/\lambda_{k}}-c_{3}\epsilon 
&  \leq   \frac{\|\bV-\bW^{*}\bH^{*}\|_{\rmF}^{2}}{\|\bV\|_{\rmF}^{2}   }\\
&  \leq  \frac{\sum_{k=1}^{K}f(\alpha_{k}  )/\lambda_{k}}{\sum_{k=1}^{K}1/\lambda_{k}}+c_{2}\epsilon .
\end{align}
Thus, with probability at least $1-6K\exp(-t^{2}F)$, we have 
\begin{equation}
  \left|\frac{\|\bV-\bW^{*}\bH^{*}\|_{\rmF}}{\|\bV\|_{\rmF}}-\sqrt{\frac{\sum_{k=1}^{K}f(\alpha_{k}  )/\lambda_{k}}{\sum_{k=1}^{K}1/\lambda_{k}}}\right| \leq c_{4}\epsilon,
\end{equation}
where $c_{4}$ depends on $K$ and $\{(\alpha_{k},\lambda_{k}): k\in [K]\}$.
\end{IEEEproof}

\section{Proof of Theorem \ref{thm:deterK}} \label{app:prf_K}
We first state and prove  the following lemma. %\ref{thm:deterK}
\begin{lemma}\label{lem:svs}
Suppose data matrix $\bV$ is generated as in Theorem \ref{thm:prob} with all the circular cones being contained in $\mathcal{P}$, then the expectation of the covariance matrix $\bv_{1}\bv_{1}^{T}$ is
\begin{align}
  & \mathbb{E}\left[\bv_{1}\bv_{1}^{T}\right]=\frac{\sum_{k=1}^{K}f(\alpha_{k} )/\lambda_{k}}{K(F-1)}\bI    \nonumber \\
  & \quad +\frac{1}{K} \sum_{k=1}^{K} \frac{ g(\alpha_{k} )-f(\alpha_{k} )/(F-1)}{\lambda_{k}} \bu_{k}\bu_{k}^{T}, \label{eqn:Evv}
\end{align}
where $\bv_{1}$ denotes the first column of $\bV$.
\end{lemma} 
\begin{IEEEproof}
From the proof in Lemma \ref{lem:lsv_special}, we know if we always take $\be_{1}$ to be the original vector for the Householder transformation, the corresponding Householder matrix for the $k$-th circular cone $\mathcal{C}_{k}$ is given by \eqref{eqn:house} and we have 
\begin{equation}
  \mathbb{E}\left[\bv_{1}\bv_{1}^{T}\right]=\frac{1}{K}\sum_{k=1}^{K}\frac{ \bP_{k}\bD_{k}\bP_{k}^{T}}{\lambda_{k}},
\end{equation}
where $\bD_{k}$ is a diagonal matrix with the first diagonal entry being $g(\alpha_{k} ):=\frac{1}{2}+\frac{\sin(2\alpha_{k}) }{4\alpha_{k}}$ and other diagonal entries are 
\begin{equation}
  \frac{f(\alpha_{k} )}{F-1}=\frac{\frac{1}{2}- \frac{\sin(2\alpha_{k}) }{4\alpha_{k}}}{F-1}.
\end{equation}
We simplify $\bP_{k}\bD_{k}\bP_{k}^{T}$ using the property that all the $F-1$ diagonal entries of $\bD_{k}$ are the same. Namely, we can write
\begin{align}
    \bP_{k}&=\bI-2\bz_{k}\bz_{k}^{T}=\bI-\frac{(\be_{1}-\bu_{k})(\be_{1}-\bu_{k})^{T}}{1-u_{k}(1)} \\
  & = \left[
\begin{matrix}
 u_{k}(1)      & u_{k}(2)      & \cdots & u_{k}(F)      \\
 u_{k}(2)      & 1-\frac{u_{k}(2)^{2}}{1-u_{k}(1)}      & \cdots & -\frac{u_{k}(2)u_{k}(F)}{1-u_{k}(1)}      \\
 \vdots & \vdots & \ddots & \vdots \\
 u_{k}(F)      & -\frac{u_{k}(F)u_{k}(2)}{1-u_{k}(1)}      & \cdots & 1-\frac{u_{k}(F)^{2}}{1-u_{k}(1)}       \\
\end{matrix}
\right].
\end{align}
Note that $\bP_{k}=\left[\bp_{1}^{k},\bp_{2}^{k},\ldots,\bp_{F}^{k}\right]$ is symmetric and the first column of $\bP_{k}$ is $\bu_{k}$. Let  $\bD_{k}$ be the diagonal matrix with diagonal entries being $d_{1},d_{2},\ldots,d_{F}$. Then we have 
\begin{align}
    \bP_{k}\bD_{k}\bP_{k}^{T}&=\sum_{j=1}^{K} d_{j}\bp_{j}^{k}(\bp_{j}^{k})^{T} \\
%&= d_{1}\bu_{k}\bu_{k}^{T}+\sum_{j=2}^{K} d_{j}\bp_{j}^{k}(\bp_{j}^{k})^{T}  \\
&= d_{1}\bu_{k}\bu_{k}^{T}+d_{2}\sum_{j=2}^{K} \bp_{j}^{k}(\bp_{j}^{k})^{T}   \\
&= g\left(\alpha_{k}\right)\bu_{k}\bu_{k}^{T}+\frac{f(\alpha_{k} )}{F-1} \left(\bI-\bu_{k}\bu_{k}^{T}\right) \\
&= \frac{f(\alpha_{k} )}{F-1}\bI+\left(g(\alpha_{k} )-\frac{f(\alpha_{k} )}{F-1}\right)\bu_{k}\bu_{k}^{T} . 
\end{align}
Thus, we obtain \eqref{eqn:Evv} as desired.  
%\begin{align}
%  & \mathbb{E}\left[\bv_{1}\bv_{1}^{T}\right]=\frac{\sum_{k=1}^{K}f\left(\alpha_{k}\right)/\lambda_{k}}{K(F-1)}\bI + \nonumber \\
%  & \frac{1}{K} \sum_{k=1}^{K} \frac{\left(g\left(\alpha_{k}\right)-f\left(\alpha_{k}\right)/(F-1)\right)}{\lambda_{k}} \bu_{k}\bu_{k}^{T} 
%\end{align} 
\end{IEEEproof}
We are now ready to prove Theorem \ref{thm:deterK}.
\begin{IEEEproof}[Proof of Theorem \ref{thm:deterK}] %\ref{thm:deterK}\\
Define 
\begin{align}
  a&:=\frac{\sum_{k=1}^{K}f(\alpha)/\lambda}{K(F-1)}=\frac{f(\alpha)/\lambda}{F-1},\;\;\mbox{and}\\
  b&:=\frac{g(\alpha)-f(\alpha)/(F-1) }{K\lambda }.
\end{align}
By exploiting the assumption that all the $\alpha_{k}$'s and $\lambda_{k}$'s are the same, we find that 
\begin{equation}
  \mathbb{E}\left[\bv_{1}\bv_{1}^{T}\right]=a\bI + b\sum_{k=1}^{K}\bu_{k}\bu_{k}^{T}.
\end{equation}
Let $\bU=\left[\bu_{1},\bu_{2},\ldots,\bu_{K}\right]$. We only need to consider the eigenvalues of $\sum_{k=1}^{K}\bu_{k}\bu_{k}^{T}=\bU\bU^{T}$. The matrix $\bU^{T}\bU$ has same non-zero eigenvalues as that of $\bU\bU^{T}$. Furthermore,
\begin{align}
  \bU^{T}\bU&=
  \left[ 
\begin{matrix}
 1      & \cos \beta      & \cdots & \cos \beta      \\
 \cos \beta      & 1      & \cdots & \cos \beta      \\
 \vdots & \vdots & \ddots & \vdots \\
 \cos \beta      & \cos \beta      & \cdots & 1       \\
\end{matrix}
\right] \\
&=( \cos\beta )\be\be^{T}+(1-\cos\beta)\bI 
\end{align}
where $\be\in\mathbb{R}^{K}$ is the vector with all entries being 1. Therefore, the eigenvalues of $\bU^{T}\bU$ are $1+(K-1)\cos\beta,1-\cos\beta,\ldots,1-\cos\beta$.
Thus, the vector  of eigenvalues of $\mathbb{E}\left[\bv_{1}\bv_{1}^{T}\right]$ is  $[a+b(1+(K-1)\cos\beta),a+b(1-\cos\beta),\ldots,a+b(1-\cos\beta),a,a,\ldots,a ] $. 

By Lemmas \ref{lem:svd_pert} and   \ref{lem:cov_est}, we deduce  that for any $t\geq 1$ and a sufficiently small $\epsilon>0$, such that
\begin{equation}\label{eq:epsilon}
 \frac{a+\epsilon}{a-\epsilon} < \frac{a+b(1-\cos\beta)-\epsilon}{a+\epsilon},
\end{equation}
then if $N\geq c(t/\epsilon)^{2}F$ (where $c>0$ depends only on $\lambda$, $\alpha$, and $\beta$),  then with probability at least $1-2\left(K_{\max}-K_{\min}+1\right)\exp\left(-t^{2}F\right)$,  Eqn.~\eqref{eqn:sigma_max} holds. 
%  \begin{equation}
%    \frac{\sigma_{K}}{\sigma_{K+1}}=\max_{j \in \left[K_{\min},K_{\max}\right]} \frac{\sigma_{j}}{\sigma_{j+1}},
%  \end{equation} 
%  where $c$ is a positive constant depending only on $\lambda$, $\alpha$ and $\beta$.
\end{IEEEproof}

\section{Invariance of $(\bW^*,\bH^*)$} \label{app:inv}
\begin{lemma}\label{lem:fix_pt}
The $\left(\bW^{*},\bH^{*}\right)$ pair generated by Algorithm \ref{algo:approx} remains unchanged in the iterations of standard multiplicative update algorithm~\cite{Lee_00} for NMF.
\end{lemma}
\begin{IEEEproof}
There is at most one non-zero entry in each column of $\bH^{*}$. When updating $\bH^{*}$, the zero entries remain zero. For the non-zero entries of $\bH^{*}$, we consider partitioning $\bV$ into $K$ submatrices corresponding to the $K$ circular cones. Clearly,
\begin{equation}
  \|\bV-\bW^{*}\bH^{*}\|_{\rmF}^{2}=\sum_{k=1}^{K} \|\bV_{k}-\bw_{k}\bh_{k}^{T}\|_{\rmF}^{2},
\end{equation}
where $\bV_{k} \in \mathbb{R}^{F\times |\mathcal{I}_{k}|}$ and $\bh_{k} \in \mathbb{R}^{|\mathcal{I}_{k}|}_{+}$. Because  of the property of rank-one NMF (Lemma \ref{lem:rank_one}), for any $k$, when $\bw_{k}$ is fixed, $\bh_{k}\in \mathbb{R}^{|\mathcal{I}_{k}|}_{+}$ minimizes $\|\bV_{k}-\bw_{k}\bh^{T}\|_{\rmF}^{2}$. Also, for the standard multiplicative update algorithm, the objective function is non-increasing for each update~\cite{Lee_00}. Thus $\bh_{k}$ for each $ k \in[K]$ (i.e., $\bH^*$) will remain unchanged.  A completely symmetric argument holds for $\bW^*$. %For the update of $\bW^{*}$ given $\bH^{*}$, similarly, because the objective function is non-increasing, $\bW^{*}$ will keep unchanged.  
\end{IEEEproof}
\subsection*{Acknowledgements}

The authors would like to thank the three anonymous reviewers for their excellent and detailed comments that helped to improve the presentation of the results in the paper.

\bibliographystyle{unsrt}
\bibliography{mydraft_ref}

\begin{thebibliography}{10}

\bibitem{Ci_09}
A.~Cichocki, R.~Zdunek, A.~Phan, and S.~Amari.
\newblock {\em Nonnegative Matrix and Tensor Factorizations: Applications to
  Exploratory Multi-Way Data Analysis and Blind Source Separation}.
\newblock John Wiley \& Sons, 2009.

\bibitem{Bu_08}
I.~Buciu.
\newblock Non-negative matrix factorization, a new tool for feature extraction:
  theory and applications.
\newblock {\em Int. J. Comput. Commun.}, 3:67--74, 2008.

\bibitem{Lee_00}
D.~D. Lee and H.~S. Seung.
\newblock Algorithms for non-negative matrix factorization.
\newblock In {\em Proc.\ NIPS}, pages 556--562, 2000.

\bibitem{Chu_04}
M.~Chu, F.~Diele, R.~Plemmons, and S.~Ragni.
\newblock Optimality, computation, and interpretation of nonnegative matrix
  factorizations.
\newblock {\em SIAM J. Matrix Anal.}, 2004.

\bibitem{Kim_08b}
H.~Kim and H.~Park.
\newblock Nonnegative matrix factorization based on alternating nonnegativity
  constrained least squares and active set method.
\newblock {\em SIAM J. Matrix Anal. A.}, 30(2):713--730, 2008.

\bibitem{Kim_08a}
J.~Kim and H.~Park.
\newblock Toward faster nonnegative matrix factorization: A new algorithm and
  comparisons.
\newblock In {\em Proc.\ ICDM}, pages 353--362, Dec 2008.

\bibitem{Kim_11}
J.~Kim and H.~Park.
\newblock Fast nonnegative matrix factorization: An active-set-like method and
  comparisons.
\newblock {\em SIAM J. Sci. Comput.}, 33(6):3261--3281, 2011.

\bibitem{Cichocki07}
A.~Cichocki, R.~Zdunek, and S.~I. Amari.
\newblock Hierarchical {ALS} algorithms for nonnegative matrix and 3d tensor
  factorization.
\newblock In {\em Proc.\ ICA}, pages 169--176, Sep 2007.

\bibitem{Ho_11}
N.-D. Ho, P.~{Van Dooren}, and V.~D. Blondel.
\newblock {\em Descent methods for Nonnegative Matrix Factorization}.
\newblock Springer Netherlands, 2011.

\bibitem{Va_09}
S.~A. Vavasis.
\newblock On the complexity of nonnegative matrix factorization.
\newblock {\em SIAM J. Optim.}, 20:1364--1377, 2009.

\bibitem{Lin_07a}
C.-J. Lin.
\newblock Projected gradient methods for nonnegative matrix factorization.
\newblock {\em Neural Comput.}, 19(10):2756--2779, Oct. 2007.

\bibitem{Lin_07b}
C.-J. Lin.
\newblock On the convergence of multiplicative update algorithms for
  nonnegative matrix factorization.
\newblock {\em IEEE Trans. Neural Netw.}, 18(6):1589--1596, Nov 2007.

\bibitem{Donoho_04}
D.~Donoho and V.~Stodden.
\newblock When does non-negative matrix factorization give correct
  decomposition into parts?
\newblock In {\em Proc.\ NIPS}, pages 1141--1148. MIT Press, 2004.

\bibitem{Arora_12}
S.~Arora, R.~Ge, R.~Kannan, and A.~Moitra.
\newblock Computing a nonnegative matrix factorization--provably.
\newblock In {\em Proc.\ STOC}, pages 145--162, May 2012.

\bibitem{Gillis_14a}
N.~Gillis and S.~A. Vavasis.
\newblock Fast and robust recursive algorithms for separable nonnegative matrix
  factorization.
\newblock {\em IEEE Trans. Pattern Anal. Mach. Intell.}, 36(4):698--714, 2014.

\bibitem{Bittorf_12}
V.~Bittorf, B.~Recht, C.~R\'{e}, and J.~A. Tropp.
\newblock Factoring nonnegative matrices with linear programs.
\newblock In {\em Proc.\ NIPS}, pages 1214--1222, 2012.

\bibitem{Kumar_13}
A.~Kumar, V.~Sindhwani, and P.~Kambadur.
\newblock Fast conical hull algorithms for near-separable non-negative matrix
  factorization.
\newblock In {\em Proc.\ ICML}, pages 231--239, Jun 2013.

\bibitem{Ben_14}
A.~Benson, J.~Lee, B.~Rajwa, and D.~Gleich.
\newblock Scalable methods for nonnegative matrix factorizations of
  near-separable tall-and-skinny matrice.
\newblock In {\em Proc.\ NIPS}, pages 945--953, 2014.

\bibitem{Gillis_14b}
N.~Gillis and R.~Luce.
\newblock Robust near-separable nonnegative matrix factorization using linear
  optimization.
\newblock {\em J. Mach. Learn. Res.}, 15(1):1249--1280, 2014.

\bibitem{Gon_09}
E.~F. Gonzalez.
\newblock {\em Efficient alternating gradient-$type$ algorithms for the
  approximate non-negative matrix factorization problem}.
\newblock PhD thesis, Rice University, Houston, Texas, 2009.

\bibitem{Ack_09}
M.~Ackerman and S.~Ben-David.
\newblock Clusterability: A theoretical study.
\newblock In {\em Proc.\ AISTATS}, volume~5, pages 1--8, 2009.

\bibitem{Wild04}
S.~Wild, J.~Curry, and A.~Dougherty.
\newblock Improving non-negative matrix factorizations through structured
  initialization.
\newblock {\em Pattern Recognit.}, 37:2217--2232, 2004.

\bibitem{Bou08}
C.~Boutsidis and E.~Gallopoulos.
\newblock {SVD} based initialization: A head start for nonnegative matrix
  factorization.
\newblock {\em Pattern Recognit.}, 41:1350--1362, 2008.

\bibitem{Dhi01}
I.~S. Dhillon and D.~S. Modha.
\newblock Concept decompositions for large sparse text data using clustering.
\newblock {\em Mach. Learn.}, 42:143--175, 2001.

\bibitem{Xue08}
Y.~Xue, C.~S. Chen, Y.~Chen, and W.~S. Chen.
\newblock Clustering-based initialization for non-negative matrix
  factorization.
\newblock {\em Appl. Math. Comput.}, 205:525--536, 2008.

\bibitem{Zheng07}
Z.~Zheng, J.~Yang, and Y.~Zhu.
\newblock Initialization enhancer for non-negative matrix factorization.
\newblock {\em Eng. Appl. Artif. Intell.}, 20:101--110, 2007.

\bibitem{Langville06}
A.~N. Langville, C.~D. Meyer, and R.~Albright.
\newblock Initializations for the nonnegative matrix factorization.
\newblock In {\em Proc.\ SIGKDD}, pages 23--26, Aug 2006.

\bibitem{Golub_89}
G.~H. Golub and C.~F. {Van Loan}.
\newblock {\em Matrix computations}.
\newblock JHU Press, 1989.

\bibitem{Boser_92}
B.~E. Boser, I.~M. Guyon, and V.~N. Vapnik.
\newblock A training algorithm for optimal margin classifiers.
\newblock In {\em Proc.\ COLT}, pages 144--152, Jul 1992.

\bibitem{Ver_10}
R.~Vershynin.
\newblock Introduction to the non-asymptotic analysis of random matrices, 2010.
\newblock arXiv:1011.3027.

\bibitem{Tan13}
V.~Y.~F. Tan and C.~F\'evotte.
\newblock Automatic relevance determination in nonnegative matrix factorization
  with the $\beta$-divergence.
\newblock {\em IEEE Trans. on Pattern Anal. Mach. Intell.}, 35:1592--1605,
  2013.

\bibitem{Rou87}
P.~J. Rousseeuw.
\newblock Silhouettes: a graphical aid to the interpretation and validation of
  cluster analysis.
\newblock {\em J. Comput. Appl. Math.}, 20:53--65, 1987.

\bibitem{Moo00}
D.~Pelleg and A.~Moore.
\newblock X-means: Extending k-means with efficient estimation of the number of
  clusters.
\newblock In {\em Proc.\ ICML}, 2000.

\bibitem{Has01}
R.~Tibshirani, G.~Walther, and T.~Hastie.
\newblock Estimating the number of clusters in a data set via the gap
  statistic.
\newblock {\em J. R. Stat. Soc. Series B Stat. Methodol.}, 63:411--423, 2001.

\bibitem{Thorndike53whobelongs}
R.~L. Thorndike.
\newblock Who belongs in the family.
\newblock {\em Psychometrika}, pages 267--276, 1953.

\bibitem{Burden05}
R.~L. Burden and J.~D. Faires.
\newblock {\em Numerical Analysis}.
\newblock Thomson/Brooks/Cole, 8 edition, 2005.

\bibitem{Alex02}
A.~Strehl and J.~Ghosh.
\newblock Cluster ensembles--a knowledge reuse framework for combining multiple
  partitions.
\newblock {\em J. Mach. Learn. Res.}, pages 583--617, 2002.

\bibitem{Salton89}
G.~Salton.
\newblock {\em Automatic text processing : the transformation, analysis, and
  retrieval of information by computer}.
\newblock Reading: Addison-Wesley, 1989.

\bibitem{Manning08}
C.~D. Manning, P.~Raghavan, and H.~Schütze.
\newblock {\em Introduction to information retrieval}.
\newblock Cambridge University Press, 2008.

\bibitem{Li13}
Y.~Li and A.~Ngom.
\newblock The non-negative matrix factorization toolbox for biological data
  mining.
\newblock {\em Source Code Biol. Med.}, 8, 2013.

\bibitem{Jung}
A.~Jung, Y.~C. Eldar, and N.~G\"ortz.
\newblock On the minimax risk of dictionary learning.
\newblock {\em IEEE Trans. Inf. Theory}, 62(3):1501--1515, 2016.

\bibitem{ZhaoTan17}
R.~Zhao and V.~Y.~F. Tan.
\newblock Online nonnegative matrix factorization with outliers.
\newblock {\em IEEE Trans. Signal Process.}, 65(3):555--570, 2017.

\bibitem{ZhaoTanXu16}
R.~Zhao, V.~Y.~F. Tan, and H.~Xu.
\newblock Online nonnegative matrix factorization with general divergences.
\newblock In {\em Proc.\ AISTATS}, 2017.
\newblock arXiv:1608.00075.

\bibitem{arthur2007}
D.~Arthur and S.~Vassilvitskii.
\newblock $k$-means++: The advantages of careful seeding.
\newblock In {\em Proc.\ SODA}, pages 1027--1035, 2007.

\end{thebibliography}

%\vspace{-1.5cm}
\begin{IEEEbiography}[{\includegraphics[width=1in,height=1.25in,clip,keepaspectratio]{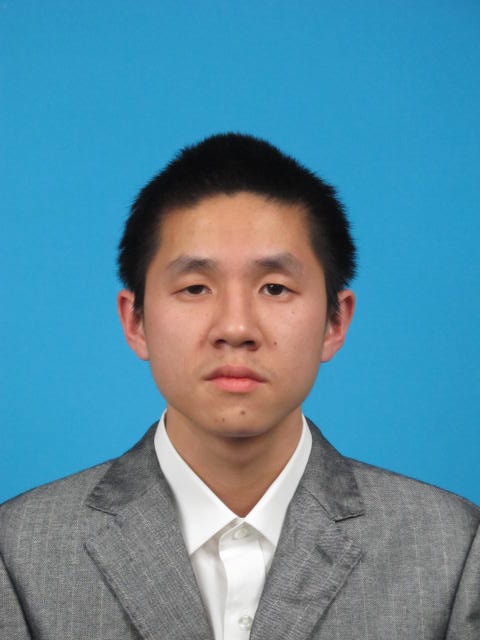}}] {Zhaoqiang Liu} was born in China in 1991. He is currently a Ph.D. candidate in the Department of Mathematics at National University of Singapore (NUS). He received the B.Sc.\ degree in Mathematics from the Department of Mathematical Sciences at Tsinghua University (THU) in 2013. His research interests are in machine learning, including unsupervised learning such as matrix factorization and deep learning. 
\end{IEEEbiography}

%\vspace{-1.5cm}

\begin{IEEEbiography}[{\includegraphics[width=1in,height=1.25in,clip,keepaspectratio]{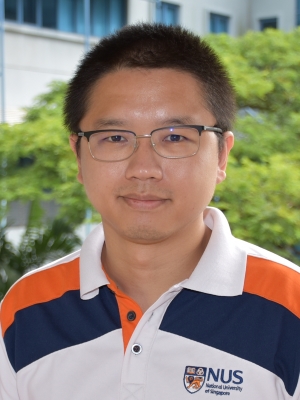}}] {Vincent Y.\ F.\ Tan} (S'07-M'11-SM'15) was born in Singapore in 1981. He is currently an Assistant Professor in the Department of Electrical and Computer Engineering (ECE) and the Department of Mathematics at the National University of Singapore (NUS). He received the B.A.\ and M.Eng.\ degrees in Electrical and Information Sciences from Cambridge University in 2005 and the Ph.D.\ degree in Electrical Engineering and Computer Science (EECS) from the Massachusetts Institute of Technology in 2011. He was a postdoctoral researcher in the Department of ECE at the University of Wisconsin-Madison and a research scientist at the Institute for Infocomm (I$^2$R) Research, A*STAR, Singapore. His research interests include network information theory, machine learning, and statistical signal processing. 

Dr.\ Tan received the MIT EECS Jin-Au Kong outstanding doctoral thesis prize in 2011 and the NUS Young Investigator Award in 2014. He has authored a research monograph on {\em ``Asymptotic Estimates in Information Theory with Non-Vanishing Error Probabilities''} in the Foundations and Trends in Communications and Information Theory Series (NOW Publishers). He is currently an Editor of the IEEE Transactions on Communications and the IEEE Transactions on Green Communications and Networking.
\end{IEEEbiography}
\end{document}